\documentclass{article}
\usepackage[a4paper, total={6.6in, 9.6in}]{geometry}

\PassOptionsToPackage{square, numbers, compress}{natbib}
\usepackage[utf8]{inputenc} 
\usepackage[T1]{fontenc}    
\usepackage{hyperref}       
\usepackage{url}            
\usepackage{booktabs}       
\usepackage{amsfonts}       
\usepackage{nicefrac}       
\usepackage{microtype}      
\usepackage{xcolor}         
\usepackage{subcaption}

\usepackage{tabularx, makecell}

\usepackage{authblk}

\usepackage{amsmath,amssymb}
\usepackage{amsthm}
\usepackage{mathtools}

\usepackage{verbatim}
\usepackage{bm}

\newcommand{\bi}[1]{\begin{itemize}#1\end{itemize}}
\newcommand{\eqn}[1]{\begin{align}#1\end{align}}

\newcommand{\mc}[1]{\mathcal{#1}}

\newcommand{\RR}{\mathbb{R}}

\newcommand{\e}[1]{^{(#1)}}

\newtheorem{theorem}{Theorem}
\newtheorem{lemma}[theorem]{Lemma}
\newtheorem{definition}[theorem]{Definition}
\newtheorem{recap}[theorem]{Known Result}
\newtheorem{corollary}[theorem]{Corollary}

\title{Value bounds and Convergence Analysis for Averages of LRP attributions}

\author[1]{Alexander Binder}
\author[2]{Nastaran Takmil-Homayouni}
\author[3]{Urun Dogan}

\affil[1]{ScaDS.AI \& Faculty of Math and CS\\
  University of Leipzig, Germany\footnote{  \texttt{alexander.binder@uni-leipzig.de}}}
\affil[2]{OvGU Magdeburg, Germany}  
\affil[3]{Microsoft Research, USA}

\begin{document}

\maketitle

\begin{abstract}

We analyze numerical properties of Layer-wise relevance propagation (LRP)-type attribution methods by representing them as a product of modified gradient matrices. This representation creates an analogy to matrix multiplications of Jacobi-matrices which arise from the chain rule of differentiation.
In order to shed light on the distribution of attribution values, we derive upper bounds for singular values. Furthermore we derive component-wise bounds for attribution map values. 
As a main result, we apply these component-wise bounds to obtain multiplicative constants. These constants govern the convergence of empirical means of attributions to expectations of attribution maps.
This finding has important implications for scenarios where multiple non-geometric data augmentations are applied to individual test samples, as well as for Smoothgrad-type attribution methods. In particular, our analysis reveals that the constants for LRP-$\beta$ remain independent of weight norms, a significant distinction from both gradient-based methods and LRP-$\epsilon$.

\end{abstract}

\section{Introduction}

%



In various domains such as healthcare or the sciences it is  not only important to achieve high predictive accuracies but it also matters in some use cases to understand what part of an input sample contributed to the prediction. To this end the field of explainable deep learning has developed several algorithms to explain predictions. Early approaches in deep learning considered gradient-based attributions \cite{DBLP:journals/corr/SimonyanVZ13}.

Several attribution methodologies for deep neural networks are based on the idea of using a modified gradient such as \cite{Springenberg,gradcam,DBLP:conf/icml/ShrikumarGK17} in order to address shortcomings of gradients in deep neural networks such as high noise content \cite{DBLP:conf/icml/BalduzziFLLMM17}. As part of modified gradient approaches, attribution methods based on Layer-wise relevance propagation (LRP)  \cite{Bach_15} have consistently produced explanations with high faithfulness to network output scores across diverse deep neural architectures, including CNNs\cite{Lapuschkin2019, Binder_2023_CVPR}, Transformers\cite{pmlr-v235-achtibat24a}, and Mamba-type networks \cite{NEURIPS2024_d6d0e41e}. However, the theoretical underpinnings of their properties and the mechanisms driving their high faithfulness remain insufficiently understood.

\bi{

\item 
We derive upper and lower bounds for the value ranges of two LRP-type attributions. We establish a formal framework by analyzing transition matrices for LRP-type attributions analogously to Jacobian matrices for gradient-based methods.  

\item 
We establish convergence properties for LRP-type attribution maps in settings involving predictions with augmentations of independently sampled data. Our analysis demonstrates that while the LRP attribution maps converge at the same asymptotic rate of $\mathcal{O}(1/\sqrt{m})$  with respect to the sample size $m$  as gradient methods, the constant factors governing this convergence differ fundamentally. In particular, for the $\beta$-rule, these constants are decoupled from weight norms - a critical distinction that grants robustness against large model weights. This theoretical finding explains the empirically observed low sensitivity of LRP to top-down model parameter randomization tests \cite{DBLP:conf/nips/AdebayoGMGHK18} reported in prior work \cite{Binder_2023_CVPR}.

}

\section{Related Work}

A number of methods can be applied scalably to deep neural networks which modify the gradient such as Guided Backpropagation \cite{Springenberg} and Grad-CAM \cite{gradcam} or which define attributions which share certain properties of the gradient \cite{DBLP:conf/iclr/AnconaCO018} such as DeepLIFT \cite{DBLP:conf/icml/ShrikumarGK17} and LRP \cite{Bach_15}. Other methods employ the gradient or gradient times input in input space and devise schemes to reduce the noise content of the gradient, such as Integrated Gradients \cite{DBLP:conf/icml/SundararajanTY17} and SmoothGrad \cite{DBLP:journals/corr/SmilkovTKVW17}. Smoothing by adding noise and averaging has been applied to LRP as well \cite{DBLP:conf/esann/RaulfDHMF21}. Gradient-free alternatives can be based, among others, on Shapley values \cite{strumbelj2014, DBLP:conf/nips/LundbergL17}, occlusion approaches \cite{DBLP:conf/bmvc/PetsiukDS18} or learned perturbations \cite{DBLP:conf/iccv/FongV17}. \cite{DBLP:conf/accv/AgarwalN20} advocates for generative inpainting. 

Due to the lack of ground truth, quality measures have been devised for attribution methods \cite{DBLP:conf/nips/Alvarez-MelisJ18,DBLP:conf/iclr/AnconaCO018,DBLP:journals/tnn/SamekBMLM17,DBLP:conf/nips/AdebayoGMGHK18,DBLP:conf/nips/YehHSIR19,DBLP:conf/ijcai/BhattWM20,DBLP:journals/corr/abs-2003-08747,DBLP:journals/corr/abs-2007-07584,DBLP:conf/icml/DasguptaFM22,DBLP:journals/corr/abs-2203-06877,DBLP:conf/icml/RongLBKK22, DBLP:journals/corr/abs-2401-06465, DBLP:conf/iccv/HesseS023,NEURIPS2024_b17799e0} covering various aspects. Importantly, satisfying them can often be achieved only with trade-offs \cite{NEURIPS2022_22b11181,Binder_2023_CVPR}.

The noise in gradients of deep neural nets has been quantified in \cite{DBLP:conf/icml/BalduzziFLLMM17,Woerl_2023_CVPR} as increasing in depth. \cite{pmlr-v139-agarwal21c} derived an $\mc{O}(m^{-1/2})$ convergence result for SmoothGrad as a function of the maximal gradient norm which implicitly depends on the network weights. A number of works have analyzed the effects of SmoothGrad, KernelShap, and others, for example from the perspective of smoothing \cite{NEURIPS2022_22b11181, zhou2024rethinkingprinciplegradientsmooth}. \cite{simpson2024probabilisticlipschitznessstablerank} uses Lipschitz-continuity valid with high probability and establishes a link between function and SmoothGrad attribution for this measure. 

\section{The problem setup: Convergence problems considered}





  We investigate the convergence properties of attribution maps when averaged over $m$ conditionally independent samples. Let $A(f,x) \in \RR^d $ be an attribution map for a classifier $f: \RR^d \rightarrow \RR^1$ in sample $x$. Then the quantity of interest is given as
\eqn{\label{eq:avg_base}
\frac{1}{m}\sum_{i=1}^m A(f,x\e{i})
}

A canonical application of this framework arises when analyzing predictions across multiple variants $x\e{i}$ of a single input sample $x$, where these variants are generated through data augmentation procedures.

\bi{
\item This procedure is known as Test-time data augmentation. Our study is motivated by the common use of Test-time data augmentation in medical imaging and sciences, see for example \cite{scalbert2022testtimeimagetoimagetranslationensembling,dippel2024aibasedanomalydetectionclinicalgrade,10.1007/978-3-031-16434-7_15,s22082988,gaillochet2022taal,MAHBOD2024669,jimaging8030071,jahanifar2021stainrobustmitoticfiguredetection,ma2024testtimegenerativeaugmentationmedical,shanmugam2021betteraggregationtesttimeaugmentation,ayhan2018testtime} . Specifically, we consider scenarios where a base sample $x$ undergoes multiple transformations via independently and identically distributed (i.i.d.) random augmentations $T_{c_i}(x)$ with parameters $c_i$ sampled from a distribution $\mathcal{Q}$:
\eqn{\label{eq:avg_random}
x\e{i} = T_{c_i}(x),\ c_i \sim \mathcal{Q}
}

These augmentations typically consist of photometric transformations that preserve the semantic content while altering superficial characteristics of the input. Notable examples include color-space transformations in histopathology imaging or spectral band mixing in hyperspectral remote sensing applications. 
In such cases, the stability of attribution maps across photometric variations provides insight into the model's reliance on invariant structural features versus incidental color or intensity patterns. 

\item Another use case for the above equation are Smoothgrad \cite{DBLP:journals/corr/SmilkovTKVW17} and SmoothLRP \cite{DBLP:conf/esann/RaulfDHMF21}.  These methods explicitly leverage statistical averaging of attribution maps computed over perturbed inputs, where each $x\e{i}$ is generated by adding independently sampled noise to the original test image.

}




\subsection{Base quantities and Notation}

Let $g \circ h $ denote the composition of functions.
 We consider a neural network of $n$ layers. 
\eqn{ \label{eq:f_k}
f_k(x) &= g_k\e{n} \circ \sigma \e{n-1} \circ g \e{n-1} \circ \sigma\e{n-2} \circ g\e{n-2} \circ \ldots \circ g\e{r} \circ  \ldots \circ g\e{1} (x)\\
z\e{r} &:= \sigma\e{r}\circ g\e{r-1} \circ \ldots \circ g\e{1} (x)
}
where $\sigma\e{r}$ is an activation function and $g\e{r}$ is a neural network layer, which is usually an affine transformation and which receives a feature map $z\e{r-1}$ as input.

We assume that one layer is a mapping $g: \RR^S \longrightarrow \RR^R$. We consider here affine layers such as convolution layers or fully connected layers, where $u \cdot v$ denotes here the usual Euclidean inner product
\eqn{
g(z) = W z + b = ( g_1(z),\ldots, g_R(z) ) = (W_{1,:} \cdot z +b_1, \ldots, W_{R,:} \cdot z +b_R )
}

\subsubsection{Gradient}

Using Jacobian matrices  $J g\e{r}$, $J \sigma\e{r}$ for the corresponding layers, the gradient of the above network for output component $f_k$ can be expressed as a series of matrix multiplications:
\eqn{
\label{eq:gradchainrule}
Df_k(x) & = \nabla^\top g_k\e{n}  J^\top\sigma\e{n-1}  \cdot J^\top g\e{n-1}  \cdot J^\top\sigma\e{n-2}  \cdot J^\top g\e{n-2} \cdot \ldots \cdot J^\top g\e{1} (x)
}
This uses chain rule to compute the derivative of the composite function $f_k(x)$ defined in equation \eqref{eq:f_k}.

\subsubsection{LRP}

LRP can be derived from the chainrule along a neural network graph. 
The neural network computation for $f_k(x)$ can be represented as a graph. Let $x_i \rightarrow g(\ldots,x_i,\ldots)$ be an edge in the forward pass of the neural network $f_k$. 

LRP-type modified gradients follow the same principle of chain-rule along graph edges as the gradient: For the gradient, an edge $x_i \rightarrow g(\ldots,x_i,\ldots)$ in the forward pass is assigned the partial derivative $\frac{\partial g}{\partial x_i}$ in the backward pass. In analogy to this, for LRP the assigned term in the backward pass is the corresponding modified gradient attribution $Att(g , x_i)$.

We define $Att(g_{a}, z_b) $ as the attribution of input $z_b$ for the output neuron $g_{a}$, using the analogy to the scalar partial derivative $\frac{\partial g_{a} }{\partial z_b}$ of output $g_{a}$ with respect to input $z_b$.

\paragraph{LRP-$\beta$}\cite{Bach_15} For LRP-$\beta$ the term $Att( g_{a} , z_{b}) $  is defined as:
\eqn{
Att( g_{a} , z_{b}) = (1+\beta) \frac{ (w_{ab} z_{b})_+  }{ \sum_{b'} (w_{ab'} z_{b'})_+  } - \beta \frac{ (w_{ab} z_{b})_-  }{ \sum_{b'} (w_{ab'} z_{b'})_-  }
}
where $(z)_+ = \max(z,0)$, $(z)_- = \min(z,0)$. It requires $\beta \ge 0$. 

Notably, $Att( g_{a} , z_{b})$  sums up over all inputs $z_{b}$ to $1+\beta -\beta = 1$ due to: 
\eqn{
\sum_b \frac{ (w_{ab} z_{b})_+  }{ \sum_{b'} (w_{ab'} z_{b'})_+  } = 1, \
\sum_b \frac{ (w_{ab} z_{b})_-  }{ \sum_{b'} (w_{ab'} z_{b'})_-  } = 1
}

\paragraph{LRP-$\gamma$}\cite{montavon2019layer} For LRP-$\gamma$ the term $Att( g_{a} , z_{b}) $  is defined as:
\eqn{
Att( g_{a} , z_{b})   =  \frac{ w_{ab} z_{b} + \gamma (w_{ab} z_{b})_+}{  \sum_{b'} w_{ab'} z_{b'} +  \gamma (w_{ab'} z_{b'})_+ } \label{eq:lrpgamma}
}
This method has been used recently to provide high faithfulness explanations for Transformer \cite{pmlr-v235-achtibat24a} and Mamba architectures \cite{NEURIPS2024_d6d0e41e}. It requires $\gamma \ge 0$. LRP-$\gamma$ also satisfies the property that $Att( g_{a} , z_{b})$  sums up $1$ over the set of all inputs $z_{b}$. 

LRP-$\gamma$ has a known convergence property towards LRP-$\beta$ with $\beta=0$:
\begin{recap}[Convergence of LRP-$\gamma$ attributions]\label{knownres:conv_lrpgamma}
\eqn{
&\lim_{\gamma \rightarrow \infty} \frac{ w_{ab} z_{b} + \gamma (w_{ab} z_{b})_+}{  \sum_{b'} w_{ab'} z_{b'} +  \gamma (w_{ab'} z_{b'})_+ } 
=  \frac{   (w_{ab} z_{b})_+}{  \sum_{b'}    (w_{ab'} z_{b'})_+ }
}
\end{recap}

Equation \eqref{eq:gradchainrule} expressed the derivative of a neural network using a product of Jacobian matrices. In analogy to the Jacobian matrix $J(g)$ for gradients we define here the corresponding matrix $M(g)$ for modified gradients:
\eqn{
J(g) &= \begin{pmatrix} \nabla g_1 , \nabla g_2 , \ldots , \nabla g_R
\end{pmatrix}\\
& = \begin{pmatrix}
\frac{\partial g_{1} }{\partial z_1} & \frac{\partial g_{2} }{\partial z_1} & \ldots & \frac{\partial g_{R} }{\partial z_1}  \\
\frac{\partial g_{1} }{\partial z_2} & \frac{\partial g_{2} }{\partial z_2} & \ldots & \frac{\partial g_{R} }{\partial z_2}   \\
\vdots & \vdots & \ddots & \vdots\\
\frac{\partial g_{1} }{\partial z_S} & \frac{\partial g_{2} }{\partial z_S} & \ldots & \frac{\partial g_{R} }{\partial z_S}  \\
\end{pmatrix} \\
M (g) &= \begin{pmatrix}
Att(g_{1} , z_1) & Att(g_{2} , z_1) & \ldots & Att( g_{R} , z_1)   \\
Att(g_{1} , z_2) & Att(g_{2} , z_2) & \ldots & Att( g_{R} , z_2)   \\
\vdots & \vdots & \ddots & \vdots\\
Att(g_{1} , z_S) & Att(g_{2} , z_S) & \ldots & Att( g_{R} , z_S)   \\
\end{pmatrix}
}

With this we can formulate an analogous result for LRP to equation \eqref{eq:gradchainrule}. Before stating it, we will use two common assumptions for LRP which are justified for example in \cite{montavon2019layer}:
\bi{
\item The attribution map for most LRP-type approaches uses in the backward pass an identity mapping for activation functions, even if the activation is not piece-wise linear. In practice this works also for GeLU units.
\item We assume the batch-normalization layers are fused into the subsequent MLP or convolution layers. This results in an equivalent network at inference time.
}
With this, we can express an attribution map computed using LRP in a matrix-based formalism.

An attribution map is usually computed for a prediction $f_u$ of a particular class $u$. This can be generalized to a weighted sum over multiple output classes. Lets assume that we apply LRP-type modified gradients to a weighted sum of network outputs $\sum_{u=1} q_u f_u(x)$ such that the weights for the network outputs satisfy $\sum_u q_u =1$. Therefore we can express the LRP attribution map, given the output initialization weight vector $q =(q_u)_u$ as
\eqn{
\sum_u q_u Att(f_u, x) &= q^\top M^\top (g\e{n}) \cdot M^\top (g\e{n-1}) \cdot M^\top (g\e{n-2}) \cdot \ldots \cdot M^\top (g\e{1}) (x) \label{eq:attrmap_equation} \\
\text{for } f(x) &= g\e{n} \circ \sigma \e{n-1} \circ g \e{n-1} \circ \sigma\e{n-2} \circ g\e{n-2} \circ  \ldots \circ g\e{1} (x)
}

Next we bring a definition for the property of $Att( g_{a} , z_{b})$ summing up to one.

\begin{definition}[Relevance conserving modified gradient method]

We say that a modified gradient method is relevance conserving, if every column of the modified gradient attribution matrix $M(g)$ sums up to 1. 
\end{definition}
This property will be crucial for proving the results shown below. It can be ensured for a non-zero attribution by normalization of attributions. Importantly, for any relevance-conserving attribution method we have
\eqn{
1_S^\top M = 1_R^\top
}

As a remark, $M(g)$ is not a stochastic matrix but rather a generalization to non-square matrix shapes and negative values.

\section{Analysis of Singular values}\label{sec:singularvalues}

Before we bring our main theoretical result in section \ref{sec:valueandconvergence}, we would like to show the results which can be obtained when we analysing LRP from the perspective of singular values of the above attribution matrices $M(g)$. 

This provides an easy to obtain insight into the scales of attribution map values across layers and may serve as an initial comparison to the gradient. It will also reveal a limitation of this SVD-based approach.

\begin{theorem}[Singular value for the vector of ones]

For any relevance conserving rule of a neural network layer which maps an $S$-dimensional input onto an $R$-dimensional output, a singular value of its one-layer transition is given by $\frac{\sqrt{R}}{\sqrt{S}}$, attained for the singular vector $\frac{1}{\sqrt{S}} 1_S = \frac{1}{\sqrt{S}}(\underbrace{1,\ldots,1}_{S \text{ times}})^\top$, where $R$ is the output dimension and $S$ the input dimension for the layer in consideration.
\end{theorem}

Proof:
\eqn{
\frac{1}{\sqrt{S}} 1_S^\top M M^\top \frac{1}{\sqrt{S}}1_S = \frac{1}{S} 1_R^\top 1_R = \frac{R}{S} 
}
The importance of this simple theorem is to show a dependence of the singular values on the output dimensionality $R$ of a layer. This motivates the insight, that observing a term $\sqrt{R}$ in the next theorem \ref{theorem:lrpbeta_svd_upperbound} is not an artefact of suboptimal proof technique but rather a necessity.


\begin{theorem}[Upper bound for singular values for LRP-$\beta$]\label{theorem:lrpbeta_svd_upperbound}
Let a neural network layer compute a mapping of an $S$-dimensional input onto an $R$-dimensional output. For the $\beta$-rule we can derive an upper bound on the singular values $\sqrt{R}\sqrt{(1+\beta)^2 + \beta^2}$, and as a better readable relaxation $\sqrt{R}(1+\sqrt{2}\beta)$
\end{theorem}

The proof for it is in the Supplemental material in Section \ref{sec:proof_lrpbeta_svd_upperbound}.\\

\begin{corollary}[Upper bound for singular values for LRP-$\gamma$ in the limit case]

Let a neural network layer compute a mapping of an $S$-dimensional input onto an $R$-dimensional output. In the limit of $\gamma \rightarrow \infty$ the upper bound of singular values for LRP-$\gamma$ is $\sqrt{R}$

\end{corollary}
This follows from the combination of Known Result \ref{knownres:conv_lrpgamma}, which establishes a convergence to LRP-$\beta$ with $\beta=0$, the fact that singular values of a real-valued matrix $M$ are the positive eigenvalues of a matrix
\eqn{
\begin{pmatrix}
0 & M \\
M^\top & 0
\end{pmatrix}
}
and a continuity result such as Weyl's eigenvalue bound for additive perturbations \cite{Weyl1912} which can be found in textbooks like \cite{horn_johnson_matrixanalysis}, and which ensures convergence when taking the limit $\gamma \rightarrow \infty$ for the above result $\sqrt{R}(1+\sqrt{2}\beta)$ for the case $\beta=0$.

\subsection{Comparison to the norm of the gradient attribution map}

Let us assume that we have Lipschitz continuity for the activation functions $\sigma_i$ with constant $L$. Then
\eqn{
&\|Dg \e{n} \cdot D \sigma \e{n-1}  \cdot D  g\e{n-1}  \cdot  \ldots  \cdot D \sigma \e{1} \cdot D g\e{1} (x)\|_2 \nonumber\\
&\le L^{n-1} \|Dg\e{n} \|_2  \ldots \|D g\e{1} (x)\|_2 = L^{n-1}  \|W\e{n} \|_2 \|W\e{n-1} \|_2 \ldots \|W\e{1} (x)\|_2
}

This scales as a function of the norms of the weights of a layer. For $\beta$-LRP we see an upper bound which is insensitive to weight norms:
\eqn{\label{eq:svdbound_beta}
&\|Mg\e{n} \cdot M \sigma\e{n-1}  \cdot M  g\e{n-1}  \cdot M \sigma\e{n-2}  \cdot M g\e{n-2} \cdot \ldots \cdot M g\e{1} (x)\|\nonumber\\
&\le \|Mg\e{n} \|_2 \|Mg\e{n-1} \|_2 \ldots \|M g\e{1} (x)\|_2 \le (1+\sqrt{2}\beta)^n \prod_l \sqrt{R_l}
}

\paragraph{Discussion:} There are two observations. Firstly, the independence of the singular values of the LRP-$\beta$ transition matrices of neural network weights $W$ shows a robustness property of LRP-$\beta$ and corresponds to an interpretation of LRP-$\beta$ attributions as an analogy of gradient clipping for modified gradients. 

Secondly, equation \eqref{eq:svdbound_beta} contains a term $\prod_l \sqrt{R_l}$ which depends on the output dimensions $R_l$ of each layer. This is a typically large quantity. Therefore, this might be of lesser value for deriving concentration inequalities. Therefore, we devise an improved, tighter, bound in the next section, using a different approach.

\section{Analysis of Value Ranges and Convergence Speed for  \texorpdfstring{LRP-$\beta$ and LRP-$\gamma$}{}}\label{sec:valueandconvergence}

We consider here averages of attribution maps, which arise when one predicts using multiple independent colorimetric augmentations $T_{c_i}$ of a test image $x$:
\eqn{
&\frac{1}{m}\sum_{i=1}^m A(f,x\e{i}), \ x\e{i} = T_{c_i}(x), \ c_i \sim \mathcal{Q}, \text{ independently}
}
In this case the distribution of the $x\e{i}$ is independent conditioned on $x$.

One natural candidate for quantifying convergence of this average towards its expectation is Hoeffding's inequality \cite{Hoeffding01031963}.
\begin{recap}[Hoeffding's inequality for identically distributed variables]\label{knownres_hoeffdings}

Let us assume that $Z\e{i}$ are iid with expectation $E[Z]$, with values almost surely in $[  z_l,  z_u]$: $ z_l \le Z\e{i} \le z_u$. Then:
\eqn{
P\left(  \left| \frac{1}{m}\sum_{i=1}^m Z\e{i} -  E[Z]  \right| \ge t    \right) \le 2 \exp\left(  -2 \frac{t^2m}{  (z_u-z_l)^2  }\right)
}
If the probability of large deviations is bounded by $\delta>0$, it results in a  valid with probability $1-\delta$:
\eqn{
t(\delta) =  (z_u-z_l)\sqrt{-\frac{1}{2}\ln\left(\frac{\delta}{2}\right)}\frac{1}{\sqrt{m}}
}
\end{recap}

As an application, for a fixed threshold of deviation $t$ we can use this to find out the required sample size $m$ as
\eqn{
m = (z_u-z_l)^2 \frac{1}{2}\ln\left(\frac{2}{\delta}\right) \frac{1}{t^2}
}
This tells us, that the average should contain the above amount $m$ of elements, in order to have a deviation of at most $t$ valid with probability of at least $1-\delta$ over draws of noised samples $x^{(i)}$ as defined in equations \eqref{eq:avg_base} and \eqref{eq:avg_random}. Using these results requires us to derive a bound on the value range $z_u-z_l$ of $Z\e{i}$ modulo events of zero measure, which we will do next.


For the next two lemmas it is important to understand the quantities used: $\sum_u q_u Att( g_u\e{n} , z_b\e{n-t})
$ corresponds to an attribution map for element $z_b$ of the vector of feature map values $z\e{n-t}$ from layer $n-t$, that is with reference to equation \eqref{eq:attrmap_equation}:
\eqn{
\sum_u q_u Att( g_u\e{n} , z\e{n-t}) &=  q^\top M^\top (g\e{n}) \cdot M^\top (g\e{n-1})  \cdot \ldots \cdot M^\top (g\e{n-t+1})(z\e{n-t}) \label{eq:attrmap_equation_part}
}
We will look at positive and negative elements of the corresponding attribution maps computed for the feature map values $z\e{n-t}$ from layer $n-t$ with respect to the weighted output logits $\sum_u q_u g_u\e{n}$ , that is
\eqn{
Z\e{n-t}_{+}:=\left\{z_b\e{n-t}:\, \sum_u q_u Att( g_u\e{n} , z_b\e{n-t}) > 0 \right\} \\
Z\e{n-t}_{-}:=\left\{z_b\e{n-t}:\, \sum_u q_u Att( g_u\e{n} , z_b\e{n-t}) < 0 \right\}
}

\begin{lemma}[Sequential bound for values under LRP-$\beta$]\label{lemma:seqbound_lrpbeta}

Suppose that the network has $n$ layers, the initializing weights $q_u$ at the output layer satisfy $\sum_u q_u =1, \ q_u \ge 0$. For LRP-$\beta$ with $\beta \ge 0$ the range of attribution map values at layer $n-t$ is given for each component $z_b\e{n-t}$ by
\eqn{
 \begin{matrix*}[l]     
 \sum_{z_b\e{n-t} \in Z\e{n-t}_{+}  } \sum_u q_u Att( g_u\e{n} , z_b\e{n-t}) \le +2^{t-1} (1+\beta)^{t} &  \text{ and } \\
 \sum_{z_b\e{n-t} \in Z\e{n-t}_{-} } \sum_u q_u Att( g_u\e{n} , z_b\e{n-t})  \ge  - 2^{t-1} \beta (1+\beta)^{t-1}  &  \text{ if }\beta >0\\
\sum_{b} \sum_u q_u Att( g_u\e{n} , z_b\e{n-t}) \in [0,1 ]& \text{ if }\beta=0
 \end{matrix*}
} 
\end{lemma}
The proof of Lemma \ref{lemma:seqbound_lrpbeta} is in Appendix Section \ref{sec:prooflemmaseqbound}.

The lemma states that the sum of positive attribution map scores is upper bounded by $+2^{t-1} (1+\beta)^{t}$, 

while the sum of negative attribution map scores
is lower bounded by $- 2^{t-1} \beta (1+\beta)^{t-1}$. 

In the special case of $\beta=0$ there are only non-negative contributions in the range of $[0,1]$.\\

Next we consider LRP-$\gamma$. It is easy to see in Equation \eqref{eq:lrpgamma} that one could have a divisor close to $0$ if $\gamma$ is too small, due to negative terms $w_{ab}z_b <0$. Therefore, we require a condition on $\gamma$ which keeps the contribution from negative activations to one neuron bounded relative to the positive ones. This condition is stated in equation \eqref{eq:lrpgammacondition} of the next lemma. 
\begin{lemma}[Sequential bound for values under LRP-$\gamma$]\label{lemma:seqbound_lrpgamma}

Suppose that the network has $n$ layers, the initializing weights $q_u$ at the output layer satisfy $\sum_u q_u =1, \ q_u \ge 0$. 

Furthermore we assume that $\gamma > 1$ is chosen large enough so that the following bound holds for all layers simultaneously which have positive connections in the sense of $\sum_{b: w_{ab}z_b>0} w_{ab}z_b > 0$ from the input to the output: 
\eqn{\label{eq:lrpgammacondition}
\gamma^{-1/2}\sum_{b: w_{ab}z_b<0} -w_{ab}z_b&< \sum_{b: w_{ab}z_b>0} w_{ab}z_b
}
Then the range of attribution map values at layer $n-t$ is given for each component $z_b\e{n-t}$ by
\eqn{
 \begin{matrix*}[l]
b(\gamma) &= \max(\frac{1}{  \gamma^{1/2}-1 } ,\frac{1+\gamma}{ 1+\gamma -\gamma^{1/2}} )\\
 \sum_{z_b\e{n-t} \in Z\e{n-t}_{+} } \sum_u q_u  Att( g_u\e{n} , z_b\e{n-t})&\le 2^{t-1} \frac{1+\gamma}{ 1+\gamma -\gamma^{1/2}}b(\gamma)^{t-1}   \\
 \sum_{z_b\e{n-t} \in Z\e{n-t}_{-}  } \sum_u q_u Att( g_u\e{n} , z_b\e{n-t})  &\ge   2^{t-1}\frac{-1}{(  \gamma^{1/2}-1 )} b(\gamma)^{t-1}
 \end{matrix*}
}
\end{lemma}
The proof of Lemma \ref{lemma:seqbound_lrpgamma} is in Appendix Section \ref{sec:prooflemmaseqbound_gamma}. For $\gamma \ge 4$ we have $b(\gamma)= \frac{1+\gamma}{ 1+\gamma -\gamma^{1/2}}$.

\paragraph{Value range for LRP-$\beta$:}

We can now calculate the value range $z_u-z_l$ required for Hoeffding's inequality for a network with $n$ layers. We have for LRP-$\beta$ the bound for the term $z_u-z_l$ appearing in Hoeffdings inequality for a network with $n$ layers given as:
\eqn{\label{eq:b-a_beta}
z_u-z_l = \begin{cases}
 2^{n-1}  (1+2\beta) (1+\beta)^{n-1}    & \beta > 0\\
1 & \beta = 0
\end{cases}
}
As a notable observation, this shows a lack of sensitivity to the norms of model weights, same as for the Singular value based analysis in section \ref{sec:singularvalues}. Unlike a bound derived from singular values, it does also not depend on the output dimensionality of layers $R$.

\paragraph{Comparison to a gradient-based bound:}

If we would compute $z_u-z_l$ for the gradient, we would obtain a term
\eqn{
z_u-z_l = 2 L^{n-1}  \|W_n \|_2 \|W_{n-1} \|_2 \ldots \|W_1 (x)\|_2. \label{eq:valuerange_gradient}
}
This bound for the gradient depends on the scale of weights $\|W_l \|_2$ in each layer, which may become large as a consequence of the high dimensionality of weights for each layer. Note that if $w_d \sim N(0, \sigma^2)$ and $\|W\|_2^2 = \sum_{d=1}^{R_l} w_d^2$, then it is known that $w_d^2$ is $\chi^2$-distributed with one degree of freedom and mean $\sigma^2$ and  $\|W\|_2^2$ is $\chi^2$-distributed with $R_l$ degrees of freedom and mean $R_l\sigma^2$. As such the expectation of $\|W_l\|_2^2$ is equal to $\sqrt{R_l}\sigma$ at initialization time. 

\begin{recap}[Expected norm of weights]
If $w_d \sim N(0, \sigma^2)$ and $W_l$ has dimensionality $R_l$, then $E[\|W_l\|_2] = \sqrt{R_l}\sigma$

\end{recap}

This is reminiscent of the bound obtained by SVD methods in Section \ref{sec:singularvalues}. It also shows that the gradient-based bounds will scale with $\prod_l \sqrt{R_l}$ and thus attain comparatively large values.

\paragraph{Value range for LRP-$\gamma$:}

 We have for LRP-$\gamma$ the bound for the term $z_u-z_l$ appearing in Hoeffdings inequality for a network with $n$ layers given as:
\eqn{\label{eq:b-a_gamma}
z_u-z_l = 
 2^{n-1}  b(\gamma)^{n-1} \left( \frac{1+\gamma}{ 1+\gamma -\gamma^{1/2}} + \frac{1}{\gamma^{1/2}-1}\right)
}
This is not fully independent of weights because the condition from equation \eqref{eq:lrpgammacondition} has to be satisfied. The value of $\gamma$ depends implicitly on the scale of activation values.


\section{Experimental validation of convergence speed}\label{sec:experiments_all}

The above convergence results
provide upper bounds on the deviation between an average $\frac{1}{m}\sum_{i=1}^m A(f,x\e{i}) $ and its expectation $E[  A(f,x)]$ via Known Result 6 and the bounds on $z_u-z_l$ in equations \eqref{eq:b-a_beta} and \eqref{eq:b-a_gamma}. In this section we measure empirically a lower bound on the deviation
\eqn{
&\left\|\frac{1}{m}\sum_{i=1}^m A(f,x\e{i,1})- \frac{1}{m}\sum_{k=1}^m A(f,x\e{k,2})\right\|_2 
}
\eqn{
=& \left\|\frac{1}{m}\sum_{i=1}^m A(f,x\e{i,1})- E [ A(f,x) ] + E [ A(f,x) ]-\frac{1}{m}\sum_{k=1}^m A(f,x\e{k,2})\right\|
}
\eqn{
\le&  \left\|\frac{1}{m}\sum_{i=1}^m A(f,x\e{i,1})- E [ A(f,x) ]\right\| + \left\|\frac{1}{m}\sum_{i=1}^m A(f,x\e{i,2})- E [ A(f,x) ]\right\|
}

The motivation to consider this lower bound is to verify experimentally a comparison of convergence of averages for gradient-based attribution maps and for LRP-based attribution maps. It also addresses the potential concern that our value range for the gradient in equation \eqref{eq:valuerange_gradient} might be less sophisticated compared to the bound for LRP-$\beta$ and LRP-$\gamma$. We investigate it by looking at computable lower bounds for all attribution maps.
This lower bound converges against the deviation:
\eqn{
\left\|\frac{1}{m}\sum_{i=1}^m A(f,x\e{i,1})- \frac{1}{n}\sum_{k=1}^n A(f,x\e{k,2})\right\|_2 \stackrel{n \rightarrow \infty}{\longrightarrow} \left\|\frac{1}{m}\sum_{i=1}^m A(f,x\e{i,1})- E[ A(f,x)]\right\|_2 
}

We will measure two statistics here. Each statistic will be computed for averages of the squared gradient, which is known as sensitivity attribution maps, averages of gradient times input, averages for LRP-$\beta$ with $\beta \in \{0,1\}$ and LRP-$\gamma$ with $\gamma \in \{100,1000\}$.

The first statistic is 
\eqn{\label{eq:diff_nonorm}
s_{1,m}(x) =&\left\|\frac{1}{m}\sum_{i=1}^m A(f,x\e{i,1})- \frac{1}{m}\sum_{k=1}^m A(f,x\e{k,2})\right\|_2 \ ,
}
where the samples $x\e{i,1}, x\e{k,2}$ in both sums come from two disjoint sets. This measures the statistics for unnormalized attribution maps. The concentration inequality above can be applied to it right away. However, a valid methodological concern arises from the fact that attribution maps generated by different techniques often exist on substantially different numerical scales, potentially confounding direct comparisons. To address this scaling issue and ensure fair comparative analysis, we additionally evaluate the differences between averages of $\ell_2$-normalized attribution maps. This normalization procedure isolates the directional properties of the attribution vectors from their magnitude, allowing us to quantify convergence characteristics in a scale-invariant manner that better captures the spatial distribution of feature importance.
\eqn{\label{eq:diff_l2norm}
s_{2,m}(x) = &\left\| \frac{\frac{1}{m}\sum_{i=1}^m A(f,x\e{i,1})}{\|\frac{1}{m}\sum_{i'}A(f,x\e{i',1})\|_2}-  \frac{\frac{1}{m}\sum_{k=1}^m A(f,x\e{k,2})}{\|\frac{1}{m}\sum_{k'} A(f,x\e{k',2})\|_2}\right\|_2
}

We deliberately avoid normalization by the maximum value of attribution maps, as this approach introduces heightened sensitivity to outliers and fails to provide meaningful constraints on the expected distribution of attribution scores. Although such normalization constrains the values to the $[-1,+1]$ interval, it does not offer guarantees regarding the statistical properties of the resulting distribution.

Instead, our adoption of $\ell_2$-normalization is motivated by the fundamental property that zero serves as the baseline value in many attribution methods, indicating the absence of influence on the prediction. As demonstrated in previous work \cite{Binder_2023_CVPR}, this normalization technique ensures that the mean square difference of the attribution values from zero is equal to one. Although this specific property falls outside of our derived value bounds for LRP-$\beta$ it establishes a principled basis for cross-method comparison by standardizing the mean deviation from the zero baseline. This approach facilitates more meaningful comparative analyses of attribution methods that may otherwise operate on incomparable numerical scales.

\subsection{Experimental details}\label{sec:experimental_details}

We consider three networks. ResNet-50 \cite{Resnet:he2016deep} and EfficientNet-V2-S \cite{Tan_21} are representatives of a classical and a more recent deep convolutional neural network, which we use with the pretrained weights provided in {\texttt torchvision}\cite{Paszke_19}. Furthermore we show the statistics for a Swin-V2-Tiny transformer network \cite{DBLP:conf/cvpr/Liu0LYXWN000WG22} as a representative of transformer-based models.
The experiments were done using PyTorch 2.6.0$+$cu124, torchvision 0.21$+$cu124 and two RTX A6000 GPUs. They required less than 47 GByte GPU Ram and 21 hours of time.

We consider photometric augmentation using {\texttt RandomPhotometricDistort} with boundaries $[0.875,1.125]$ for brightness, $[0.5,1.5]$ for contrast, $[0.8,1.2]$ for saturation and $[-0.1,0.1]$ for Hue. For each augmentation the parameters are drawn uniformly from the ranges shown above. For SmoothGrad-type additive augmentation we employ standard normal noise with a variance of $\sigma^2=1$.

The augmentations are applied $m=25, 50, 100$ times for one given image. The reason to use these seemingly small values of $m$ lies in the typical ranges for values of $m$ used in test-time averaging and in SmoothGrad \cite{DBLP:journals/corr/SmilkovTKVW17} and SmoothLRP. These are in the orders of tens of samples. Larger values of $m$ such as high hundreds would excessively slow down test time averaging and make it impractical in applications.

For each augmentation sample size $m$, this results in one average statistic $s_{1,m}(x)$, $s_{2,m}(x)$ based on a single image $x$ according to equations \eqref{eq:diff_nonorm} or \eqref{eq:diff_l2norm}. We compute these average statistics for the first 1000 images of the ImageNet validation set \cite{DenDon09Imagenet}, and report means in Section \ref{sec:exp_convergence} and boxplots in the appendix section \ref{app:stats_grad}. For comparing statistics, we thus employ 1000 paired samples for a pair consisting of one statistic for a gradient-based attribution, and one statistic for a LRP-based attribution

To avoid misunderstandings, with reference to $s_{1,m}(x)$, $s_{2,m}(x)$ from equations \eqref{eq:diff_nonorm} and \eqref{eq:diff_l2norm}, we are computing for each of $m=25, 50, 100$ 
the box plots and medians for the set 
\eqn{ \label{eq:statstomeasure}
\{ s_{1,m}(x), x \in S\}, \ \{ s_{2,m}(x), x \in S\}, |S|=1000
}
where each statistic $s_{1,m}(x)$ in the set $S$ is an average of $m$ attribution maps for data augmentations of the base sample $x$. We use the first 1000 images in the Imagenet validation set for samples $x$.

We compute these statistics for the gradient, for gradient-times-input, for LRP-$\beta=0$, LRP-$\beta=1$, LRP-$\gamma=10^3$ and LRP-$\gamma=10^2$. LRP-$\beta$ was computed while setting bias terms to zero in the backward pass (they were kept in the forward pass). We also included gradient-times-input, because it and its SmoothGrad-type variant often have a notably higher faithfulness compared to the plain squared gradient.
By using the SmoothGrad-type augmentation, we are measuring lower convergence bounds for SmoothGrad and SmoothLRP. The code is in the supplement.



\subsection{Experimental results}\label{sec:exp_convergence}

The results can be seen in Section \ref{ssec:unnorm} using the statistic $s_{1,m}(x)$ for attribution maps without normalization, and in Section \ref{ssec:l2norm} using the statistic $s_{2,m}(x)$  for $\ell_2$-normalization. The tables show two measures: 

The first measure is the ratio of the medians of the sets of statistics ($\{s_{1,m}(x), x \in S\}$ in Section \ref{ssec:unnorm}, $\{s_{2,m}(x), x \in S\}$ in Section \ref{ssec:l2norm}) obtained by Equations \eqref{eq:diff_nonorm} and \eqref{eq:diff_l2norm}. We compute one median for the squared gradient or the gradient $\times$ input, and one median for the LRP-based attribution maps. From that we take the ratio of the two medians. A ratio above 1 implies that the median of statistics for the gradient-based attribution maps is larger than the median for the LRP-based attribution maps. 

We use the median here because it aligns well with the statistical test used, which is a one-sided paired Wilcoxons signed rank test. Since the statistics are non-negative and converge towards zero, we refrain from using Gaussianity assumptions in statistical testing.

We use the one-sided paired Wilcoxons signed rank test on the differences between $s_{1,m}(x)$ computed for a gradient-based variant and for an LRP-variant in Section \ref{ssec:unnorm}. In Section \ref{ssec:l2norm} we apply the one-sided paired Wilcoxons signed rank test on the differences between $s_{2,m}(x)$ computed for a gradient-based variant and for an LRP-variant.

The second measure shown in the tables is the p-value from this statistical test. 

We can see two general outcomes:
\bi{
\item For the unnormalized results in Section \ref{ssec:unnorm}, computed from $s_{1,m}(x)$, the distances between the averages are always much larger for the gradient compared to both LRP-$\beta$ variants, and to both LRP-$\gamma$ variants.

We can see this by the large ratios in the order of hundreds. This implies a much faster convergence by all LRP variants. Note that we never verified for LRP-$\gamma$ whether the condition in Equation \eqref{eq:lrpgammacondition} holds. The comparison for LRP-$\gamma$ thus may include a number of samples which are not covered by Lemma \ref{lemma:seqbound_lrpgamma}.

\item For the $\ell_2$-normalized results in Section \ref{ssec:l2norm}, computed from $s_{2,m}(x)$, the picture is more mixed, yet with larger distances for the gradient when compared to LRP variants in the majority of cases. 

The distances between the averages are larger for the gradient compared to LRP-$\beta=0$ in all cases, compared to LRP-$\beta=1$ in most cases. They are larger for the gradient compared to LRP-$\gamma=10^3$ in the majority of cases. Note that this case is not covered by the results in Lemmata \ref{lemma:seqbound_lrpbeta} and \ref{lemma:seqbound_lrpgamma}. Still we can see ratios larger than $1$ in many cases indicating a faster convergence also in this normalized case.

}

\newcommand\Tstrut{\rule{0pt}{2.6ex}}         
\newcommand\Bstrut{\rule[-0.9ex]{0pt}{0pt}}
\newcolumntype{s}{>{\hsize=.5\hsize}X}

\subsection{Unnormalized case, covered by the theoretical results}\label{ssec:unnorm}

{ }

Effnet-V2-S, no normalization, Gradient, Comparison with LRP-$\beta$

\begin{tabularx}{\textwidth}{XXXXXX}\hline\rule{0pt}{1.0\normalbaselineskip}
Augmentation & Sample size  & \multicolumn{2}{c}{ Grad vs LRP-$\beta=0$} & \multicolumn{2}{c}{ Grad vs LRP-$\beta=1$}  \\
             &          &  p-value & ratio & p-value & ratio   \\\hline \rule{0pt}{2\normalbaselineskip}
Gaussian & \makecell[l]{ $m=25$ \\ $m=50$ \\ $m=100$  } & \makecell{  $1.7\cdot 10^{-165}$\\ $1.7\cdot 10^{-165}$ \\ $1.7\cdot 10^{-165}$ } & \makecell{ 2377.5 \\ 2860.0 \\ 3560.0 } & \makecell{ $1.7\cdot 10^{-165}$ \\ $1.7\cdot 10^{-165}$ \\ $1.7\cdot 10^{-165}$ } & \makecell{ 726.1 \\  875.3 \\ 1084.6 } \\
photometric & \makecell[l]{ $m=25$ \\ $m=50$ \\ $m=100$  }  &  \makecell{ $1.7\cdot 10^{-165}$ \\ $1.7\cdot 10^{-165}$ \\ $1.7\cdot 10^{-165}$ } & \makecell{ 5371.8 \\ 6336.5 \\ 7671.2  }  & \makecell{ $1.7\cdot 10^{-165}$ \\ $1.7\cdot 10^{-165}$ \\ $1.7\cdot 10^{-165}$ }  & \makecell{ 1640.7 \\ 1939.3  \\ 2337.2  } \\\hline
\end{tabularx}
\newpage

ResNet-50, no normalization, Gradient, Comparison with LRP-$\beta$

\begin{tabularx}{\textwidth}{XXXXXX}\hline\rule{0pt}{1.0\normalbaselineskip}
Augmentation & Sample size  & \multicolumn{2}{c}{ Grad vs LRP-$\beta=0$} & \multicolumn{2}{c}{ Grad vs LRP-$\beta=1$}  \\
             &          &  p-value & ratio & p-value & ratio   \\\hline \rule{0pt}{2\normalbaselineskip}
Gaussian & \makecell[l]{ $m=25$ \\ $m=50$ \\ $m=100$  } & \makecell{ $1.67\cdot 10^{-165}$ \\ $1.67\cdot 10^{-165}$ \\ $1.67\cdot 10^{-165}$  } & \makecell{447.1 \\ 545.9  \\ 655.3 } & \makecell{ $1.9\cdot 10^{-165}$ \\ $1.8\cdot 10^{-165}$ \\ $1.67\cdot 10^{-165}$ } & \makecell{ 69.2 \\ 84.5 \\ 101.6 } \\
photometric & \makecell[l]{ $m=25$ \\ $m=50$ \\ $m=100$  }  &  \makecell{ $1.67\cdot 10^{-165}$ \\ $1.67\cdot 10^{-165}$ \\ $1.67\cdot 10^{-165}$ } & \makecell{ 1246.9 \\ 1509.0 \\ 1796.2  }  & \makecell{ $1.67\cdot 10^{-165}$ \\ $1.67\cdot 10^{-165}$ \\ $1.67\cdot 10^{-165}$ }  & \makecell{ 192.9 \\ 233.6 \\ 278.5 } \\\hline
\end{tabularx}

SwinTransformer-V2-Tiny, no normalization, Gradient, Comparison with LRP-$\beta$

\begin{tabularx}{\textwidth}{XXXXXX}\hline\rule{0pt}{1.0\normalbaselineskip}
Augmentation & Sample size  & \multicolumn{2}{c}{ Grad vs LRP-$\beta=0$} & \multicolumn{2}{c}{ Grad vs LRP-$\beta=1$}  \\
             &          &  p-value & ratio & p-value & ratio   \\\hline \rule{0pt}{2\normalbaselineskip}
Gaussian & \makecell[l]{ $m=25$ \\ $m=50$ \\ $m=100$  } & \makecell{ $1.67\cdot 10^{-165}$ \\ $1.67\cdot 10^{-165}$ \\ $1.67\cdot 10^{-165}$  } & \makecell{ 14156.2 \\ 17324.9  \\ 21171.6 } & \makecell{ $1.67\cdot 10^{-165}$ \\$1.67\cdot 10^{-165}$   \\ $1.67\cdot 10^{-165}$ } & \makecell{ 1944.8 \\ 2361.4  \\ 2894.5 } \\
photometric & \makecell[l]{ $m=25$ \\ $m=50$ \\ $m=100$  }   & \makecell{ $1.67\cdot 10^{-165}$ \\$1.67\cdot 10^{-165}$ \\  $1.67\cdot 10^{-165}$ } &  \makecell{ 52690.9 \\ 64170.1 \\ 78109.1 }  & \makecell{ $1.67\cdot 10^{-165}$ \\ $1.67\cdot 10^{-165}$ \\$1.67\cdot 10^{-165}$ }  & \makecell{ 7225.7 \\ 8736.8 \\ 10678.0 } \\\hline
\end{tabularx}

Effnet-V2-S, no normalization, Gradient times input, Comparison with LRP-$\beta$

\begin{tabularx}{\textwidth}{XXXXXX}\hline\rule{0pt}{1.0\normalbaselineskip}
Augmentation & Sample size  & \multicolumn{2}{c}{ $\nabla \times x$ vs LRP-$\beta=0$} & \multicolumn{2}{c}{ $\nabla \times x$ vs LRP-$\beta=1$}  \\
             &          &  p-value & ratio & p-value & ratio   \\\hline \rule{0pt}{2\normalbaselineskip}
Gaussian & \makecell[l]{ $m=25$ \\ $m=50$ \\ $m=100$  } & \makecell{ $1.67\cdot 10^{-165}$ \\ $1.67\cdot 10^{-165}$ \\ $1.67\cdot 10^{-165}$ } & \makecell{1478.5 \\ 1460.6  \\ 1477.6  } & \makecell{$1.67\cdot 10^{-165}$ \\ $1.67\cdot 10^{-165}$ \\ $1.67\cdot 10^{-165}$ } & \makecell{ 451.6 \\ 447.0  \\ 450.2  } \\
photometric & \makecell[l]{ $m=25$ \\ $m=50$ \\ $m=100$  }  &  \makecell{ $1.67\cdot 10^{-165}$ \\ $1.67\cdot 10^{-165}$ \\ $1.67\cdot 10^{-165}$ } & \makecell{ 2883.2 \\ 2818.5 \\  2765.2 }  & \makecell{$1.67\cdot 10^{-165}$\\ $1.67\cdot 10^{-165}$ \\ $1.67\cdot 10^{-165}$ }  & \makecell{ 880.6 \\ 862.6 \\ 842.5 } \\\hline
\end{tabularx}

ResNet-50, no normalization, Gradient times input, Comparison with LRP-$\beta$

\begin{tabularx}{\textwidth}{XXXXXX}\hline\rule{0pt}{1.0\normalbaselineskip}
Augmentation & Sample size  & \multicolumn{2}{c}{ $\nabla \times x$ vs LRP-$\beta=0$} & \multicolumn{2}{c}{ $\nabla \times x$ vs LRP-$\beta=1$}  \\
             &          &  p-value & ratio & p-value & ratio   \\\hline \rule{0pt}{2\normalbaselineskip}
Gaussian & \makecell[l]{ $m=25$ \\ $m=50$ \\ $m=100$  } & \makecell{ $1.67\cdot 10^{-165}$ \\ $1.67\cdot 10^{-165}$ \\ $1.67\cdot 10^{-165}$  } & \makecell{ 336.9 \\ 338.1 \\ 337.8 } & \makecell{ $1.78\cdot 10^{-163}$ \\ $4.22\cdot 10^{-161}$ \\ $1.16\cdot 10^{-163}$  } & \makecell{ 52.1 \\ 52.4 \\ 52.4 } \\
photometric & \makecell[l]{ $m=25$ \\ $m=50$ \\ $m=100$  }   & \makecell{ $1.67\cdot 10^{-165}$ \\ $1.67\cdot 10^{-165}$ \\ $1.67\cdot 10^{-165}$  }  &  \makecell{ 819.2 \\ 835.7 \\ 834.5 }  & \makecell{ $1.77\cdot 10^{-165}$  \\ $1.67\cdot 10^{-165}$  \\  $1.69\cdot 10^{-165}$ } & \makecell{ 126.8 \\ 129.4 \\ 129.4  }  \\\hline
\end{tabularx}

SwinTransformer-V2-Tiny, no normalization, Gradient times input, Comparison with LRP-$\beta$

\begin{tabularx}{\textwidth}{XXXXXX}\hline\rule{0pt}{1.0\normalbaselineskip}
Augmentation & Sample size  & \multicolumn{2}{c}{ $\nabla \times x$ vs LRP-$\beta=0$} & \multicolumn{2}{c}{ $\nabla \times x$ vs LRP-$\beta=1$}  \\
             &          &  p-value & ratio & p-value & ratio   \\\hline \rule{0pt}{2\normalbaselineskip}
Gaussian & \makecell[l]{ $m=25$ \\ $m=50$ \\ $m=100$  } & \makecell{ $1.67\cdot 10^{-165}$ \\$1.67\cdot 10^{-165}$ \\ $1.67\cdot 10^{-165}$ } & \makecell{ 11461.4 \\ 11556.5 \\ 11507.5 } & \makecell{ $1.67\cdot 10^{-165}$\\ $1.67\cdot 10^{-165}$ \\$1.67\cdot 10^{-165}$  } & \makecell{ 1574.6  \\ 1575.2   \\ 1573.3 }  \\
photometric & \makecell[l]{ $m=25$ \\ $m=50$ \\ $m=100$  }  &  \makecell{ $1.67\cdot 10^{-165}$ \\  $1.67\cdot 10^{-165}$\\ $1.67\cdot 10^{-165}$ } & \makecell{ 37988.7 \\  38074.8 \\ 37776.8  }  & \makecell{ $1.67\cdot 10^{-165}$ \\ $1.67\cdot 10^{-165}$ \\ $1.67\cdot 10^{-165}$ }  & \makecell{ 5209.5 \\ 5183.9  \\ 5164.3 } \\\hline
\end{tabularx}

Effnet-V2-S, no normalization, Gradient, Comparison with LRP-$\gamma$

\begin{tabularx}{\textwidth}{XXXXXX}\hline\rule{0pt}{1.0\normalbaselineskip}
Augmentation & Sample size  & \multicolumn{2}{c}{ Grad vs LRP-$\gamma=10^3$} & \multicolumn{2}{c}{ Grad vs LRP-$\gamma=10^2$}  \\
             &          &  p-value & ratio & p-value & ratio   \\\hline \rule{0pt}{2\normalbaselineskip}
Gaussian & \makecell[l]{ $m=25$ \\ $m=50$ \\ $m=100$  } & \makecell{ $1.67\cdot 10^{-165}$ \\ $1.67\cdot 10^{-165}$ \\ $3.2\cdot 10^{-164}$  }& \makecell{ 2217.5 \\  2667.8 \\  3291.4 } & \makecell{ $1.61\cdot 10^{-151}$ \\ $1.64\cdot 10^{-149}$ \\ $3.82\cdot 10^{-142}$  }   & \makecell{ 1490.0 \\ 1770.3  \\ 2088.6 } \\
photometric & \makecell[l]{ $m=25$ \\ $m=50$ \\ $m=100$  }  &  \makecell{  $1.67\cdot 10^{-165}$ \\  $1.67\cdot 10^{-165}$ \\ $2.48\cdot 10^{-165}$  } & \makecell{ 5010.2 \\  5910.8 \\  7092.8 }  & \makecell{ $2.57\cdot 10^{-155}$  \\  $1.25\cdot 10^{-156}$ \\ $1.02\cdot 10^{-154}$ }  & \makecell{ 3366.5  \\ 3922.3 \\ 4500.8 } \\\hline
\end{tabularx}

ResNet-50, no normalization, Gradient, Comparison with LRP-$\gamma$

\begin{tabularx}{\textwidth}{XXXXXX}\hline\rule{0pt}{1.0\normalbaselineskip}
Augmentation & Sample size  & \multicolumn{2}{c}{ Grad vs LRP-$\gamma=10^3$} & \multicolumn{2}{c}{ Grad vs LRP-$\gamma=10^2$}  \\
             &          &  p-value & ratio & p-value & ratio   \\\hline \rule{0pt}{2\normalbaselineskip}
Gaussian & \makecell[l]{ $m=25$ \\ $m=50$ \\ $m=100$  } & \makecell{ $2.18\cdot 10^{-160}$ \\ $3.43\cdot 10^{-164}$ \\ $7.04\cdot 10^{-162}$  } & \makecell{ 437.8 \\ 519.5  \\ 576.8  } & \makecell{ $1.64\cdot 10^{-149}$ \\ $2.85\cdot 10^{-138}$ \\  $4.6\cdot 10^{-127}$ } & \makecell{ 422.8 \\ 453.2 \\ 358.5  }  \\
photometric & \makecell[l]{ $m=25$ \\ $m=50$ \\ $m=100$  }  & \makecell{ $ 6.6\cdot 10^{-163}$  \\ $2.32\cdot 10^{-165}$ \\ $1.67\cdot 10^{-165}$   } &  \makecell{ 1221.0 \\ 1252.7 \\ 1581.0}   & \makecell{ $2.02\cdot 10^{-158}$   \\ $2.83\cdot 10^{-152}$   \\ $4.68\cdot 10^{-146}$ }  & \makecell{ 1179.3 \\ 1252.7 \\ 982.6 } \\\hline
\end{tabularx}

Swin-V2-Tiny, no normalization, Gradient, Comparison with LRP-$\gamma$

\begin{tabularx}{\textwidth}{XXXXXX}\hline\rule{0pt}{1.0\normalbaselineskip}
Augmentation & Sample size  & \multicolumn{2}{c}{ Grad vs LRP-$\gamma=10^3$} & \multicolumn{2}{c}{ Grad vs LRP-$\gamma=10^2$}  \\
             &          &  p-value & ratio & p-value & ratio   \\\hline \rule{0pt}{2\normalbaselineskip}
Gaussian & \makecell[l]{ $m=25$ \\ $m=50$ \\ $m=100$  } & \makecell{ $1.67\cdot 10^{-165}$  \\ $1.67\cdot 10^{-165}$  \\ $1.67\cdot 10^{-165}$  } & \makecell{ 1235.3 \\ 1230.8  \\ 1329.6 } & \makecell{ $1.67\cdot 10^{-165}$  \\ $1.67\cdot 10^{-165}$  \\ $1.67\cdot 10^{-165}$  } & \makecell{ 850.6 \\ 799.5 \\ 840.7 } \\
photometric & \makecell[l]{ $m=25$ \\ $m=50$ \\ $m=100$  } &  \makecell{ $1.67\cdot 10^{-165}$ \\ $1.67\cdot 10^{-165}$ \\ $1.67\cdot 10^{-165}$ } & \makecell{ 4611.5 \\ 4597.8 \\ 4959.6 }  & \makecell{ $1.67\cdot 10^{-165}$ \\ $1.67\cdot 10^{-165}$ \\ $3.33\cdot 10^{-164}$ }  & \makecell{ 2932.0  \\   3152.7 \\ 3179.3 } \\\hline
\end{tabularx}

Effnet-V2-S, no normalization, Gradient times input, Comparison with LRP-$\gamma$

\begin{tabularx}{\textwidth}{XXXXXX}\hline\rule{0pt}{1.0\normalbaselineskip}
Augmentation & Sample size  & \multicolumn{2}{c}{ $\nabla \times x$ vs LRP-$\gamma=10^3$} & \multicolumn{2}{c}{ $\nabla \times x$ vs LRP-$\gamma=10^2$}  \\
             &          &  p-value & ratio & p-value & ratio   \\\hline \rule{0pt}{2\normalbaselineskip}
Gaussian & \makecell[l]{ $m=25$ \\ $m=50$ \\ $m=100$  } & \makecell{ $3.35\cdot 10^{-165}$ \\ $1.67\cdot 10^{-165}$ \\  $5.02\cdot 10^{-163}$ } & \makecell{ 1379.0 \\ 1362.5  \\  1366.2 } & \makecell{ $2.28\cdot 10^{-148}$ \\  $4.33\cdot 10^{-145}$ \\ $1.1\cdot 10^{-124}$ } & \makecell{ 926.6 \\ 904.1  \\ 867.0  } \\
photometric & \makecell[l]{ $m=25$ \\ $m=50$ \\ $m=100$  }  & \makecell{ $1.67\cdot 10^{-165}$  \\ $1.67\cdot 10^{-165}$ \\  $8.01\cdot 10^{-164}$  }   &  \makecell{ 2689.1 \\ 2629.1 \\ 2556.7  }  & \makecell{ $2.66\cdot 10^{-153}$  \\ $1.26\cdot 10^{-149}$ \\ $3.0\cdot 10^{-139}$  } & \makecell{ 1806.9 \\ 1744.6 \\ 1622.4 }  \\\hline
\end{tabularx}

ResNet-50, no normalization, Gradient times input, Comparison with LRP-$\gamma$

\begin{tabularx}{\textwidth}{XXXXXX}\hline\rule{0pt}{1.0\normalbaselineskip}
Augmentation & Sample size  & \multicolumn{2}{c}{ $\nabla \times x$ vs LRP-$\gamma=10^3$} & \multicolumn{2}{c}{ $\nabla \times x$ vs LRP-$\gamma=10^2$}  \\
             &          &  p-value & ratio & p-value & ratio   \\\hline \rule{0pt}{2\normalbaselineskip}
Gaussian & \makecell[l]{ $m=25$ \\ $m=50$ \\ $m=100$  } & \makecell{ $2.4\cdot 10^{-160}$ \\ $7.1\cdot 10^{-163}$ \\  $3.4\cdot 10^{-160}$ } & \makecell{329.9 \\ 321.8 \\ 297.3  } & \makecell{ $5.4\cdot 10^{-147}$  \\ $4.1\cdot 10^{-129}$  \\ $1.7\cdot 10^{-99}$  } & \makecell{  318.6 \\ 280.7 \\ 184.7 } \\
photometric & \makecell[l]{ $m=25$ \\ $m=50$ \\ $m=100$  }  &  \makecell{ $3.2\cdot 10^{-162}$  \\  $3.0\cdot 10^{-164}$ \\ $9.8\cdot 10^{-165}$  } & \makecell{ 802.2 \\ 795.3 \\ 734.5   }  & \makecell{ $1.8\cdot 10^{-156}$  \\ $3.8\cdot 10^{-145}$  \\ $1.6\cdot 10^{-132}$ }  & \makecell{ 774.8  \\ 693.8 \\ 456.5 } \\\hline
\end{tabularx}

Swin-V2-Tiny, no normalization, Gradient times input, Comparison with LRP-$\gamma$

\begin{tabularx}{\textwidth}{XXXXXX}\hline\rule{0pt}{1.0\normalbaselineskip}
Augmentation & Sample size  & \multicolumn{2}{c}{ $\nabla \times x$ vs LRP-$\gamma=10^3$} & \multicolumn{2}{c}{ $\nabla \times x$ vs LRP-$\gamma=10^2$}  \\
             &          &  p-value & ratio & p-value & ratio   \\\hline \rule{0pt}{2\normalbaselineskip}
Gaussian & \makecell[l]{ $m=25$ \\ $m=50$ \\ $m=100$  } & \makecell{ $1.67\cdot 10^{-165}$\\$1.67\cdot 10^{-165}$  \\ $1.67\cdot 10^{-165}$ } & \makecell{ 996.3 \\ 821.2 \\ 727.0 } & \makecell{ $1.67\cdot 10^{-165}$\\ $1.67\cdot 10^{-165}$ \\ $1.98\cdot 10^{-165}$ } & \makecell{ 686.0\\ 533.4 \\ 459.7 } \\
photometric & \makecell[l]{ $m=25$ \\ $m=50$ \\ $m=100$  } &  \makecell{ $1.67\cdot 10^{-165}$ \\ $1.67\cdot 10^{-165}$ \\ $1.67\cdot 10^{-165}$ } & \makecell{ 3382.1 \\ 2725.9 \\ 2362.5  }  & \makecell{ $1.67\cdot 10^{-165}$ \\$1.67\cdot 10^{-165}$  \\  $3.34\cdot 10^{-164}$}  & \makecell{ 2150.4 \\  1869.1 \\ 1514.5 } \\\hline
\end{tabularx}

\newpage
\subsection{$\ell_2$-normalized case, not covered by theoretical results}\label{ssec:l2norm}

\hspace{5mm}Effnet-V2-S, $\ell_2$-normalization, Gradient, Comparison with LRP-$\beta$

\begin{tabularx}{\textwidth}{XXXXXX}\hline\rule{0pt}{1.0\normalbaselineskip}
Augmentation & Sample size  & \multicolumn{2}{c}{ Grad vs LRP-$\beta=0$} & \multicolumn{2}{c}{ Grad vs LRP-$\beta=1$}  \\
             &          &  p-value & ratio & p-value & ratio   \\\hline \rule{0pt}{2\normalbaselineskip}
Gaussian & \makecell[l]{ $m=25$ \\ $m=50$ \\ $m=100$  } & \makecell{ $1.23\cdot 10^{-161}$  \\ $1.06\cdot 10^{-163}$ \\ $6.56\cdot 10^{-163}$  } & \makecell{ 3.3\\  4.0 \\ 4.9 } & \makecell{ $2.29\cdot 10^{-160}$ \\ $8.63\cdot 10^{-158}$ \\ $6.84\cdot 10^{-163}$ } & \makecell{  1.3\\  1.6 \\ 1.9  } \\
photometric & \makecell[l]{ $m=25$ \\ $m=50$ \\ $m=100$  } &  \makecell{ $1.29\cdot 10^{-161}$  \\ $3.22\cdot 10^{-160}$ \\ $6.58\cdot 10^{-163}$ } & \makecell{  2.8\\  3.4\\  4.0  }  & \makecell{ $1.48\cdot 10^{-84}$ \\ $7.46\cdot 10^{-154}$  \\ $4.36\cdot 10^{-159}$ }  & \makecell{  1.1\\ 1.3  \\ 1.5  } \\\hline
\end{tabularx}

ResNet-50, $\ell_2$-normalization, Gradient, Comparison with LRP-$\beta$

\begin{tabularx}{\textwidth}{XXXXXX}\hline\rule{0pt}{1.0\normalbaselineskip}
Augmentation & Sample size  & \multicolumn{2}{c}{ Grad vs LRP-$\beta=0$} & \multicolumn{2}{c}{ Grad vs LRP-$\beta=1$}  \\
             &          &  p-value & ratio & p-value & ratio   \\\hline \rule{0pt}{2\normalbaselineskip}
Gaussian & \makecell[l]{ $m=25$ \\ $m=50$ \\ $m=100$  } & \makecell{ $1.57\cdot 10^{-123}$  \\ $7.53\cdot 10^{-122}$ \\ $7.54\cdot 10^{-120}$ } & \makecell{ 2.2 \\ 2.6 \\  3.1 } & \makecell{ 1.0 \\ $6.13\cdot 10^{-42}$ \\ $1.27\cdot 10^{-94}$ } & \makecell{ 1.0 \\ 1.1  \\ 1.3 } \\
photometric & \makecell[l]{ $m=25$ \\ $m=50$ \\ $m=100$  } &  \makecell{ $3.60\cdot 10^{-123}$ \\ $3.94\cdot 10^{-120}$ \\ $2.95\cdot 10^{-118}$  } & \makecell{  2.1\\ 2.6 \\ 3.0  }  & \makecell{ 1.0 \\ $3.15\cdot 10^{-25}$ \\ $7.57\cdot 10^{-87}$ }  & \makecell{ 0.9 \\ 1.1 \\ 1.3 } \\\hline
\end{tabularx}

Swin-V2-Tiny, $\ell_2$-normalization, Gradient, Comparison with LRP-$\beta$

\begin{tabularx}{\textwidth}{XXXXXX}\hline\rule{0pt}{1.0\normalbaselineskip}
Augmentation & Sample size  & \multicolumn{2}{c}{ Grad  vs LRP-$\beta=0$} & \multicolumn{2}{c}{ Grad vs vs LRP-$\beta=1$}  \\
             &          &  p-value & ratio & p-value & ratio   \\\hline \rule{0pt}{2\normalbaselineskip}
Gaussian & \makecell[l]{ $m=25$ \\ $m=50$ \\ $m=100$  } & \makecell{  $2.01\cdot 10^{-156}$\\$1.74\cdot 10^{-153}$ \\ $8.11\cdot 10^{-153}$ } & \makecell{ 3.3\\ 4.10 \\5.0  } & \makecell{ $1.94\cdot 10^{-142}$\\$8.86\cdot 10^{-148}$  \\ $6.17\cdot 10^{-142}$ } & \makecell{ 1.5\\ 1.80 \\ 2.1 } \\
photometric & \makecell[l]{ $m=25$ \\ $m=50$ \\ $m=100$  } &  \makecell{ $1.99\cdot 10^{-152}$ \\ $6.24\cdot 10^{-153}$ \\ $1.08\cdot 10^{-150}$ } & \makecell{3.5  \\ 4.2\\ 5.1  }  & \makecell{ $2.75\cdot 10^{-145}$ \\$1.28\cdot 10^{-145}$  \\$2.36\cdot 10^{-145}$ }  & \makecell{ 1.5 \\ 1.8 \\ 2.2 } \\\hline
\end{tabularx}

Effnet-V2-S, $\ell_2$-normalization, Gradient times input, Comparison with LRP-$\beta$

\begin{tabularx}{\textwidth}{XXXXXX}\hline\rule{0pt}{1.0\normalbaselineskip}
Augmentation & Sample size  & \multicolumn{2}{c}{ $\nabla \times x$ vs LRP-$\beta=0$} & \multicolumn{2}{c}{ $\nabla \times x$ vs LRP-$\beta=1$}  \\
             &          &  p-value & ratio & p-value & ratio   \\\hline \rule{0pt}{2\normalbaselineskip}
Gaussian & \makecell[l]{ $m=25$ \\ $m=50$ \\ $m=100$  } & \makecell{ $1.69\cdot 10^{-165}$  \\ $1.69\cdot 10^{-165}$ \\ $1.68\cdot 10^{-165}$  } & \makecell{ 6.5\\  7.7 \\ 8.7 } & \makecell{ $3.04\cdot 10^{-165}$ \\ $1.87\cdot 10^{-165}$ \\ $1.20\cdot 10^{-164}$ } & \makecell{  2.6\\  3.0 \\ 3.3 } \\
photometric & \makecell[l]{ $m=25$ \\ $m=50$ \\ $m=100$  } &  \makecell{ $1.36\cdot 10^{-160}$  \\ $3.40\cdot 10^{-157}$ \\ $4.54\cdot 10^{-160}$ } & \makecell{  2.3\\  2.4\\  2.4  }  & \makecell{ $1$ \\ $1$  \\ $1$ }  & \makecell{  0.9\\ 0.9  \\ 0.9  } \\\hline
\end{tabularx}

ResNet-50, $\ell_2$-normalization, Gradient times input, Comparison with LRP-$\beta$

\begin{tabularx}{\textwidth}{XXXXXX}\hline\rule{0pt}{1.0\normalbaselineskip}
Augmentation & Sample size  & \multicolumn{2}{c}{ $\nabla \times x$ vs LRP-$\beta=0$} & \multicolumn{2}{c}{ $\nabla \times x$ vs LRP-$\beta=1$}  \\
             &          &  p-value & ratio & p-value & ratio   \\\hline \rule{0pt}{2\normalbaselineskip}
Gaussian & \makecell[l]{ $m=25$ \\ $m=50$ \\ $m=100$  } & \makecell{ $5.67\cdot 10^{-165}$  \\ $7.72\cdot 10^{-165}$ \\ $9.70\cdot 10^{-162}$  } & \makecell{ 5.4\\  6.8\\ 8.0 } & \makecell{ $1.37\cdot 10^{-164}$ \\ $3.16\cdot 10^{-164}$ \\ $1.46\cdot 10^{-158}$ } & \makecell{  2.4\\  2.9 \\ 3.4 } \\
photometric & \makecell[l]{ $m=25$ \\ $m=50$ \\ $m=100$  } &  \makecell{ $2.25\cdot 10^{-131}$  \\ $7.49\cdot 10^{-126}$ \\ $1.04\cdot 10^{-115}$ } & \makecell{  2.7\\  2.9\\  2.8  }  & \makecell{ $2.41\cdot 10^{-47}$ \\ $6.18\cdot 10^{-47}$  \\ $1.35\cdot 10^{-43}$ }  & \makecell{  1.2\\ 1.1  \\ 1.2  } \\\hline
\end{tabularx}

\vspace{40mm}

Swin-V2-Tiny, $\ell_2$-normalization, Gradient times input, Comparison with LRP-$\beta$

\begin{tabularx}{\textwidth}{XXXXXX}\hline\rule{0pt}{1.0\normalbaselineskip}
Augmentation & Sample size  & \multicolumn{2}{c}{ $\nabla \times x$ vs LRP-$\beta=0$} & \multicolumn{2}{c}{ $\nabla \times x$ vs LRP-$\beta=1$}  \\
             &          &  p-value & ratio & p-value & ratio   \\\hline \rule{0pt}{2\normalbaselineskip}
Gaussian & \makecell[l]{ $m=25$ \\ $m=50$ \\ $m=100$  } & \makecell{ $1.98\cdot 10^{-165}$  \\ $6.76\cdot 10^{-163}$ \\ $9.07\cdot 10^{-158}$  } & \makecell{ 6.5\\  7.4\\ 8.0 } & \makecell{ $3.85\cdot 10^{-165}$ \\ $4.01\cdot 10^{-163}$ \\ $3.50\cdot 10^{-152}$ } & \makecell{  2.9\\  3.2 \\ 3.5 } \\
photometric & \makecell[l]{ $m=25$ \\ $m=50$ \\ $m=100$  } &  \makecell{ $7.17\cdot 10^{-151}$  \\ $5.21\cdot 10^{-148}$ \\ $6.49\cdot 10^{-145}$ } & \makecell{  3.0\\  3.1\\  3.1  }  & \makecell{ $1.90\cdot 10^{-82}$ \\ $9.78\cdot 10^{-81}$  \\ $1.80\cdot 10^{-80}$ }  & \makecell{  1.3\\ 1.3  \\ 1.4  } \\\hline
\end{tabularx}

Effnet-V2-S, $\ell_2$-normalization, Gradient, Comparison with LRP-$\gamma$

\begin{tabularx}{\textwidth}{XXXXXX}\hline\rule{0pt}{1.0\normalbaselineskip}
Augmentation & Sample size  & \multicolumn{2}{c}{ Grad  vs LRP-$\gamma=10^{3}$} & \multicolumn{2}{c}{ Grad vs vs LRP-$\gamma=10^{2}$}  \\
             &          &  p-value & ratio & p-value & ratio   \\\hline \rule{0pt}{2\normalbaselineskip}
Gaussian & \makecell[l]{ $m=25$ \\ $m=50$ \\ $m=100$  } & \makecell{ $8\cdot 10^{-153}$  \\ $3\cdot 10^{-149}$ \\ $4\cdot 10^{-148}$  } & \makecell{ 3.3 \\ 4.0 \\ 4.9 } & \makecell{ $2\cdot 10^{-84}$ \\ $3\cdot 10^{-71}$ \\ $4\cdot 10^{-13}$ } & \makecell{ 3.2 \\ 3.9  \\ 4.7 } \\
photometric & \makecell[l]{ $m=25$ \\ $m=50$ \\ $m=100$  } &  \makecell{ $9\cdot 10^{-153}$  \\ $4\cdot 10^{-145}$ \\ $8\cdot 10^{-148}$ } & \makecell{ 2.8 \\ 3.4 \\ 4.0   }  & \makecell{ $9\cdot 10^{-82}$ \\ $1\cdot 10^{-6}$  \\ $1\cdot 10^{-10}$ }  & \makecell{ 2.8 \\ 3.3  \\ 3.9  } \\\hline
\end{tabularx}

ResNet-50, $\ell_2$-normalization, Gradient, Comparison with LRP-$\gamma$

\begin{tabularx}{\textwidth}{XXXXXX}\hline\rule{0pt}{1.0\normalbaselineskip}
Augmentation & Sample size  & \multicolumn{2}{c}{ Grad vs LRP-$\gamma=10^{3}$} & \multicolumn{2}{c}{ Grad  vs LRP-$\gamma=10^{2}$}  \\
             &          &  p-value & ratio & p-value & ratio   \\\hline \rule{0pt}{2\normalbaselineskip}
Gaussian & \makecell[l]{ $m=25$ \\ $m=50$ \\ $m=100$  }  & \makecell{ $7\cdot 10^{-83}$ \\ $4\cdot 10^{-60}$ \\ $1\cdot 10^{-35}$ } & \makecell{ 2.2 \\ 2.5  \\ 2.9 } & \makecell{  $7\cdot 10^{-33}$ \\ $3\cdot 10^{-5}$ \\ $1.0$  } & \makecell{2.1  \\ 2.4 \\ 2.1 } \\
photometric & \makecell[l]{ $m=25$ \\ $m=50$ \\ $m=100$  } &  \makecell{ $2\cdot 10^{-82}$  \\ $5\cdot 10^{-58}$ \\ $1\cdot 10^{-32}$ } & \makecell{ 2.1 \\ 2.5 \\ 2.8   }  & \makecell{ $2\cdot 10^{-32}$ \\ $7\cdot 10^{-5}$  \\ $1.0$ }  & \makecell{ 2.1 \\ 2.4  \\ 2.0  } \\\hline
\end{tabularx}

Swin-V2-Tiny, $\ell_2$-normalization, Gradient, Comparison with LRP-$\gamma$

\begin{tabularx}{\textwidth}{XXXXXX}\hline\rule{0pt}{1.0\normalbaselineskip}
Augmentation & Sample size  & \multicolumn{2}{c}{ Grad vs vs LRP-$\gamma=10^{3}$} & \multicolumn{2}{c}{ Grad vs vs LRP-$\gamma=10^{2}$}  \\
             &          &  p-value & ratio & p-value & ratio   \\\hline \rule{0pt}{2\normalbaselineskip}
Gaussian & \makecell[l]{ $m=25$ \\ $m=50$ \\ $m=100$  } & \makecell{ $1.0$  \\ $1.0$ \\  $1.0$  } & \makecell{ 0.4\\ 0.4 \\ 0.3 } & \makecell{  $1.0$ \\  $1.0$ \\ $1.0$ } & \makecell{ 0.4 \\ 0.3  \\ 0.3 } \\
photometric & \makecell[l]{ $m=25$ \\ $m=50$ \\ $m=100$  } &  \makecell{  $1.0$ \\$1.0$\\ $1.0$  } & \makecell{ 0.4 \\ 0.4 \\  0.4  }  & \makecell{ $1.0$ \\ $1.0$  \\ $1.0$ }  & \makecell{ 0.4 \\ 0.4  \\ 0.3  } \\\hline
\end{tabularx}

Effnet-V2-S, $\ell_2$-normalization, Gradient times input, Comparison with LRP-$\gamma$

\begin{tabularx}{\textwidth}{XXXXXX}\hline\rule{0pt}{1.0\normalbaselineskip}
Augmentation & Sample size  & \multicolumn{2}{c}{ $\nabla \times x$ vs LRP-$\gamma=10^{3}$} & \multicolumn{2}{c}{ Grad vs vs LRP-$\gamma=10^{2}$}  \\
             &          &  p-value & ratio & p-value & ratio   \\\hline \rule{0pt}{2\normalbaselineskip}
Gaussian & \makecell[l]{ $m=25$ \\ $m=50$ \\ $m=100$  } & \makecell{ 2$\cdot 10^{-165}$  \\3$\cdot 10^ {-165}$ \\ 2$\cdot 10^{-160}$  } & \makecell{ 6.5 \\ 7.7 \\ 8.7 } & \makecell{ 1$\cdot 10^{-161}$ \\ 2$\cdot 10^{-158}$ \\ 8$\cdot 10^{-88}$ } & \makecell{ 6.4 \\ 7.6  \\ 8.4 } \\
photometric & \makecell[l]{ $m=25$ \\ $m=50$ \\ $m=100$  } &  \makecell{ 9$\cdot 10^{-152}$  \\ $6\cdot 10^{-141}$ \\ $5 \cdot 10^{-140}$ } & \makecell{ 2.3 \\ 2.4 \\ 2.4   }  & \makecell{ $1\cdot 10^{-80}$ \\ 9$\cdot 10^{-59}$  \\ 4$\cdot 10^{-6}$ }  & \makecell{ 2.6 \\2.4   \\ 2.3  } \\\hline
\end{tabularx}

ResNet-50, $\ell_2$-normalization, Gradient times input, Comparison with LRP-$\gamma$

\begin{tabularx}{\textwidth}{XXXXXX}\hline\rule{0pt}{1.0\normalbaselineskip}
Augmentation & Sample size  & \multicolumn{2}{c}{ $\nabla \times x$ vs LRP-$\gamma=10^{3}$} & \multicolumn{2}{c}{ Grad vs vs LRP-$\gamma=10^{2}$}  \\
             &          &  p-value & ratio & p-value & ratio   \\\hline \rule{0pt}{2\normalbaselineskip}
Gaussian & \makecell[l]{ $m=25$ \\ $m=50$ \\ $m=100$  } & \makecell{ 1$\cdot 10^{-163}$  \\ 6$\cdot 10^{-161}$ \\8 $\cdot 10^{-149}$  } & \makecell{ 5.3\\ 6.6 \\ 7.5 } & \makecell{ 1$\cdot 10^{-155}$ \\7 $\cdot 10^{-136}$ \\1 $\cdot 10^{-78}$ } & \makecell{ 5.2 \\ 6.3  \\ 7.5 } \\
photometric & \makecell[l]{ $m=25$ \\ $m=50$ \\ $m=100$  } &  \makecell{9 $\cdot 10^{-99}$  \\ $1\cdot 10^{-66}$ \\ $5\cdot 10^{-30}$ } & \makecell{ 2.7 \\ 2.8 \\ 2.7   }  & \makecell{ $3\cdot 10^{-45}$ \\ $1\cdot 10^{-7}$  \\ $1.0$ }  & \makecell{ 2.6 \\ 2.6  \\ 1.9   } \\\hline
\end{tabularx}

Swin-V2-Tiny, $\ell_2$-normalization, Gradient times input, Comparison with LRP-$\gamma$

\begin{tabularx}{\textwidth}{XXXXXX}\hline\rule{0pt}{1.0\normalbaselineskip}
Augmentation & Sample size  & \multicolumn{2}{c}{ $\nabla \times x$ vs LRP-$\gamma=10^{3}$} & \multicolumn{2}{c}{ $\nabla \times x$ vs LRP-$\gamma=10^{2}$}  \\
             &          &  p-value & ratio & p-value & ratio   \\\hline \rule{0pt}{2\normalbaselineskip}
Gaussian & \makecell[l]{ $m=25$ \\ $m=50$ \\ $m=100$  } & \makecell{ $1.0$  \\ $1.0$ \\ $1.0$  } & \makecell{ 0.8\\ 0.7 \\ 0.6 } & \makecell{ $1.0$ \\ $1.0$ \\ $1.0$ } & \makecell{ 0.8 \\ 0.6  \\ 0.5 } \\
photometric & \makecell[l]{ $m=25$ \\ $m=50$ \\ $m=100$  } &  \makecell{ $1.0$ \\$1.0$ \\ $1.0$ } & \makecell{ 0.4  \\ 0.3 \\ 0.2   }  & \makecell{ $1.0$ \\ $1.0$  \\$1.0$ }  & \makecell{  0.3  \\ 0.3 \\ 0.2  } \\\hline
\end{tabularx}

\vspace{2mm}

More detailed box plots for the gradient versus LRP-$\beta$ can be inspected in Section \ref{app:stats_grad}. Box plots for gradient times input versus LRP-$\beta$ are shown in Section \ref{app:stats_gxi}. The box plots also contain information about inter-quartile ranges as a replacement for the variance, as these statistics of non-negative values do not fit well Normal distribution assumptions.

For LRP-$\gamma$ we show boxplots for the gradient in the Appendix Section \ref{app:stats_grad_gamma}. We see in Appendix Section \ref{app:stats_grad_gamma} usually a faster convergence for the unnormalized version, as predicted according to lemma \ref{lemma:seqbound_lrpgamma}. For the $\ell_2$-normalized variant, which is not covered by this lemma, we can observe that $\gamma=100.0$ is often a too small choice. This is apparent in cases seen in Section \ref{app:stats_grad_gamma} where the mean for $\gamma=100.0$, shown as green triangle, is higher than the median, shown by a horizontal vertical line. This discrepancy between the mean and the median indicates a presence of outlier samples with large distances implying slower convergence. 


While lemma \ref{lemma:seqbound_lrpgamma} does not make a prediction for this $\ell_2$-normalized case, the condition to $\gamma$ required in lemma \ref{lemma:seqbound_lrpgamma} might be useful in general for determining satisfactory ranges for the parameter $\gamma$.

As an outlook, this may indicate the possibility of optimizing parameters for LRP attribution maps beyond faithfulness measures.

\section{Conclusion}\label{sec:conclusion}






We have analyzed the convergence properties of averaged attribution maps—a framework relevant for predictions using multiple photometric augmentations and for noise-augmented prediction methods like SmoothGrad and SmoothLRP. Our theoretical analysis establishes a weight-independent upper bound for  LRP-$\beta$ and demonstrates that LRP-$\gamma$'s convergence can be decoupled from weights when $\gamma$ satisfies conditions related to the relative scales of positive and negative activations.

Experimentally, we quantified lower bounds of convergence, revealing that  LRP-$\beta$ converges notably faster than gradient-based methods. This advantage persists even after $\ell_2$-normalization, suggesting additional beneficial convergence properties yet to be theoretically characterized. 

Regarding limitations, we explicitly do not aim at a discussion on which attribution method can be considered superior in general, acknowledging that selection depends on specific requirements of the use case and involves trade-offs between different evaluation measures. We also do not aim at the question which set of criteria should be used to select an attribution method. This would require a much broader discussion of evaluation measures beyond the scope of averages of attribution maps considered here.

In practice, faster convergence translates to computational efficiency through reduced sampling requirements.






\newpage


\bibliography{neurips2025}

\begin{thebibliography}{10}

\bibitem{DBLP:journals/corr/SimonyanVZ13}
K.~Simonyan, A.~Vedaldi, and A.~Zisserman, ``Deep inside convolutional
  networks: Visualising image classification models and saliency maps,'' in
  {\em 2nd International Conference on Learning Representations, {ICLR} 2014,
  Banff, AB, Canada, April 14-16, 2014, Workshop Track Proceedings} (Y.~Bengio
  and Y.~LeCun, eds.), 2014.

\bibitem{Springenberg}
J.~Springenberg, A.~Dosovitskiy, T.~Brox, and M.~Riedmiller, ``Striving for
  simplicity: The all convolutional net,'' in {\em ICLR (workshop track)},
  2015.

\bibitem{gradcam}
R.~R. Selvaraju, M.~Cogswell, A.~Das, R.~Vedantam, D.~Parikh, and D.~Batra,
  ``Grad-cam: Visual explanations from deep networks via gradient-based
  localization,'' in {\em 2017 IEEE International Conference on Computer Vision
  (ICCV)}, pp.~618--626, 2017.

\bibitem{DBLP:conf/icml/ShrikumarGK17}
A.~Shrikumar, P.~Greenside, and A.~Kundaje, ``Learning important features
  through propagating activation differences,'' in {\em Proceedings of the 34th
  International Conference on Machine Learning, {ICML} 2017, Sydney, NSW,
  Australia, 6-11 August 2017} (D.~Precup and Y.~W. Teh, eds.), vol.~70 of {\em
  Proceedings of Machine Learning Research}, pp.~3145--3153, {PMLR}, 2017.

\bibitem{DBLP:conf/icml/BalduzziFLLMM17}
D.~Balduzzi, M.~Frean, L.~Leary, J.~P. Lewis, K.~W. Ma, and B.~McWilliams,
  ``The shattered gradients problem: If resnets are the answer, then what is
  the question?,'' in {\em International Conference on Machine Learning
  (ICML)}, vol.~70 of {\em PMLR}, pp.~342--350, {PMLR}, 2017.

\bibitem{Bach_15}
S.~Bach, A.~Binder, G.~Montavon, F.~Klauschen, K.~R. Müller, and W.~Samek,
  ``On pixel-wise explanations for non-linear classifier decisions by
  layer-wise relevance propagation,'' {\em PLoS ONE}, vol.~10, no.~7,
  pp.~1--46, 2015.

\bibitem{Lapuschkin2019}
S.~Lapuschkin, S.~W{\"a}ldchen, A.~Binder, G.~Montavon, W.~Samek, and K.-R.
  M{\"u}ller, ``Unmasking clever hans predictors and assessing what machines
  really learn,'' {\em Nature Communications}, vol.~10, p.~1096, Mar 2019.

\bibitem{Binder_2023_CVPR}
A.~Binder, L.~Weber, S.~Lapuschkin, G.~Montavon, K.-R. M\"uller, and W.~Samek,
  ``Shortcomings of top-down randomization-based sanity checks for evaluations
  of deep neural network explanations,'' in {\em Proceedings of the IEEE/CVF
  Conference on Computer Vision and Pattern Recognition (CVPR)},
  pp.~16143--16152, June 2023.

\bibitem{pmlr-v235-achtibat24a}
R.~Achtibat, S.~M.~V. Hatefi, M.~Dreyer, A.~Jain, T.~Wiegand, S.~Lapuschkin,
  and W.~Samek, ``{A}ttn{LRP}: Attention-aware layer-wise relevance propagation
  for transformers,'' in {\em Proceedings of the 41st International Conference
  on Machine Learning} (R.~Salakhutdinov, Z.~Kolter, K.~Heller, A.~Weller,
  N.~Oliver, J.~Scarlett, and F.~Berkenkamp, eds.), vol.~235 of {\em
  Proceedings of Machine Learning Research}, pp.~135--168, PMLR, 21--27 Jul
  2024.

\bibitem{NEURIPS2024_d6d0e41e}
F.~Rezaei~Jafari, G.~Montavon, K.-R. M\"{u}ller, and O.~Eberle, ``Mambalrp:
  Explaining selective state space sequence models,'' in {\em Advances in
  Neural Information Processing Systems} (A.~Globerson, L.~Mackey, D.~Belgrave,
  A.~Fan, U.~Paquet, J.~Tomczak, and C.~Zhang, eds.), vol.~37,
  pp.~118540--118570, Curran Associates, Inc., 2024.

\bibitem{DBLP:conf/nips/AdebayoGMGHK18}
J.~Adebayo, J.~Gilmer, M.~Muelly, I.~J. Goodfellow, M.~Hardt, and B.~Kim,
  ``Sanity checks for saliency maps,'' in {\em Advances in Neural Information
  Processing Systems 31}, pp.~9525--9536, 2018.

\bibitem{DBLP:conf/iclr/AnconaCO018}
M.~Ancona, E.~Ceolini, C.~{\"{O}}ztireli, and M.~Gross, ``Towards better
  understanding of gradient-based attribution methods for deep neural
  networks,'' in {\em 6th International Conference on Learning Representations,
  {ICLR} 2018, Vancouver, BC, Canada, April 30 - May 3, 2018, Conference Track
  Proceedings}, OpenReview.net, 2018.

\bibitem{DBLP:conf/icml/SundararajanTY17}
M.~Sundararajan, A.~Taly, and Q.~Yan, ``Axiomatic attribution for deep
  networks,'' in {\em Proceedings of the 34th International Conference on
  Machine Learning, {ICML} 2017, Sydney, NSW, Australia, 6-11 August 2017}
  (D.~Precup and Y.~W. Teh, eds.), vol.~70 of {\em Proceedings of Machine
  Learning Research}, pp.~3319--3328, {PMLR}, 2017.

\bibitem{DBLP:journals/corr/SmilkovTKVW17}
D.~Smilkov, N.~Thorat, B.~Kim, F.~B. Vi{\'{e}}gas, and M.~Wattenberg,
  ``Smoothgrad: removing noise by adding noise,'' {\em arXiv preprint
  arXiv:1706.03825}, 2017.

\bibitem{DBLP:conf/esann/RaulfDHMF21}
A.~P. Raulf, S.~D{\"{a}}ubener, B.~Hack, A.~Mosig, and A.~Fischer, ``Smoothlrp:
  Smoothing {LRP} by averaging over stochastic input variations,'' in {\em 29th
  European Symposium on Artificial Neural Networks, Computational Intelligence
  and Machine Learning, {ESANN} 2021, Online event (Bruges, Belgium), October
  6-8, 2021}, 2021.

\bibitem{strumbelj2014}
E.~{\v{S}}trumbelj and I.~Kononenko, ``Explaining prediction models and
  individual predictions with feature contributions,'' {\em Knowledge and
  Information Systems}, vol.~41, pp.~647--665, Dec 2014.

\bibitem{DBLP:conf/nips/LundbergL17}
S.~M. Lundberg and S.~Lee, ``A unified approach to interpreting model
  predictions,'' in {\em Advances in Neural Information Processing Systems 30:
  Annual Conference on Neural Information Processing Systems 2017, December
  4-9, 2017, Long Beach, CA, {USA}} (I.~Guyon, U.~von Luxburg, S.~Bengio, H.~M.
  Wallach, R.~Fergus, S.~V.~N. Vishwanathan, and R.~Garnett, eds.),
  pp.~4765--4774, 2017.

\bibitem{DBLP:conf/bmvc/PetsiukDS18}
V.~Petsiuk, A.~Das, and K.~Saenko, ``{RISE:} randomized input sampling for
  explanation of black-box models,'' in {\em British Machine Vision Conference
  2018, {BMVC} 2018, Newcastle, UK, September 3-6, 2018}, p.~151, {BMVA} Press,
  2018.

\bibitem{DBLP:conf/iccv/FongV17}
R.~C. Fong and A.~Vedaldi, ``Interpretable explanations of black boxes by
  meaningful perturbation,'' in {\em {IEEE} International Conference on
  Computer Vision, {ICCV} 2017, Venice, Italy, October 22-29, 2017},
  pp.~3449--3457, {IEEE} Computer Society, 2017.

\bibitem{DBLP:conf/accv/AgarwalN20}
C.~Agarwal and A.~Nguyen, ``Explaining image classifiers by removing input
  features using generative models,'' in {\em Computer Vision - {ACCV} 2020 -
  15th Asian Conference on Computer Vision, Kyoto, Japan, November 30 -
  December 4, 2020, Revised Selected Papers, Part {VI}} (H.~Ishikawa, C.~Liu,
  T.~Pajdla, and J.~Shi, eds.), vol.~12627 of {\em Lecture Notes in Computer
  Science}, pp.~101--118, Springer, 2020.

\bibitem{DBLP:conf/nips/Alvarez-MelisJ18}
D.~Alvarez{-}Melis and T.~S. Jaakkola, ``Towards robust interpretability with
  self-explaining neural networks,'' in {\em Advances in Neural Information
  Processing Systems 31: Annual Conference on Neural Information Processing
  Systems 2018, NeurIPS 2018, December 3-8, 2018, Montr{\'{e}}al, Canada}
  (S.~Bengio, H.~M. Wallach, H.~Larochelle, K.~Grauman, N.~Cesa{-}Bianchi, and
  R.~Garnett, eds.), pp.~7786--7795, 2018.

\bibitem{DBLP:journals/tnn/SamekBMLM17}
W.~Samek, A.~Binder, G.~Montavon, S.~Lapuschkin, and K.~M{\"{u}}ller,
  ``Evaluating the visualization of what a deep neural network has learned,''
  {\em {IEEE} Trans. Neural Networks Learn. Syst.}, vol.~28, no.~11,
  pp.~2660--2673, 2017.

\bibitem{DBLP:conf/nips/YehHSIR19}
C.~Yeh, C.~Hsieh, A.~S. Suggala, D.~I. Inouye, and P.~Ravikumar, ``On the
  (in)fidelity and sensitivity of explanations,'' in {\em Advances in Neural
  Information Processing Systems 32: Annual Conference on Neural Information
  Processing Systems 2019, NeurIPS 2019, December 8-14, 2019, Vancouver, BC,
  Canada} (H.~M. Wallach, H.~Larochelle, A.~Beygelzimer,
  F.~d'Alch{\'{e}}{-}Buc, E.~B. Fox, and R.~Garnett, eds.), pp.~10965--10976,
  2019.

\bibitem{DBLP:conf/ijcai/BhattWM20}
U.~Bhatt, A.~Weller, and J.~M.~F. Moura, ``Evaluating and aggregating
  feature-based model explanations,'' in {\em Proceedings of the Twenty-Ninth
  International Joint Conference on Artificial Intelligence, {IJCAI} 2020}
  (C.~Bessiere, ed.), pp.~3016--3022, ijcai.org, 2020.

\bibitem{DBLP:journals/corr/abs-2003-08747}
L.~Rieger and L.~K. Hansen, ``{IROF:} a low resource evaluation metric for
  explanation methods,'' {\em CoRR}, vol.~abs/2003.08747, 2020.

\bibitem{DBLP:journals/corr/abs-2007-07584}
A.~Nguyen and M.~R. Mart{\'{\i}}nez, ``On quantitative aspects of model
  interpretability,'' {\em CoRR}, vol.~abs/2007.07584, 2020.

\bibitem{DBLP:conf/icml/DasguptaFM22}
S.~Dasgupta, N.~Frost, and M.~Moshkovitz, ``Framework for evaluating
  faithfulness of local explanations,'' in {\em International Conference on
  Machine Learning, {ICML} 2022, 17-23 July 2022, Baltimore, Maryland, {USA}}
  (K.~Chaudhuri, S.~Jegelka, L.~Song, C.~Szepesv{\'{a}}ri, G.~Niu, and
  S.~Sabato, eds.), vol.~162 of {\em Proceedings of Machine Learning Research},
  pp.~4794--4815, {PMLR}, 2022.

\bibitem{DBLP:journals/corr/abs-2203-06877}
C.~Agarwal, N.~Johnson, M.~Pawelczyk, S.~Krishna, E.~Saxena, M.~Zitnik, and
  H.~Lakkaraju, ``Rethinking stability for attribution-based explanations,''
  {\em CoRR}, vol.~abs/2203.06877, 2022.

\bibitem{DBLP:conf/icml/RongLBKK22}
Y.~Rong, T.~Leemann, V.~Borisov, G.~Kasneci, and E.~Kasneci, ``A consistent and
  efficient evaluation strategy for attribution methods,'' in {\em
  International Conference on Machine Learning, {ICML} 2022, 17-23 July 2022,
  Baltimore, Maryland, {USA}} (K.~Chaudhuri, S.~Jegelka, L.~Song,
  C.~Szepesv{\'{a}}ri, G.~Niu, and S.~Sabato, eds.), vol.~162 of {\em
  Proceedings of Machine Learning Research}, pp.~18770--18795, {PMLR}, 2022.

\bibitem{DBLP:journals/corr/abs-2401-06465}
A.~Hedstr{\"{o}}m, L.~Weber, S.~Lapuschkin, and M.~M. H{\"{o}}hne, ``Sanity
  checks revisited: An exploration to repair the model parameter randomisation
  test,'' {\em CoRR}, vol.~abs/2401.06465, 2024.

\bibitem{DBLP:conf/iccv/HesseS023}
R.~Hesse, S.~Schaub{-}Meyer, and S.~Roth, ``Funnybirds: {A} synthetic vision
  dataset for a part-based analysis of explainable {AI} methods,'' in {\em
  {IEEE/CVF} International Conference on Computer Vision, {ICCV} 2023, Paris,
  France, October 1-6, 2023}, pp.~3958--3968, {IEEE}, 2023.

\bibitem{NEURIPS2024_b17799e0}
R.~Hesse, S.~Schaub-Meyer, and S.~Roth, ``Benchmarking the attribution quality
  of vision models,'' in {\em Advances in Neural Information Processing
  Systems} (A.~Globerson, L.~Mackey, D.~Belgrave, A.~Fan, U.~Paquet,
  J.~Tomczak, and C.~Zhang, eds.), vol.~37, pp.~97928--97947, Curran
  Associates, Inc., 2024.

\bibitem{NEURIPS2022_22b11181}
T.~Han, S.~Srinivas, and H.~Lakkaraju, ``Which explanation should i choose? a
  function approximation perspective to characterizing post hoc explanations,''
  in {\em Advances in Neural Information Processing Systems} (S.~Koyejo,
  S.~Mohamed, A.~Agarwal, D.~Belgrave, K.~Cho, and A.~Oh, eds.), vol.~35,
  pp.~5256--5268, Curran Associates, Inc., 2022.

\bibitem{Woerl_2023_CVPR}
A.-C. Woerl, J.~Disselhoff, and M.~Wand, ``Initialization noise in image
  gradients and saliency maps,'' in {\em Proceedings of the IEEE/CVF Conference
  on Computer Vision and Pattern Recognition (CVPR)}, pp.~1766--1775, June
  2023.

\bibitem{pmlr-v139-agarwal21c}
S.~Agarwal, S.~Jabbari, C.~Agarwal, S.~Upadhyay, S.~Wu, and H.~Lakkaraju,
  ``Towards the unification and robustness of perturbation and gradient based
  explanations,'' in {\em Proceedings of the 38th International Conference on
  Machine Learning} (M.~Meila and T.~Zhang, eds.), vol.~139 of {\em Proceedings
  of Machine Learning Research}, pp.~110--119, PMLR, 18--24 Jul 2021.

\bibitem{zhou2024rethinkingprinciplegradientsmooth}
L.~Zhou, C.~Ma, Z.~Wang, and X.~Shi, ``Rethinking the principle of gradient
  smooth methods in model explanation,'' 2024.

\bibitem{simpson2024probabilisticlipschitznessstablerank}
L.~Simpson, K.~Millar, A.~Cheng, C.-C. Lim, and H.~G. Chew, ``Probabilistic
  lipschitzness and the stable rank for comparing explanation models,'' 2024.

\bibitem{scalbert2022testtimeimagetoimagetranslationensembling}
M.~Scalbert, M.~Vakalopoulou, and F.~Couzinié-Devy, ``Test-time image-to-image
  translation ensembling improves out-of-distribution generalization in
  histopathology,'' 2022.

\bibitem{dippel2024aibasedanomalydetectionclinicalgrade}
J.~Dippel, N.~Prenißl, J.~Hense, P.~Liznerski, T.~Winterhoff, S.~Schallenberg,
  M.~Kloft, O.~Buchstab, D.~Horst, M.~Alber, L.~Ruff, K.-R. Müller, and
  F.~Klauschen, ``Ai-based anomaly detection for clinical-grade
  histopathological diagnostics,'' 2024.

\bibitem{10.1007/978-3-031-16434-7_15}
C.~Xu, Z.~Wen, Z.~Liu, and C.~Ye, ``Improved domain generalization for cell
  detection in histopathology images via test-time stain augmentation,'' in
  {\em Medical Image Computing and Computer Assisted Intervention -- MICCAI
  2022} (L.~Wang, Q.~Dou, P.~T. Fletcher, S.~Speidel, and S.~Li, eds.), (Cham),
  pp.~150--159, Springer Nature Switzerland, 2022.

\bibitem{s22082988}
A.~Vulli, P.~N. Srinivasu, M.~S.~K. Sashank, J.~Shafi, J.~Choi, and M.~F. Ijaz,
  ``Fine-tuned densenet-169 for breast cancer metastasis prediction using
  fastai and 1-cycle policy,'' {\em Sensors}, vol.~22, no.~8, 2022.

\bibitem{gaillochet2022taal}
M.~Gaillochet, C.~Desrosiers, and H.~Lombaert, ``Taal: Test-time augmentation
  for active learning in medical image segmentation,'' in {\em MICCAI Workshop
  on Data Augmentation, Labelling, and Imperfections}, pp.~43--53, Springer,
  2022.

\bibitem{MAHBOD2024669}
A.~Mahbod, G.~Dorffner, I.~Ellinger, R.~Woitek, and S.~Hatamikia, ``Improving
  generalization capability of deep learning-based nuclei instance segmentation
  by non-deterministic train time and deterministic test time stain
  normalization,'' {\em Computational and Structural Biotechnology Journal},
  vol.~23, pp.~669--678, 2024.

\bibitem{jimaging8030071}
Y.~Liu, S.~J. Wagner, and T.~Peng, ``Multi-modality microscopy image style
  augmentation for nuclei segmentation,'' {\em Journal of Imaging}, vol.~8,
  no.~3, 2022.

\bibitem{jahanifar2021stainrobustmitoticfiguredetection}
M.~Jahanifar, A.~Shephard, N.~Z. Tajeddin, R.~M.~S. Bashir, M.~Bilal, S.~A.
  Khurram, F.~Minhas, and N.~Rajpoot, ``Stain-robust mitotic figure detection
  for the mitosis domain generalization challenge,'' 2021.

\bibitem{ma2024testtimegenerativeaugmentationmedical}
X.~Ma, Y.~Tao, Y.~Zhang, Z.~Ji, Y.~Zhang, and Q.~Chen, ``Test-time generative
  augmentation for medical image segmentation,'' 2024.

\bibitem{shanmugam2021betteraggregationtesttimeaugmentation}
D.~Shanmugam, D.~Blalock, G.~Balakrishnan, and J.~Guttag, ``Better aggregation
  in test-time augmentation,'' 2021.

\bibitem{ayhan2018testtime}
M.~S. Ayhan and P.~Berens, ``Test-time data augmentation for estimation of
  heteroscedastic aleatoric uncertainty in deep neural networks,'' in {\em
  Medical Imaging with Deep Learning}, 2018.

\bibitem{montavon2019layer}
G.~Montavon, A.~Binder, S.~Lapuschkin, W.~Samek, and K.-R. M{\"u}ller,
  ``Layer-wise relevance propagation: an overview,'' in {\em Explainable AI:
  interpreting, explaining and visualizing deep learning}, pp.~193--209,
  Springer, 2019.

\bibitem{Weyl1912}
H.~Weyl, ``Das asymptotische {V}erteilungsgesetz der {E}igenwerte linearer
  partieller {D}ifferentialgleichungen (mit einer {A}nwendung auf die {T}heorie
  der {H}ohlraumstrahlung),'' {\em Mathematische Annalen}, vol.~71,
  pp.~441--479, Dec 1912.

\bibitem{horn_johnson_matrixanalysis}
R.~A. Horn and C.~R. Johnson, {\em Matrix Analysis}.
\newblock Cambridge University Press, 1990.

\bibitem{Hoeffding01031963}
W.~H. and, ``Probability inequalities for sums of bounded random variables,''
  {\em Journal of the American Statistical Association}, vol.~58, no.~301,
  pp.~13--30, 1963.

\bibitem{Resnet:he2016deep}
K.~He, X.~Zhang, S.~Ren, and J.~Sun, ``Deep residual learning for image
  recognition,'' in {\em {IEEE} Conference on Computer Vision and Pattern
  Recognition (CVPR)}, pp.~770--778, 2016.

\bibitem{Tan_21}
M.~Tan and Q.~V. Le, ``Efficient{N}et{V}2: Smaller models and faster
  training,'' in {\em Proceedings of the 38th International Conference on
  Machine Learning, {ICML} 2021, 18-24 July 2021, Virtual Event} (M.~Meila and
  T.~Zhang, eds.), vol.~139 of {\em Proceedings of Machine Learning Research},
  pp.~10096--10106, {PMLR}, 2021.

\bibitem{Paszke_19}
A.~Paszke, S.~Gross, F.~Massa, A.~Lerer, J.~Bradbury, G.~Chanan, T.~Killeen,
  Z.~M. Lin, N.~Gimelshein, L.~Antiga, A.~Desmaison, A.~Köpf, E.~Yang,
  Z.~DeVito, M.~Raison, A.~Tejani, S.~Chilamkurthy, B.~Steiner, L.~Fang, J.~J.
  Bai, and S.~Chintala, ``Pytorch: An imperative style, high-performance deep
  learning library,'' {\em Advances in Neural Information Processing Systems 32
  (Nips 2019)}, vol.~32, 2019.
\newblock Bp0id Times Cited:23510 Cited References Count:43 Advances in Neural
  Information Processing Systems.

\bibitem{DBLP:conf/cvpr/Liu0LYXWN000WG22}
Z.~Liu, H.~Hu, Y.~Lin, Z.~Yao, Z.~Xie, Y.~Wei, J.~Ning, Y.~Cao, Z.~Zhang,
  L.~Dong, F.~Wei, and B.~Guo, ``Swin transformer {V2:} scaling up capacity and
  resolution,'' in {\em {IEEE/CVF} Conference on Computer Vision and Pattern
  Recognition, {CVPR} 2022, New Orleans, LA, USA, June 18-24, 2022},
  pp.~11999--12009, {IEEE}, 2022.

\bibitem{DenDon09Imagenet}
J.~Deng, W.~Dong, R.~Socher, L.-J. Li, K.~Li, and L.~Fei-Fei, ``Imagenet: A
  large-scale hierarchical image database,'' in {\em Computer Vision and
  Pattern Recognition, 2009. CVPR 2009. IEEE Conference on}, pp.~248--255,
  IEEE, 2009.

\end{thebibliography}


\appendix

\section{Technical Appendices and Supplementary Material}

\subsection{Proof of Theorem \ref{theorem:lrpbeta_svd_upperbound}}

\label{sec:proof_lrpbeta_svd_upperbound}

\textbf{Proof:} We know that for a vector $v$ such that $\|v\|_2 = 1$, to be a singular vector for value $c \ge 0$ implies that there exists another unit norm vector $\|u\|_2=1$ such that:
\eqn{
M^\top v = c u \Rightarrow  v^\top MM^\top v  = c^2 \|u\|_2^2 = c^2 
}
Now:
\eqn{
v^\top MM^\top v  = \sum_{k} ( v\cdot M[:,k]) (M^\top[k,:] \cdot v) =  \sum_{k} ( v\cdot M[:,k])^2 
}
We can maximize an inner product by choosing $v = \frac{M[:,k]}{\|M[:,k]\|}  $. This yields here an upper bound because we have $k$ vectors $M[:,k]$ but only one vector $v$.
\eqn{
&\sup_{v:\|v\|_2=1}v^\top MM^\top v = \sup_{v:\|v\|_2=1} \sum_{k} ( v\cdot M[:,k])^2 \le \sum_{k} (\frac{1}{\|M[:,k]\|}  M[:,k] \cdot M[:,k])^2 \\
&= \sum_k \|M[:,k]\|^2_2
}
Next we consider the specific shape of $M[:,k]$ under LRP-$\beta$. $M[s,k]$ contains in every sum exclusively either a term $(1+\beta)(w_{ks} \cdot z_s)_+ / C_{k+}$ or a term $-\beta (w_{ks} \cdot z_s)_- / C_{k-}$. Both cannot be present at the same time.\\

We aim to compute $\|M[:,k]\|^2_2$. This norm is invariant to reordering the components of the vector $M[:,k]$. We can assume without loss of generality after ordering the terms according to the sign of $w_{ks} \cdot z_s$ that 
\eqn{
M[:,k] = ( (1+\beta)p_{1,+},\ldots,  (1+\beta)p_{t,+}, -\beta p_{t+1,-} , \ldots, -\beta p_{S,-} )
}
where $\sum_{i=1}^t p_{i,+} = 1$ and $\sum_{i=t+1}^S p_{i,-} = 1$. 

This is due to the fact that in LRP-$\beta$ the positive entries $\frac{ (w_{ab} z_{b})_+  }{ \sum_{b'} (w_{ab'} z_{b'})_+  }$ and the negative entries $\frac{ (w_{ab} z_{b})_-  }{ \sum_{b'} (w_{ab'} z_{b'})_-  }$ are separately normalized to sum up to 1 for both signs. Now
\eqn{
\|M[:,k]\|^2_2 =&(  (1+\beta)^2\sum_{i=1}^t p_{i,+}^2 +\beta^2 \sum_{i=t+1}^S p_{i,-}^2 \le (1+\beta)^2\sum_{i=1}^t p_{i,+} +\beta^2 \sum_{i=t+1}^S p_{i,-}\\
=& (1+\beta)^2 + \beta^2 
}

To obtain an upper bound on the largest singular value, we have to consider
\eqn{
\sup_{v:\|v\|_2=1}v^\top MM^\top v  \le \sum_{k=1}^R \|M[:,k]\|^2_2 \le R ((1+\beta)^2 + \beta^2)
}
Taking the square root results in 
\eqn{
\sqrt{R} \sqrt{(1+\beta)^2 + \beta^2}
}

A more interpretable form can be derived by
\eqn{
\sqrt{R}\sqrt{(1+\beta)^2 + \beta^2} = \sqrt{R}\sqrt{ 1 + 2\beta +2\beta^2} \le \sqrt{R}\sqrt{ 1 + 2\sqrt{2}\beta +(\sqrt{2}\beta)^2} = \sqrt{R}(1+\sqrt{2}\beta)
}

\subsection{Proof of Lemma \ref{lemma:seqbound_lrpbeta}}
\label{sec:prooflemmaseqbound}

\textbf{Proof:}

We considering the LRP-$\beta$ term 
\eqn{
Att( g\e{r}_a , z_b\e{r-1}) = (1+\beta) \frac{ (w_{ab} z_b)_+  }{ \sum_{b'} (w_{ab'} z_{b'})_+  } - \beta  \frac{ (w_{ab} z_b)_-  }{ \sum_{b'} (w_{ab'} z_{b'})_-  } 
}
for a layer $r$ computing $g\e{r}(z) = Wz +c$. To simplify notation, we define
\eqn{
p_{ab+}&:=  \frac{ (w_{ab} z_b)_+  }{ \sum_{b'} (w_{ab'} z_{b'})_+  }\\
p_{ab-}&:= \frac{ (w_{ab} z_b)_-  }{ \sum_{b'} (w_{ab'} z_{b'})_-  } \\
Att( g\e{r}_a , z_b\e{r-1})  &= (1+\beta) p_{ab+} -\beta p_{ab-}
}
We note that $p_{ab+} \in [0,1]$, $p_{ab-} \in [0,1] $ . One can observe that $w_{ab} z_b$ is either non-negative or non-positive. Therefore,  one of the terms $p_{ab+}  $ and $p_{ab-}$ must always be a zero term.\\

\paragraph{Induction start $t=1$:} We initially the computation of the attribution map at the network output across output components $g_1\e{n},\ldots, g_{D_n}\e{n}$ at the last layer $n$ using a vector $q$ such that $\sum_{u=1}^{D_n} q_u = 1,\ q_u \ge 0$, that is we compute the attribution map for the weighted sum of outputs $\sum_{u=1}^{D_n} q_u g_{u}\e{n}$.\\

The attribution map in the next upstream layer, for the component $z_b$ of the feature map $z\e{n-1}$, for which we have to prove the bounds, will be
\eqn{
\sum_{u=1}^{D_n} q_u Att( g\e{n}_u , z_b\e{n-1})
}

Applying LRP-$\beta$ to $g_{u}\e{n}$ with the weights $q_u$ results in
\eqn{
\sum_{u=1}^{D_n} q_u Att( g\e{n}_u , z_b\e{n-1}) = \sum_{u=1}^{D_n} q_u ( (1+\beta) p_{ub+} -\beta p_{ub-}) 
}
Using the observation that one of $p_{ub+}$, $p_{ub-}$ is always zero, we can write it as
\eqn{
=  \sum_{u}  q_u ( \underbrace{(1+\beta) p_{ub+} }_{\ge 0}) +  \sum_{ u }  q_u ( \underbrace{-\beta p_{ub-} }_{\le 0})
}
Lets prove the upper bound. We need to bound
\eqn{
\sum_{b: \sum_{u=1} q_u Att( g\e{n}_u , z_b) > 0 } \sum_{u=1}^{D_n} q_u Att( g\e{n}_u , z_b\e{n-1})
}
For this we observe: if $b$ satisfies $\sum_{u=1} q_u Att( g\e{n}_u , z_b\e{n-1}) > 0$,  there must exist $u: p_{ub+}>0$ (due to $v_u \ge 0$).

Lets define the following true/false logical functions

$Y_{+}(b)= \sum_{u=1} q_u Att( g\e{n}_u , z_b\e{n-1}) > 0 $

$Y_{-}(b)= \sum_{u=1} q_u Att( g\e{n}_u , z_b\e{n-1}) < 0$

Therefore, using $\sum_{b: Y_{+}(b)}$ to denote those $b$ for which the function evaluates to true: 
\eqn{
&\sum_{\{b: \sum_{u=1} q_u Att( g\e{n}_u , z_b) > 0 \}} \sum_{u=1}^{D_n} q_u Att( g\e{n}_u , z_b) = \sum_{b: Y_{+}(b) } \sum_{u=1}^{D_n} q_u Att( g\e{n}_u , z_b) \\
=& \sum_{b: Y_{+}(b) } \sum_{u }  \underbrace{q_u ( (1+\beta) p_{ub+} )}_{\ge 0} +  \sum_{ u  }  \underbrace{q_u ( -\beta p_{ub-} )}_{\le 0}\\
\le & \sum_{b: Y_{+}(b) } \sum_{u}  q_u ( (1+\beta) p_{ub+} ) +0\\
\le & \sum_{b } \sum_{u }  q_u ( (1+\beta) p_{ub+}   \\
= &   \sum_{u }  q_u  (1+\beta) \sum_{b } p_{ub+} = \sum_{u }q_u  (1+\beta)=  (1+\beta)
}

For the lower bound we use an analogous argument:
\eqn{
&\sum_{\{b: \sum_{u=1} q_u Att( g\e{n}_u , z_b) < 0 \}} \sum_{u=1}^{D_n} q_u Att( g\e{n}_u , z_b) 
= \sum_{b: Y_{-}(b) } \sum_{u=1}^{D_n} q_u Att( g\e{n}_u , z_b)  \\
=& \sum_{b: Y_{-}(b) } \sum_{u }  \underbrace{q_u ( (1+\beta) p_{ub+}) }_{\ge 0} +  \sum_{ u  }  \underbrace{q_u ( -\beta p_{ub-} )}_{\le 0} \\
\ge & \sum_{b: Y_{-}(b) } 0+ \sum_{ u  }  q_u ( -\beta p_{ub-} )\\
\ge & \sum_{b } \sum_{u }  q_u ( -\beta p_{ub-})  \\
= & \sum_{u}  q_u ( -\beta)  \sum_{b } p_{ub-} = \sum_{u }q_u  (-\beta) =  -\beta
}

\paragraph{Induction step $t-1 \rightarrow t$:}

We are given now attribution scores $v_u$ such that 
\eqn{v_u = \sum_r q_r Att(g\e{n}_r, z\e{n-(t-1)}_u) 
}
These are the attribution scores for the feature map $z\e{n-(t-1)}$ of layer $n-(t-1)$ backpropagated from the weighted output of the network $\sum_r q_r g\e{n}_r$. 

$v_u$ is the score for the $u$-th component $z\e{n-(t-1)}_u$ of vector $z\e{n-(t-1)}$.

We can assume that for these attribution scores $v_u$ from the layer $z\e{n-(t-1)}$ downstream we have $\sum_u v_u = 1$, however, we can have now both signs for values $v_u$.

Note that the set of scores $\{v_u\}$ satisfies the induction assumption as stated above for layer $n-(t-1)$.

It should be noted here that due to equations \eqref{eq:attrmap_equation} or \eqref{eq:attrmap_equation_part} we have
\eqn{
&\sum_r q_r Att(g\e{n}_r, z\e{n-t)}_b) =
  q^\top M^\top (g\e{n}) \cdot M^\top (g\e{n-1})  \cdot \ldots \cdot M^\top (g\e{n-(t-1)})(z\e{n-t}_b)\\
=&    \left(q^\top M^\top (g\e{n}) \cdot M^\top (g\e{n-1})  \cdot \ldots \cdot M^\top (g\e{n-t+2})(z\e{n-(t-1)})\right) \cdot M^\top (g\e{n-(t-1)})(z\e{n-t}_b) \\
=& \sum_u  \left(\sum_r q_r Att(g\e{n}_r, z\e{n-(t-1)}_u)\right)  Att( g\e{n-(t-1)}_u , z_b\e{n-t})\\
=& \sum_u v_u  Att( g\e{n-(t-1)}_u , z_b\e{n-t}) \label{eq:vueq1}\\
=& v \cdot M^\top (g\e{n-(t-1)})(z\e{n-t}_b)
}

Therefore, as a consequence of equation \eqref{eq:vueq1}, we need to obtain bounds for
\eqn{
\sum_{b: \sum_u v_u Att( g\e{n-(t-1)}_u , z_b\e{n-t}) > 0 }\sum_u v_u Att( g\e{n-(t-1)}_u , z_b\e{n-t})\\
\sum_{b: \sum_u v_u Att( g\e{n-(t-1)}_u , z_b\e{n-t}) < 0 }\sum_u v_u Att( g\e{n-(t-1)}_u , z_b\e{n-t})
}

Lets define the true/false-valued functions 

$Y_+(b)= \sum_u v_u Att( g_u\e{n-(t-1)} , z_b\e{n-t})> 0$,

$Y_-(b)= \sum_u v_u Att( g_u\e{n-(t-1)} , z_b\e{n-t}) < 0$ 

We will shorten $g_u\e{n-(t-1)}$ to $g_u$ and $z\e{n-t}_b$ to $z_b$ further below.

Lets consider the upper bound first. For the upper bound we can separate terms by their signs to obtain
\eqn{
&\sum_{b: Y_{+}(b)} \sum_u v_u Att( g_u , z_b) = \sum_{b: Y_{+}(b)} \sum_u v_u (1+\beta) p_{ub+} + \sum_u v_u (-\beta) p_{ub-}\\
=& \sum_{b: Y_{+}(b)} \sum_{u: v_u >0} v_u (1+\beta) p_{ub+} + \sum_{u: v_u >0} v_u (-\beta) p_{ub-}\\
+& \sum_{b: Y_{+}(b)} \sum_{u: v_u <0} v_u (1+\beta) p_{ub+} + \sum_{u: v_u <0} v_u (-\beta) p_{ub-}\\
\le& \sum_{b: Y_{+}(b)} \sum_{u: v_u >0} v_u (1+\beta) p_{ub+} + \sum_{u: v_u >0} \mathcolor{orange}{0}\\
+& \sum_{b: Y_{+}(b)} \sum_{u: v_u <0} \mathcolor{orange}{0} + \sum_{u: v_u <0} v_u (-\beta) p_{ub-}\\
=& \sum_{b: Y_{+}(b)} \sum_{u: v_u >0} v_u (1+\beta) p_{ub+}  + \sum_{u: v_u <0} v_u (-\beta) p_{ub-}
}
The above inequality comes from checking the signs of the terms.

In the following $\sum_b f(b)$ denotes the sum over all $b$, while $\sum_{b: Y_{+}(b)} f(b)$ is the sum over the subset of input indices $b$ for which $Y_{+}(b)$ evaluates to true.   

All terms in the last statement are non-negative (note $p_{ub+} \in [0,1], \ p_{ub-} \in [0,1]$). 

Therefore we can upper bound
\eqn{
&\sum_{b: Y_{+}(b)} \sum_{u: v_u >0} v_u (1+\beta) p_{ub+}  + \sum_{u: v_u <0} v_u (-\beta) p_{ub-}\\
\le& \sum_{b} \sum_{u: v_u >0} v_u (1+\beta) p_{ub+}  + \sum_{u: v_u <0} v_u (-\beta) p_{ub-} \\
=& \sum_{u: v_u >0}  v_u (1+\beta)\sum_{b} p_{ub+}  + \sum_{u: v_u <0}  v_u (-\beta) \sum_{b}p_{ub-} \\
=&  \sum_{u: v_u >0} v_u (1+\beta)  + \sum_{u: v_u <0} v_u (-\beta)
}

Now $\sum_{v_u >0 } v_u$ are the positive scores from the next downstream layer $n-(t-1)$. They satisfy according to the induction assumption 
\eqn{
\sum_{ v_u >0 } v_u \le +2^{t-2} (1+\beta)^{t-1}
}
Analogously, $\sum_{ v_u <0 }  v_u$ are the negative scores from the next downstream layer $n-(t-1)$. They satisfy according to the induction assumption 
\eqn{
\sum_{ v_u <0 }  v_u \ge -2^{t-2} \beta (1+\beta)^{t-2} \Leftrightarrow \sum_{ v_u <0 }  (-v_u) \le +2^{t-2} \beta (1+\beta)^{t-2}
}
Plugging these inequalities in, results in
\eqn{
&  \sum_{ v_u >0 } v_u (1+\beta) + \sum_{ v_u <0 }  (-v_u) \beta \\
\le & +2^{t-2} (1+\beta)^{t-1} (1+\beta) + 2^{t-2} \beta^2 (1+\beta)^{t-2} \le  +2^{t-2} (1+\beta)^{t} + 2^{t-2} (1+\beta)^{t} \\
= & 2^{t-1} (1+\beta)^{t}
}

For the lower bound we can use an analogous reasoning:  We can look at the signs to obtain
\eqn{
&\sum_{b: Y_{-}(b)} \sum_u v_u Att( g_u , z_b) = \sum_{b: Y_{-}(b)} \sum_u v_u (1+\beta) p_{ub+} + \sum_u v_u (-\beta) p_{ub-}\\
=& \sum_{b: Y_{-}(b)} \sum_{u: v_u >0} v_u (1+\beta) p_{ub+} + \sum_{u: v_u >0} v_u (-\beta) p_{ub-}\\
+& \sum_{b: Y_{-}(b)} \sum_{u: v_u <0} v_u (1+\beta) p_{ub+} + \sum_{u: v_u <0} v_u (-\beta) p_{ub-}\\
\ge& \sum_{b: Y_{-}(b)} \sum_{u: v_u >0} \mathcolor{orange}{0} + \sum_{u: v_u >0} v_u (-\beta) p_{ub-}\\
+& \sum_{b: Y_{-}(b)} \sum_{u: v_u <0} v_u (1+\beta) p_{ub+} + \sum_{u: v_u <0} \mathcolor{orange}{0} \\
=& \sum_{b: Y_{-}(b)} \sum_{u: v_u >0} v_u (-\beta) p_{ub-}  +\sum_{u: v_u <0} v_u (1+\beta) p_{ub+}
}



All terms are non-positive (note $p_{ub+} \in [0,1], \ p_{ub-} \in [0,1]$). 

Therefore we can lower bound
\eqn{
&\sum_{b: Y_{-}(b)} \sum_{u: v_u >0} v_u (-\beta) p_{ub-}  +\sum_{u: v_u <0} v_u (1+\beta) p_{ub+} \\
\ge&
\sum_{b} \sum_{u: v_u >0} v_u (-\beta) p_{ub-}  +\sum_{u: v_u <0} v_u (1+\beta) p_{ub+}\\
=&  \sum_{u: v_u >0}v_u (-\beta)  \sum_{b} p_{ub-}  +\sum_{u: v_u <0} v_u (1+\beta) \sum_{b}p_{ub+}\\
=& \sum_{u: v_u >0} v_u (-\beta)  +\sum_{u: v_u <0} v_u (1+\beta)
}

By induction assumption we have bounds as follows:
\eqn{
&\sum_{ v_u >0 } v_u \le 2^{t-2} (1+\beta)^{t-1} \\
& \sum_{ v_u <0 } v_u \ge  -2^{t-2} \beta(1+\beta)^{t-2}
}
Plugging them in yields:
\eqn{
&\sum_{ v_u >0 } v_u (-\beta) + \sum_{ v_u <0 } v_u (1+\beta)\\
\ge & 2^{t-2} (1+\beta)^{t-1}(-\beta)  -2^{t-2} \beta (1+\beta)^{t-2} (1+\beta) = -2^{t-2} \beta(1+\beta)^{t-1} -2^{t-2} \beta (1+\beta)^{t-1}  \\
=& -2^{t-1} \beta(1+\beta)^{t-1}
}
This concludes the upper and the lower bound in the induction step.

\newpage
\subsection{Proof of Lemma \ref{lemma:seqbound_lrpgamma}}
\label{sec:prooflemmaseqbound_gamma}

We considering the LRP-$\gamma$ term 
\eqn{
Att( g_a , z_b) = \frac{ w_{ab}z_b + \gamma (w_{ab}z_b)_+ }{\sum_{b'} w_{ab'}z_{b'} + \gamma (w_{ab}z_{b'})_+ } 
}
Let us omit layer indices again, and define as notation
\eqn{
y_{ab}=w_{ab}z_b 
}

\paragraph{Induction start $t=1$:} We initially the computation of the attribution map at the network output across output components $g_1\e{n},\ldots, g_{D_n}\e{n}$ at the last layer $n$ using a vector $q$ such that $\sum_{u=1}^{D_n} q_u = 1,\ q_u \ge 0$, that is we compute the attribution map for the weighted sum of outputs $\sum_{u=1}^{D_n} q_u g_{u}\e{n}$.\\

The attribution map in the next upstream layer, for which we have to prove the bounds, will be
\eqn{
\sum_{u=1}^{D_n} q_u Att( g\e{n}_u , z_b)
}

Applying LRP-$\gamma$ to $g_{u}\e{n}$ with weights $q_u$ results in
\eqn{
 \sum_{u=1}^{D_n} q_u Att( g_u , z_b) &= \sum_{u=1}^{D_n} q_u\frac{ y_{ub} + \gamma (y_{ub})_+ }{\sum_{b'} y_{ub'} + \gamma (y_{ub'})_+ } = \sum_{u=1}^{D_n} q_u \frac{ \gamma^{-1} y_{ub} + (y_{ub})_+ }{\sum_{b'}  \gamma^{-1}y_{ub'} + (y_{ub'})_+ } \\
=&  \sum_{u : y_{ub}>0 } q_u \frac{ \gamma^{-1} y_{ub} + (y_{ub})_+ }{\sum_{b'}  \gamma^{-1}y_{ub'} + (y_{ub'})_+ } 
+ \sum_{u : y_{ub}<0  } q_u \frac{ \gamma^{-1} y_{ub} + (y_{ub})_+ }{\sum_{b'}  \gamma^{-1}y_{ub'} + (y_{ub'})_+ }\\
=& \sum_{u : y_{ub}>0 } q_u \frac{   (1+\gamma^{-1})(y_{ub})_+ }{\sum_{b'}  \gamma^{-1}y_{ub'} + (y_{ub'})_+ } 
+ \sum_{u : y_{ub}<0  } q_u \frac{ \gamma^{-1} y_{ub} }{\sum_{b'}  \gamma^{-1}y_{ub'} + (y_{ub'})_+ }
}

If $\sum_b (y_{ub})_+ = 0$ it means that all $y_{ub}<0$, then we get:
\eqn{
& \sum_{b: Y_{+}(b)}  \sum_{u=1}^{D_n} q_u Att( g_u , z_b) = 0
+ \sum_{b: Y_{+}(b)} \sum_{u : y_{ub}<0  } q_u \frac{ \gamma^{-1} y_{ub} }{\sum_{b'}  \gamma^{-1}y_{ub'} + 0 } \\
=& \sum_{b: Y_{+}(b)} \sum_{u : y_{ub}<0  } q_u \frac{  y_{ub} }{\sum_{b'}  y_{ub'}  } =  \sum_{b: Y_{+}(b)}\sum_{u : y_{ub}<0  } q_u \frac{ (y_{ub})_{-} }{ \sum_{b'}  (y_{ub'})_{-}} \\
\le & \sum_{b}\sum_{u  } q_u \frac{ y_{ub} }{ \sum_{b'}  y_{ub'}} =\sum_{u :  } q_u  \sum_{b} \frac{ y_{ub} }{ \sum_{b'}  y_{ub'}}= \sum_{u   } q_u = 1
}

If $\sum_b (y_{ub})_+ > 0$, then we require by the assumption of the lemma (in the lemma we have set $\alpha = \gamma^{-1/2}$  as seen further below, while here we execute it for a general $\alpha \in(0,1) $ )
\eqn{
\gamma^{-1}\sum_{b: y_{ub}<0} y_{ub}&> -\alpha \sum_{b: y_{ub}>0}  (y_{ub})_+\\
\Leftrightarrow \sum_{b: y_{ub}<0} \gamma^{-1}y_{ub} + \sum_{b: y_{ub}>0} (1+\gamma^{-1})(y_{ub})_+ &> -\alpha \sum_{b: y_{ub}>0}  (y_{ub})_+ + (1+\gamma^{-1}) \sum_{b: y_{ub}>0}  (y_{ub})_+ \\
\Leftrightarrow \sum_{b'}  \gamma^{-1}y_{ub'} + (y_{ub'})_+  &
> (1+\gamma^{-1} - \alpha ) \sum_{b'} (y_{ub'})_+
}
The left hand side holds due to
\eqn{
 \sum_{b: y_{ub}<0} \gamma^{-1}y_{ub} + \sum_{b: y_{ub}>0} (1+\gamma^{-1})(y_{ub})_+ &=  \sum_{b: y_{ub}<0} \gamma^{-1}y_{ub} + (y_{ub})_+  + \sum_{b: y_{ub}>0} \gamma^{-1} y_{ub} +(y_{ub})_+ \\
 &= \sum_{b'}  \gamma^{-1}y_{ub'} + (y_{ub'})_+ 
}

Then, an upper bound will be
\eqn{
& \sum_{b: Y_{+}(b)}  \sum_{u=1}^{D_n} q_u Att( g_u , z_b)\\
\le& \sum_{b: Y_{+}(b)}\sum_{u : y_{ub}>0 } q_u \frac{   (1+\gamma^{-1})(y_{ub})_+ }{\sum_{b'}  \gamma^{-1}y_{ub'} + (y_{ub'})_+ } \\
\le & \sum_{b: Y_{+}(b)}\sum_{u : y_{ub}>0 } q_u \frac{   (1+\gamma^{-1})(y_{ub})_+ }{ (1+\gamma^{-1} - \alpha ) \sum_{b'} (y_{ub'})_+ }\\
=& \frac{1+\gamma^{-1}}{1+\gamma^{-1} - \alpha} \sum_{b: Y_{+}(b)}\sum_{u : y_{ub}>0 } q_u \frac{ (y_{ub})_+ }{\sum_{b'} (y_{ub'})_+ }
}
Next we use the trick, that we can drop the conditioning on $u : y_{ub}>0$ because the terms in the upper bound would be simply zero if $y_{ub}<0$. After that we can sum over all input dimensions $b$ because all terms have the same positive sign or are zero.
\eqn{
\le& \frac{1+\gamma^{-1}}{1+\gamma^{-1} - \alpha}\sum_{b : Y_{+}(b)}\sum_{u  } q_u \frac{ (y_{ub})_+ }{\sum_{b'} (y_{ub'})_+ } \le\frac{1+\gamma^{-1}}{1+\gamma^{-1} - \alpha} \sum_{b }\sum_{u  } q_u \frac{ (y_{ub})_+ }{\sum_{b'} (y_{ub'})_+ }\\
&= \frac{1+\gamma^{-1}}{1+\gamma^{-1} - \alpha}\sum_{u  } q_u \frac{ \sum_{b }(y_{ub})_+ }{\sum_{b'} (y_{ub'})_+ }= \frac{1+\gamma^{-1}}{1+\gamma^{-1} - \alpha} \sum_{u  } q_u = \frac{1+\gamma^{-1}}{1+\gamma^{-1} - \alpha}
}
This proves an upper bound in the induction step of $\frac{1+\gamma^{-1}}{1+\gamma^{-1} - \alpha}$.\\

For a lower bound we use
\eqn{
\gamma^{-1}\sum_{b: y_{ub}<0} y_{ub}&> -\alpha \sum_{b: y_{ub}>0}  (y_{ub})_+
\\
\Leftrightarrow 0 \le -\gamma^{-1}\alpha^{-1}\sum_{b: y_{ub}<0} y_{ub}&<  \sum_{b: y_{ub}>0}  (y_{ub})_+ = \sum_{b'}  (y_{ub'})_+
}
so that
\eqn{
&\sum_{b: Y_{-}(b)}  \sum_{u=1}^{D_n} q_u Att( g_u , z_b) \ge 0
+ \sum_{u : y_{ub}<0  } q_u \frac{ \gamma^{-1} y_{ub} }{\sum_{b'}  \gamma^{-1}y_{ub'} + (y_{ub'})_+ } \\
\ge& \sum_{b: Y_{-}(b)}\sum_{u : y_{ub}<0  } q_u \frac{ \gamma^{-1} y_{ub} }{\sum_{b': y_{ub'}<0}  \gamma^{-1}(1-\alpha^{-1} )y_{ub'}  } \\
=& (1-\alpha^{-1} )^{-1} \sum_{b: Y_{-}(b)}\sum_{u : y_{ub}<0  } q_u  \frac{  y_{ub} }{\sum_{b': y_{ub'}<0} y_{ub'}  }\\
=& (1-\alpha^{-1} )^{-1} \sum_{b: Y_{-}(b)}\sum_{u : y_{ub}<0  } q_u  \frac{  (y_{ub})_{-} }{\sum_{b': y_{ub'}<0} (y_{ub'})_{-}  }\\
=& (1-\alpha^{-1} )^{-1} \sum_{b: Y_{-}(b)}\sum_{u   } q_u  \frac{  (y_{ub})_{-} }{\sum_{b'} (y_{ub'})_{-}  }\\
\ge & (1-\alpha^{-1} )^{-1} \sum_{b}\sum_{u   } q_u  \frac{  (y_{ub})_{-} }{\sum_{b'} (y_{ub'})_{-}  }\\
=&(1-\alpha^{-1} )^{-1} \sum_{u   } q_u = \frac{\alpha}{\alpha-1}
}


We obtain

 for the sum of positive attributions
\eqn{
\sum_{b: Y_{+}(b)}  \sum_{u=1}^{D_n} q_u Att( g_u\e{n} , z_b\e{n-1})\le \frac{   1+\gamma^{-1}}{1+\gamma^{-1}-\alpha}
}
for the sum of negative attributions as 
\eqn{
\sum_{b: Y_{-}(b)}  \sum_{u=1}^{D_n} q_u Att( g_u\e{n} , z_b\e{n-1})\ge \frac{ \alpha}{ \alpha -1}
}
To simplify, set $\alpha := \gamma^{-1/2}$, resulting in a requirement of 
\eqn{
\gamma^{-1/2}\sum_{b: y_{ub}<0} (-1)y_{ub}<  \sum_{b: y_{ub}>0}  (y_{ub})_+
}
then we get
 for the sum of positive attributions
\eqn{
\le \frac{   1+\gamma^{-1}}{1+\gamma^{-1}-\gamma^{-1/2}}= \frac{1+\gamma}{1+\gamma - \gamma^{1/2}}
}
for the sum of negative attributions as 
\eqn{
\ge \frac{ 1}{ 1 -\gamma^{1/2}}
}

\paragraph{Induction step $t-1 \rightarrow t$:}

To start with, by our assumption of the lemma, we have set $\gamma$ such that for all activations $y_{ub} = w_{ub}z_{b}$ we have
\eqn{
\gamma^{-1}\sum_{b: y_{ub}<0} y_{ub} > -\gamma^{-1/2} \sum_{b: y_{ub}>0}  (y_{ub})_+
}

We are given now attribution scores $v_u$ such that 
\eqn{v_u = \sum_r q_r Att(g\e{n}_r, z\e{n-(t-1)}_u) 
}
These are the attribution scores for the feature map $z\e{n-(t-1)}$ of layer $n-(t-1)$ backpropagated from the weighted output of the network $\sum_r q_r g\e{n}_r$.


Note that the set of scores $\{v_u\}$ satisfies the induction assumption as stated above for layer $n-(t-1)$, that is
\eqn{
&\sum_{u: v_u <0} v_u \ge 2^{t-2}\frac{ 1}{ 1 -\gamma^{1/2}}b(\gamma)^{t-2}\\
&\sum_{u: v_u >0} v_u \le  2^{t-2} \frac{1+\gamma}{1+\gamma - \gamma^{1/2}}  b(\gamma)^{t-2}
}
Take note that for $\gamma > 1$ : $\frac{ 1}{ 1 -\gamma^{1/2}} <0$\\

It should be noted here that due to equations \eqref{eq:attrmap_equation} or \eqref{eq:attrmap_equation_part} we have
\eqn{
&\sum_r q_r Att(g\e{n}_r, z\e{n-t)}_b) =
  q^\top M^\top (g\e{n}) \cdot M^\top (g\e{n-1})  \cdot \ldots \cdot M^\top (g\e{n-(t-1)})(z\e{n-t}_b)\\
=&    \left(q^\top M^\top (g\e{n}) \cdot M^\top (g\e{n-1})  \cdot \ldots \cdot M^\top (g\e{n-t+2})(z\e{n-(t-1)})\right) \cdot M^\top (g\e{n-(t-1)})(z\e{n-t}_b) \\
=& \sum_u  \left(\sum_r q_r Att(g\e{n}_r, z\e{n-(t-1)}_u)\right)  Att( g\e{n-(t-1)}_u , z_b\e{n-t})\\
=& \sum_u v_u  Att( g\e{n-(t-1)}_u , z_b\e{n-t}) \label{eq:vueq2}\\
=& v \cdot M^\top (g\e{n-(t-1)})(z\e{n-t}_b)
}

Therefore, as a consequence of equation \eqref{eq:vueq2}, we need to obtain bounds for
\eqn{
\sum_{b: \sum_u v_u Att( g\e{n-(t-1)}_u , z_b\e{n-t}) > 0 }\sum_u v_u Att( g\e{n-(t-1)}_u , z_b\e{n-t})\\
\sum_{b: \sum_u v_u Att( g\e{n-(t-1)}_u , z_b\e{n-t}) < 0 }\sum_u v_u Att( g\e{n-(t-1)}_u , z_b\e{n-t})
}

Lets define the true/false-valued functions 

$Y_+(b)= \sum_u v_u Att( g_u\e{n-(t-1)} , z_b\e{n-t})> 0$,

$Y_-(b)= \sum_u v_u Att( g_u\e{n-(t-1)} , z_b\e{n-t}) < 0$ 

We will shorten $g_u\e{n-(t-1)}$ to $g_u$ and $z\e{n-t}_b$ to $z_b$ further below.

Let $b(\gamma):= \max(  -\frac{ 1}{ 1 -\gamma^{1/2}}, \frac{1+\gamma}{1+\gamma - \gamma^{1/2}} ) $

Applying LRP-$\gamma$ to $g_{u}$ with weights $v_u$ results in
\eqn{
 \sum_{u=1} v_u Att( g_u , z_b) &= \sum_{u=1} v_u\frac{ y_{ub} + \gamma (y_{ub})_+ }{\sum_{b'} y_{ub'} + \gamma (y_{ub'})_+ } = \sum_{u=1} v_u \frac{ \gamma^{-1} y_{ub} + (y_{ub})_+ }{\sum_{b'}  \gamma^{-1}y_{ub'} + (y_{ub'})_+ } \\
=&  \sum_{u : y_{ub}>0 } v_u \frac{ \gamma^{-1} y_{ub} + (y_{ub})_+ }{\sum_{b'}  \gamma^{-1}y_{ub'} + (y_{ub'})_+ } 
+ \sum_{u : y_{ub}<0  } v_u \frac{ \gamma^{-1} y_{ub} + (y_{ub})_+ }{\sum_{b'}  \gamma^{-1}y_{ub'} + (y_{ub'})_+ }\\
=& \sum_{u : y_{ub}>0 } v_u \frac{   (1+\gamma^{-1})(y_{ub})_+ }{\sum_{b'}  \gamma^{-1}y_{ub'} + (y_{ub'})_+ } 
+ \sum_{u : y_{ub}<0  } v_u \frac{ \gamma^{-1} y_{ub} }{\sum_{b'}  \gamma^{-1}y_{ub'} + (y_{ub'})_+ }
}
Now we have to split this further according to signs of $v_u$:
\eqn{
 = & \sum_{u : v_u>0, y_{ub}>0 } v_u \frac{   (1+\gamma^{-1})(y_{ub})_+ }{\sum_{b'}  \gamma^{-1}y_{ub'} + (y_{ub'})_+ } 
+ \sum_{u : v_u>0, y_{ub}<0  } v_u \frac{ \gamma^{-1} y_{ub} }{\sum_{b'}  \gamma^{-1}y_{ub'} + (y_{ub'})_+ } \\
+& \sum_{u : v_u<0,y_{ub}>0 } v_u \frac{   (1+\gamma^{-1})(y_{ub})_+ }{\sum_{b'}  \gamma^{-1}y_{ub'} + (y_{ub'})_+ } 
+ \sum_{u : v_u<0,y_{ub}<0  } v_u \frac{ \gamma^{-1} y_{ub} }{\sum_{b'}  \gamma^{-1}y_{ub'} + (y_{ub'})_+ } 
}

For the upper bound we can derive from that:
\eqn{
&\sum_{b: Y_{+}(b)}  \sum_{u=1} v_u Att( g_u , z_b)\\
 = & \sum_{b: Y_{+}(b)} \sum_{u : v_u>0, y_{ub}>0 } v_u \frac{   (1+\gamma^{-1})(y_{ub})_+ }{\sum_{b'}  \gamma^{-1}y_{ub'} + (y_{ub'})_+ } 
+ \sum_{u : v_u>0, y_{ub}<0  } v_u \frac{ \gamma^{-1} y_{ub} }{\sum_{b'}  \gamma^{-1}y_{ub'} + (y_{ub'})_+ } \\
+& \sum_{b: Y_{+}(b)}\sum_{u : v_u<0,y_{ub}>0 } v_u \frac{   (1+\gamma^{-1})(y_{ub})_+ }{\sum_{b'}  \gamma^{-1}y_{ub'} + (y_{ub'})_+ } 
+ \sum_{u : v_u<0,y_{ub}<0  } v_u \frac{ \gamma^{-1} y_{ub} }{\sum_{b'}  \gamma^{-1}y_{ub'} + (y_{ub'})_+ } \\
\le &  \sum_{b: Y_{+}(b)}\sum_{u : v_u>0, y_{ub}>0 } v_u \frac{   (1+\gamma^{-1})(y_{ub})_+ }{\sum_{b'}  \gamma^{-1}y_{ub'} + (y_{ub'})_+ } 
+ 0 \\
+& \sum_{b: Y_{+}(b)} 0+\sum_{u : v_u<0,y_{ub}<0  } v_u \frac{ \gamma^{-1} y_{ub} }{\sum_{b'}  \gamma^{-1}y_{ub'} + (y_{ub'})_+ } 
}

For the upper line we will use from our requirement
\eqn{
\gamma^{-1}\sum_{b: y_{ub}<0} y_{ub}&> -\gamma^{-1/2} \sum_{b: y_{ub}>0}  (y_{ub})_+\\
\Leftrightarrow \sum_{b'}  \gamma^{-1}y_{ub'} + (y_{ub'})_+  &
> (1+\gamma^{-1} - \gamma^{-1/2} ) \sum_{b'} (y_{ub'})_+
\label{eq:lowerboundgamma1}
}
For the lower line, we employ
\eqn{
&\sum_{b'}  \gamma^{-1}y_{ub'} + (y_{ub'})_+ =  
\sum_{y_{ub'}<0 }  \gamma^{-1}y_{ub'} + (y_{ub'})_+ 
+ \sum_{y_{ub'}>0 }  \gamma^{-1}y_{ub'} + (y_{ub'})_+\\ 
=& \sum_{y_{ub'}<0 }  \gamma^{-1}y_{ub'} +  \sum_{y_{ub'}>0 } (1+\gamma^{-1} ) (y_{ub'})_+ = \gamma^{-1} \sum_{y_{ub'}<0 }  y_{ub'} + (1+\gamma^{-1} ) \sum_{y_{ub'}>0 } (y_{ub'})_+ \\
>&  \gamma^{-1} \sum_{y_{ub'}<0 }  y_{ub'} + (1+\gamma^{-1} )(-1) \gamma^{-1/2} \sum_{y_{ub'}<0 }  y_{ub'}\\
=& (\gamma^{-1} - (1+\gamma^{-1} ) \gamma^{-1/2} ) \sum_{y_{ub'}<0 }  y_{ub'} \label{eq:lowerboundgamma2}
}

Also note that all terms are non-negative, so that replacing positive terms in the divisor by smaller positive ones yields an upper bound.  We obtain:
\eqn{
&\sum_{b: Y_{+}(b)}  \sum_{u=1} v_u Att( g_u , z_b)\\
\le &  \sum_{b: Y_{+}(b)}\sum_{u : v_u>0, y_{ub}>0 } v_u \frac{   (1+\gamma^{-1})(y_{ub})_+ }{ \mathcolor{orange}{(1+\gamma^{-1} - \gamma^{-1/2} )\sum_{b'}   (y_{ub'})_+} } 
\\
+& \sum_{b: Y_{+}(b)}\sum_{u : v_u<0,y_{ub}<0  } v_u \frac{ \gamma^{-1} y_{ub} }{\mathcolor{orange}{(\gamma^{-1} - (1+\gamma^{-1} ) \gamma^{-1/2} ) \sum_{y_{ub'}<0 }  y_{ub'}} } \\
=&  \sum_{b: Y_{+}(b)}\sum_{u : v_u>0, y_{ub}>0 } v_u \frac{   (1+\gamma^{-1})(y_{ub})_+ }{ (1+\gamma^{-1} - \gamma^{-1/2} )\sum_{b'}   (y_{ub'})_+ } 
\\
+& \sum_{b: Y_{+}(b)}\sum_{u : v_u<0,y_{ub}<0  } v_u \frac{ \gamma^{-1} \mathcolor{orange}{(y_{ub})_{-}} }{(\gamma^{-1} - (1+\gamma^{-1} ) \gamma^{-1/2} ) \sum_{y_{ub'}<0 }  \mathcolor{orange}{(y_{ub'})_{-}} }
}
Next we use the trick that for $y_{ub}<0$ we have  $y_{ub}= (y_{ub})_{-}$, however terms $(y_{ub})_{-}$ can be summed over all $b$ because for those where $y_{ub}>0$ it would be just zero: $(y_{ub})_{-}=0$. 

The same idea holds for $y_{ub}>0$ and $(y_{ub})_{+}$.

Therefore we can replace $\sum_{u : v_u<0,y_{ub}<0  }$ by $\sum_{u : v_u<0  }$ and $\sum_{u : v_u>0,y_{ub}>0  }$ by $\sum_{u : v_u>0  }$:
\eqn{
= & \sum_{b: Y_{+}(b)}\sum_{\mathcolor{orange}{u : v_u>0 }} v_u \frac{   (1+\gamma^{-1})(y_{ub})_+ }{ (1+\gamma^{-1} - \gamma^{-1/2} )\sum_{b'}   (y_{ub'})_+ } 
\\
+& \sum_{b: Y_{+}(b)}\sum_{\mathcolor{orange}{u : v_u<0}  } v_u \frac{ \gamma^{-1} (y_{ub})_{-} }{(\gamma^{-1} - (1+\gamma^{-1} ) \gamma^{-1/2} ) \sum_{b' }  (y_{ub'})_{-} }
}
Now all terms are non-negative [note that $\gamma >1$, so $1-\gamma^{-1/2}>0$ and that 

$(\gamma^{-1} - (1+\gamma^{-1} ) \gamma^{-1/2} ) = \gamma^{-1}( 1-(1+\gamma^{-1})\gamma^{1/2}) <0$] 

so that we can upper bound by increasing the sum from $\sum_{b: Y_{+}(b)}$ to $\sum_{b}$:
\eqn{
\le & \sum_{b}\sum_{u : v_u>0 } v_u \frac{   (1+\gamma^{-1})(y_{ub})_+ }{ (1+\gamma^{-1} - \gamma^{-1/2} )\sum_{b'}   (y_{ub'})_+ } 
\\
+& \sum_{b}\sum_{u : v_u<0  } v_u \frac{ \gamma^{-1} (y_{ub})_{-} }{(\gamma^{-1} - (1+\gamma^{-1} ) \gamma^{-1/2} ) \sum_{b' }  (y_{ub'})_{-} }\\
=&  \sum_{u : v_u>0  } v_u \frac{1+\gamma^{-1}}{1+\gamma^{-1} - \gamma^{-1/2}}  \sum_{b} \frac{(y_{ub})_+}{\sum_{b'}   (y_{ub'})_+}\\
+& \sum_{u : v_u<0  } v_u \frac{\gamma^{-1}}{(\gamma^{-1} - (1+\gamma^{-1} ) \gamma^{-1/2} )}\sum_{b}\frac{  (y_{ub})_{-} }{ \sum_{b' }  (y_{ub'})_{-} }\\
=&  \sum_{u : v_u>0  } v_u \frac{1+\gamma^{-1}}{1+\gamma^{-1} - \gamma^{-1/2}} \\
+&  \sum_{u : v_u<0  } v_u \frac{\gamma^{-1}}{(\gamma^{-1} - (1+\gamma^{-1} ) \gamma^{-1/2} )}\\
=& \sum_{u : v_u>0  } v_u \frac{1+\gamma} {1+\gamma - \gamma^{1/2}} \\
+&  \sum_{u : v_u<0  } v_u \underbrace{\frac{1}{(1 - (1+\gamma^{-1} ) \gamma^{1/2} )}}_{<0}
}
Now we can plug in the induction assumption
\eqn{
\le &  2^{t-2} \frac{1+\gamma}{1+\gamma - \gamma^{1/2}}  b(\gamma)^{t-2} \frac{1+\gamma}{1+\gamma - \gamma^{1/2}} \\
+ &2^{t-2}\frac{ 1}{ 1 -\gamma^{1/2}}b(\gamma)^{t-2} \frac{1}{(1 - (1+\gamma^{-1} ) \gamma^{1/2} )}\\
=& 2^{t-2} \frac{1+\gamma}{1+\gamma - \gamma^{1/2}}  b(\gamma)^{t-2} \frac{1+\gamma}{1+\gamma - \gamma^{1/2}} \\
+ &2^{t-2}\frac{ 1}{ \gamma^{1/2}-1}b(\gamma)^{t-2} \frac{-1}{(1 - (1+\gamma^{-1} ) \gamma^{1/2} )}\\
}
Finally note
\eqn{
&\frac{-1}{(1 - (1+\gamma^{-1} ) \gamma^{1/2} )} = -\frac{\gamma^{1/2}}{(\gamma^{1/2} - (1+\gamma^{-1} ) \gamma )} = -\frac{\gamma^{1/2}}{(\gamma^{1/2} - (1+\gamma )  )}\\
=&\frac{\gamma^{1/2}}{ 1+\gamma -\gamma^{1/2}} \le \frac{1+\gamma}{ 1+\gamma -\gamma^{1/2}} \label{eq:funnyeqn1}
}
Therefore:
\eqn{
&\sum_{b: Y_{+}(b)}  \sum_{u=1}v_u Att( g_u , z_b) \\
\le& 2^{t-2} \frac{1+\gamma}{1+\gamma - \gamma^{1/2}}  b(\gamma)^{t-2} \frac{1+\gamma}{1+\gamma - \gamma^{1/2}} \\
+ &2^{t-2}\frac{ 1}{ \gamma^{1/2}-1}b(\gamma)^{t-2} \frac{-1}{(1 - (1+\gamma^{-1} ) \gamma^{1/2} )}\\
\le & 2^{t-2} \frac{1+\gamma}{1+\gamma - \gamma^{1/2}}  b(\gamma)^{t-2}b(\gamma)  \\
+ & 2^{t-2}b(\gamma) b(\gamma)^{t-2}\frac{1+\gamma}{ 1+\gamma -\gamma^{1/2}}\\
&= 2^{t-1} \frac{1+\gamma}{ 1+\gamma -\gamma^{1/2}}b(\gamma)^{t-1} 
}
which proves the induction claim for the positive upper bound.

For the lower bound we can derive in similar spirit:
\eqn{
&\sum_{b: Y_{-}(b)}  \sum_{u=1} v_u Att( g_u , z_b)\\
 = & \sum_{b: Y_{-}(b)} \sum_{u : v_u>0, y_{ub}>0 } v_u \frac{   (1+\gamma^{-1})(y_{ub})_+ }{\sum_{b'}  \gamma^{-1}y_{ub'} + (y_{ub'})_+ } 
+ \sum_{u : v_u>0, y_{ub}<0  } v_u \frac{ \gamma^{-1} y_{ub} }{\sum_{b'}  \gamma^{-1}y_{ub'} + (y_{ub'})_+ } \\
+& \sum_{b: Y_{-}(b)}\sum_{u : v_u<0,y_{ub}>0 } v_u \frac{   (1+\gamma^{-1})(y_{ub})_+ }{\sum_{b'}  \gamma^{-1}y_{ub'} + (y_{ub'})_+ } 
+ \sum_{u : v_u<0,y_{ub}<0  } v_u \frac{ \gamma^{-1} y_{ub} }{\sum_{b'}  \gamma^{-1}y_{ub'} + (y_{ub'})_+ } \\
\ge & \sum_{b: Y_{-}(b)} 0
+ \sum_{u : v_u>0, y_{ub}<0  } v_u \frac{ \gamma^{-1} y_{ub} }{\sum_{b'}  \gamma^{-1}y_{ub'} + (y_{ub'})_+ } \label{eq:195} \\
+& \sum_{b: Y_{-}(b)}\sum_{u : v_u<0,y_{ub}>0 } v_u \frac{   (1+\gamma^{-1})(y_{ub})_+ }{\sum_{b'}  \gamma^{-1}y_{ub'} + (y_{ub'})_+ } 
+ 0  \label{eq:196}
}
We will use two inequalities derived in equations \eqref{eq:lowerboundgamma1} and \eqref{eq:lowerboundgamma2}.

For the upper line \eqref{eq:195} we will use
\eqn{
\sum_{b'}  \gamma^{-1}y_{ub'} + (y_{ub'})_+ > (\gamma^{-1} - (1+\gamma^{-1} ) \gamma^{-1/2} ) \sum_{y_{ub'}<0 }  y_{ub'} 
}
which works because in \eqref{eq:195}  we have $v_u>0$ and $\gamma^{-1}y_{ub}<0$.

For the lower line \eqref{eq:196}  we will use
\eqn{
\sum_{b'}  \gamma^{-1}y_{ub'} + (y_{ub'})_+ >
(1+\gamma^{-1} - \gamma^{-1/2} ) \sum_{b'} (y_{ub'})_+
}
which works because in \eqref{eq:196}  we have $v_u<0$ and $(1+\gamma^{-1})(y_{ub})_{+}>0$.

Note that the terms in \eqref{eq:195} and \eqref{eq:196} are non-positive, so that replacing positive terms in the divisor by smaller positive ones yields a lower bound. Therefore:
\eqn{
&\sum_{b: Y_{-}(b)}  \sum_{u=1} v_u Att( g_u , z_b)\\
\ge &\sum_{b: Y_{-}(b)}  \sum_{u : v_u>0, y_{ub}<0  } v_u \frac{ \gamma^{-1} y_{ub} }{ \mathcolor{orange}{\sum_{b'}  \gamma^{-1}y_{ub'} + (y_{ub'})_+ }} \\
+& \sum_{b: Y_{-}(b)}\sum_{u : v_u<0,y_{ub}>0 } v_u \frac{   (1+\gamma^{-1})(y_{ub})_+ }{\mathcolor{orange}{\sum_{b'}  \gamma^{-1}y_{ub'} + (y_{ub'})_+ }} \\
\ge &\sum_{b: Y_{-}(b)}  \sum_{u : v_u>0, y_{ub}<0  } v_u \frac{ \gamma^{-1} y_{ub} }{\mathcolor{orange}{(\gamma^{-1} - (1+\gamma^{-1} ) \gamma^{-1/2} ) \sum_{y_{ub'}<0 }  y_{ub'}} } \\
+& \sum_{b: Y_{-}(b)}\sum_{u : v_u<0,y_{ub}>0 } v_u \frac{   (1+\gamma^{-1})(y_{ub})_+ }{\mathcolor{orange}{(1+\gamma^{-1} - \gamma^{-1/2} ) \sum_{b'} (y_{ub'})_+} } \\
=&\frac{\gamma^{-1}}{(\gamma^{-1} - (1+\gamma^{-1} ) \gamma^{-1/2} )}\sum_{b: Y_{-}(b)}  \sum_{u : v_u>0, y_{ub}<0  } v_u \frac{  y_{ub} }{ \sum_{y_{ub'}<0 }  y_{ub'} } \\
+& \frac{1+\gamma^{-1}}{1+\gamma^{-1} - \gamma^{-1/2}}\sum_{b: Y_{-}(b)} \sum_{u : v_u<0,y_{ub}>0 } v_u \frac{(y_{ub})_+}{\sum_{b'}(y_{ub'})_+}\\
=& \frac{\gamma^{-1}}{(\gamma^{-1} - (1+\gamma^{-1} ) \gamma^{-1/2} )}\sum_{b: Y_{-}(b)}  \sum_{u : v_u>0, y_{ub}<0  } v_u \frac{  \mathcolor{orange}{(y_{ub})_{-}} }{ \sum_{y_{ub'}<0 }  \mathcolor{orange}{(y_{ub'})_{-}} } \\
+& \frac{1+\gamma^{-1}}{1+\gamma^{-1} - \gamma^{-1/2}}\sum_{b: Y_{-}(b)} \sum_{u : v_u<0,y_{ub}>0 } v_u \frac{(y_{ub})_+}{\sum_{b'}(y_{ub'})_+}
}
Now we use the same trick as for the positive upper bound which allows us to drop the conditioning in the $ \sum_{u : v_u<0,y_{ub}>0 }$ and  $ \sum_{u : v_u>0,y_{ub}<0 }$ on the sign of $y_{ub}$ - because the for the additional terms $(y_{ub})_{+}=0 $ and  $(y_{ub})_{-}=0 $ respectively:
\eqn{
=& \frac{\gamma^{-1}}{(\gamma^{-1} - (1+\gamma^{-1} ) \gamma^{-1/2} )}\sum_{b: Y_{-}(b)}  \sum_{\mathcolor{orange}{u : v_u>0}  } v_u \frac{  (y_{ub})_{-} }{ \sum_{b' }  (y_{ub'})_{-} } \\
+& \frac{1+\gamma^{-1}}{1+\gamma^{-1} - \gamma^{-1/2}}\sum_{b: Y_{-}(b)} \sum_{\mathcolor{orange}{u : v_u<0}} v_u \frac{(y_{ub})_+}{\sum_{b'}(y_{ub'})_+}\\
\ge& \frac{\gamma^{-1}}{(\gamma^{-1} - (1+\gamma^{-1} ) \gamma^{-1/2} )}\sum_{b}  \sum_{u : v_u>0  } v_u \frac{  (y_{ub})_{-} }{ \sum_{b' }  (y_{ub'})_{-} } \\
+& \frac{1+\gamma^{-1}}{1+\gamma^{-1} - \gamma^{-1/2}}\sum_{b} \sum_{u : v_u<0} v_u \frac{(y_{ub})_+}{\sum_{b'}(y_{ub'})_+}\\
=& \frac{\gamma^{-1}}{(\gamma^{-1} - (1+\gamma^{-1} ) \gamma^{-1/2} )}  \sum_{u : v_u>0  } v_u \frac{ \sum_{b} (y_{ub})_{-} }{ \sum_{b' }  (y_{ub'})_{-} } \\
+& \frac{1+\gamma^{-1}}{1+\gamma^{-1} - \gamma^{-1/2}} \sum_{u : v_u<0} v_u \frac{\sum_{b}(y_{ub})_+}{\sum_{b'}(y_{ub'})_+}\\
=&\frac{\gamma^{-1}}{(\gamma^{-1} - (1+\gamma^{-1} ) \gamma^{-1/2} )}  \sum_{u : v_u>0  } v_u  
+ \frac{1+\gamma^{-1}}{1+\gamma^{-1} - \gamma^{-1/2}} \sum_{u : v_u<0} v_u \\
=&\frac{1}{(1 - (1+\gamma^{-1} ) \gamma^{1/2} )} \sum_{u : v_u>0  } v_u   + \frac{1+\gamma}{1+\gamma - \gamma^{1/2}} \sum_{u : v_u<0} v_u 
}

Here we can plug in again the induction assumption to obtain
\eqn{
\ge& \frac{1}{(1 - (1+\gamma^{-1} ) \gamma^{1/2} )}  2^{t-2} \frac{1+\gamma}{1+\gamma - \gamma^{1/2}}  b(\gamma)^{t-2} + \frac{1+\gamma}{1+\gamma - \gamma^{1/2}}2^{t-2}\frac{ 1}{ 1 -\gamma^{1/2}}b(\gamma)^{t-2}\\
\ge &\frac{1}{(1 -  \gamma^{1/2} )}2^{t-2} \frac{1+\gamma}{1+\gamma - \gamma^{1/2}}  b(\gamma)^{t-2} + \frac{1+\gamma}{1+\gamma - \gamma^{1/2}}2^{t-2}\frac{ 1}{ 1 -\gamma^{1/2}}b(\gamma)^{t-2}\\
=&  \frac{1}{(1 -  \gamma^{1/2} )}2^{t-1} b(\gamma)^{t-2}  \frac{1+\gamma}{1+\gamma - \gamma^{1/2}} \\
\ge & \frac{1}{(1 -  \gamma^{1/2} )}2^{t-1} b(\gamma)^{t-2} b(\gamma)\\
=&  \frac{1}{(1 -  \gamma^{1/2} )}2^{t-1} b(\gamma)^{t-1}
}
This concludes the proof for the lower bound


\newpage

\section{Convergence Statistics for for \texorpdfstring{LRP-$\beta$}{} and the gradient}\label{app:stats_grad}
\begin{figure}[h!]
    \centering
    \includegraphics[width=0.72\linewidth]{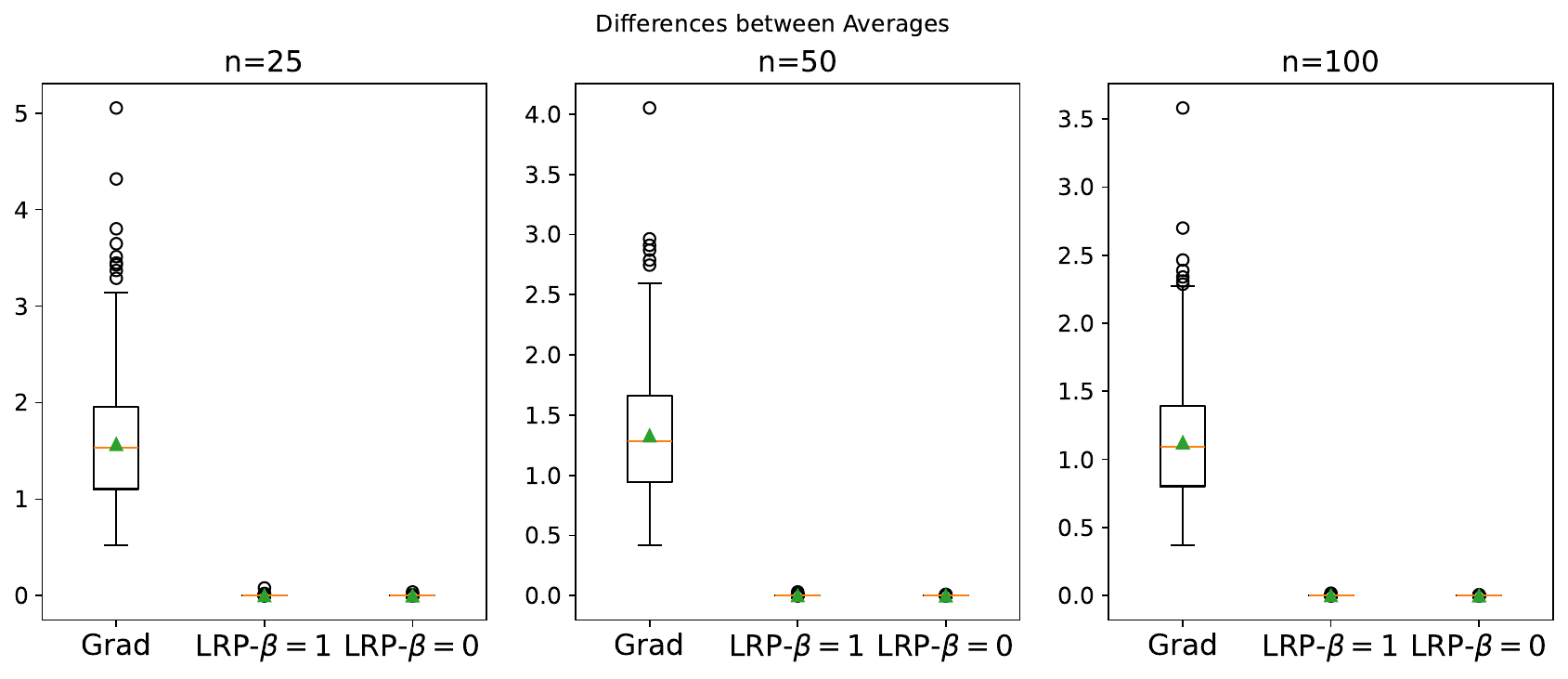}     \includegraphics[width=0.72\linewidth]{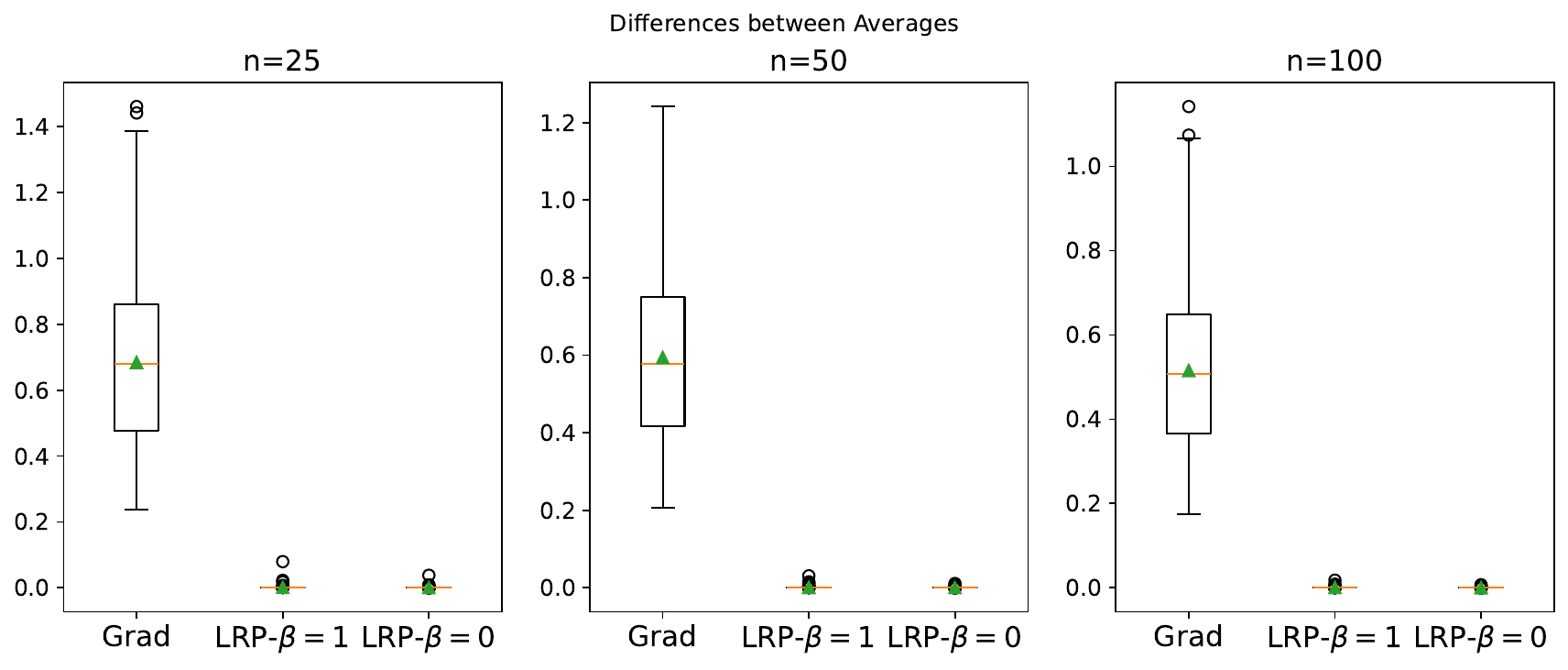}
    \includegraphics[width=0.72\linewidth]{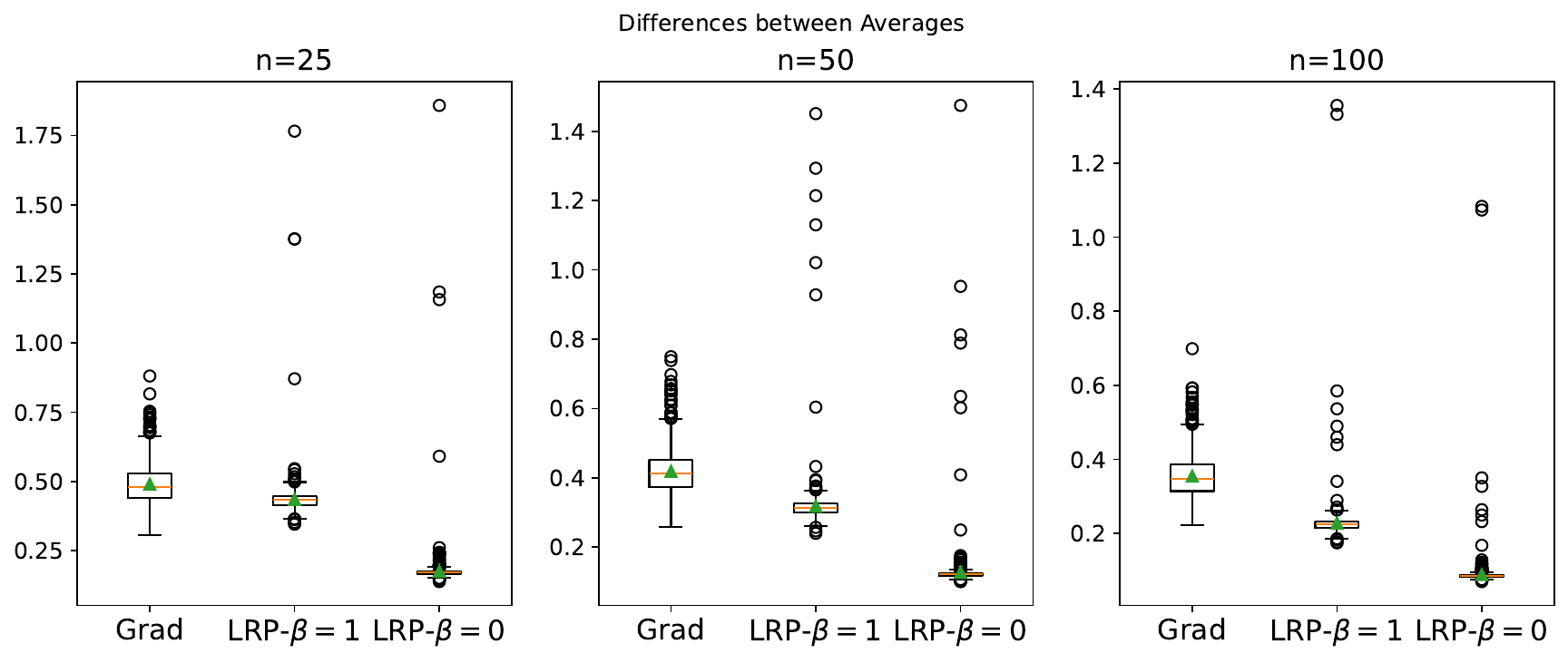}     \includegraphics[width=0.72\linewidth]{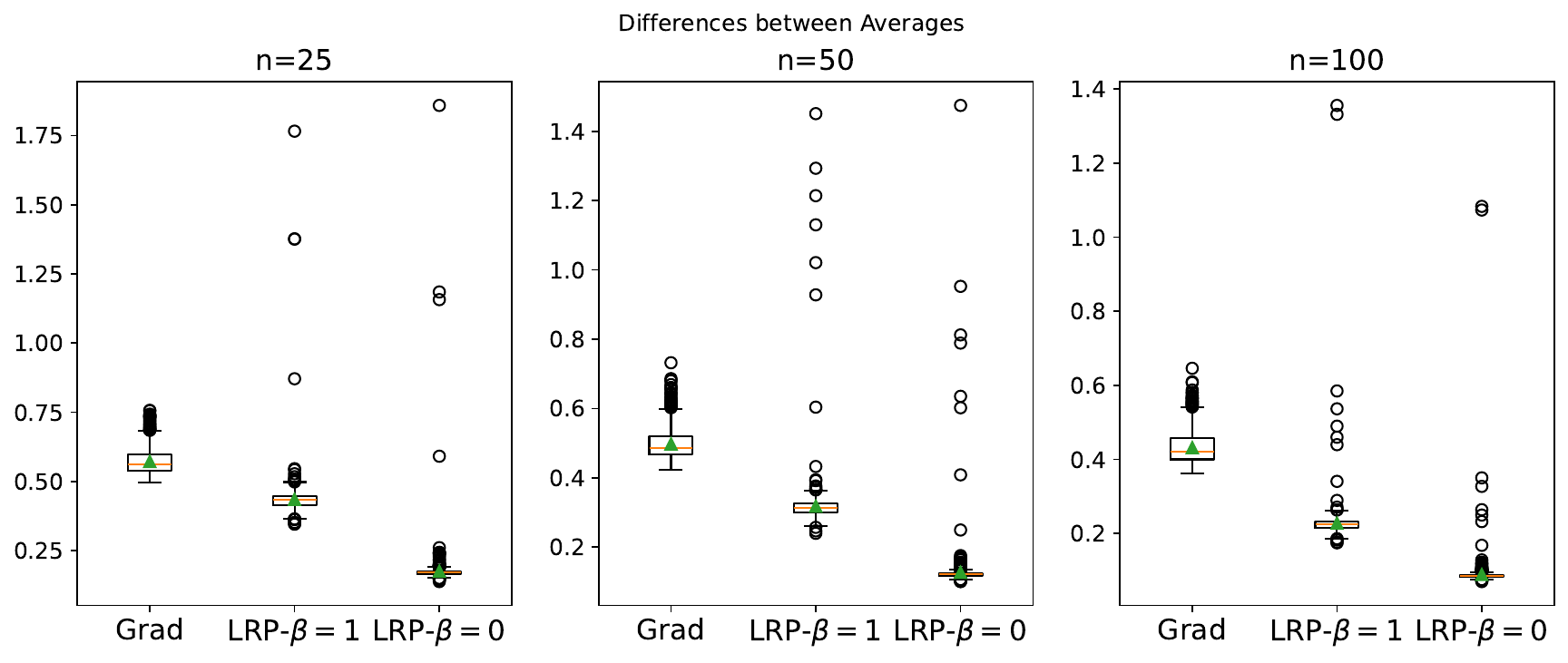}
    \caption{Convergence statistics for EfficientNet-V2-S. Lower is better. First row: no normalization, photometric augmentation. Second row: no normalization, noise augmentation. Third row: $\ell_2$-normalization, photometric augmentation. Fourth row: $\ell_2$-normalization, noise augmentation.}
    \label{fig:app:stats_grad_eff}
\end{figure}

\begin{figure}[htbp]
    \centering
    \includegraphics[width=0.72\linewidth]{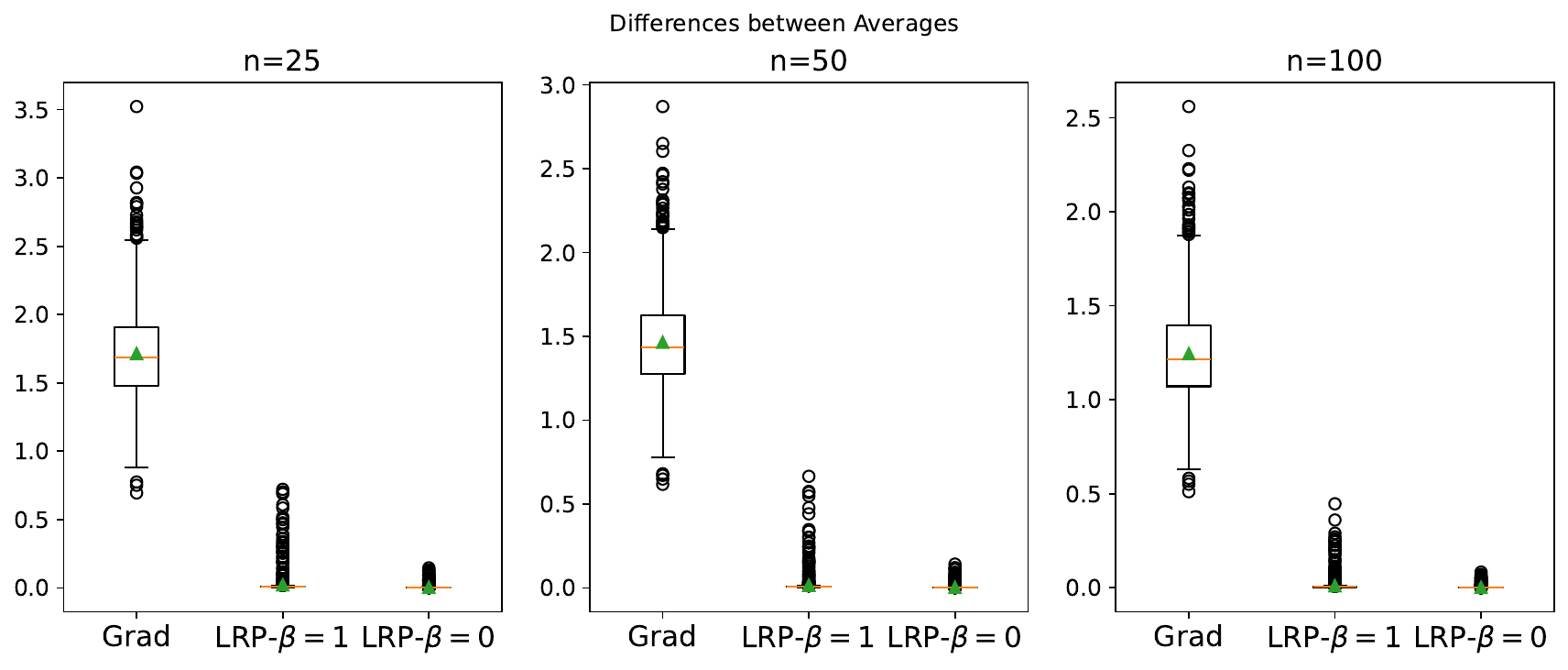}     \includegraphics[width=0.72\linewidth]{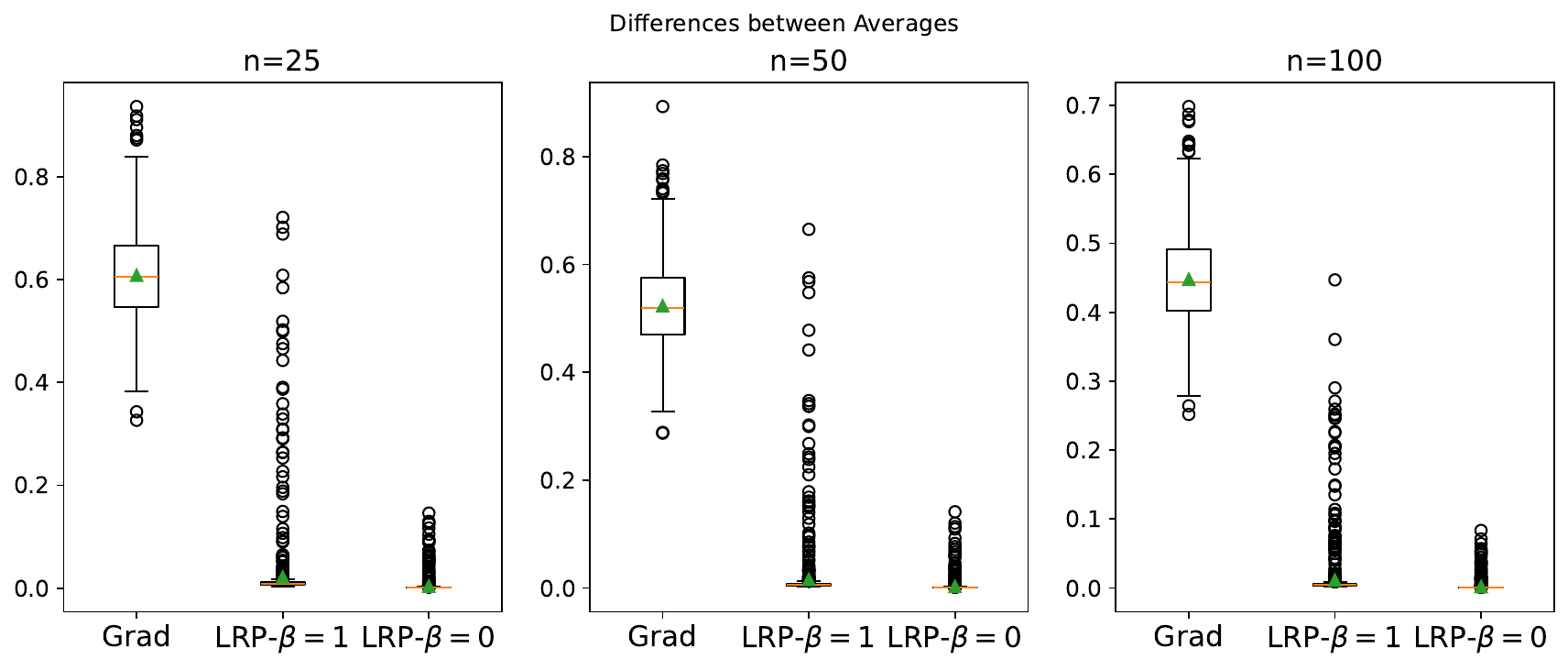}
    \includegraphics[width=0.72\linewidth]{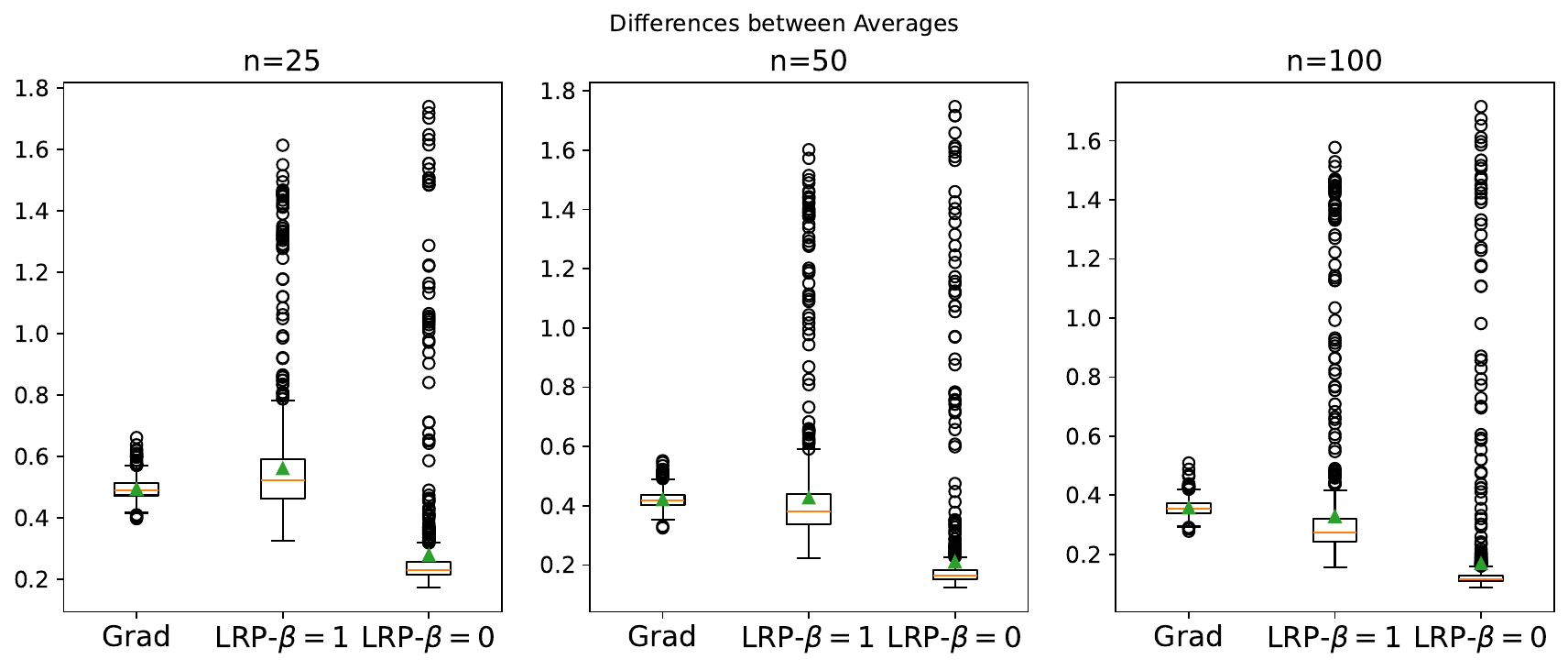}     \includegraphics[width=0.72\linewidth]{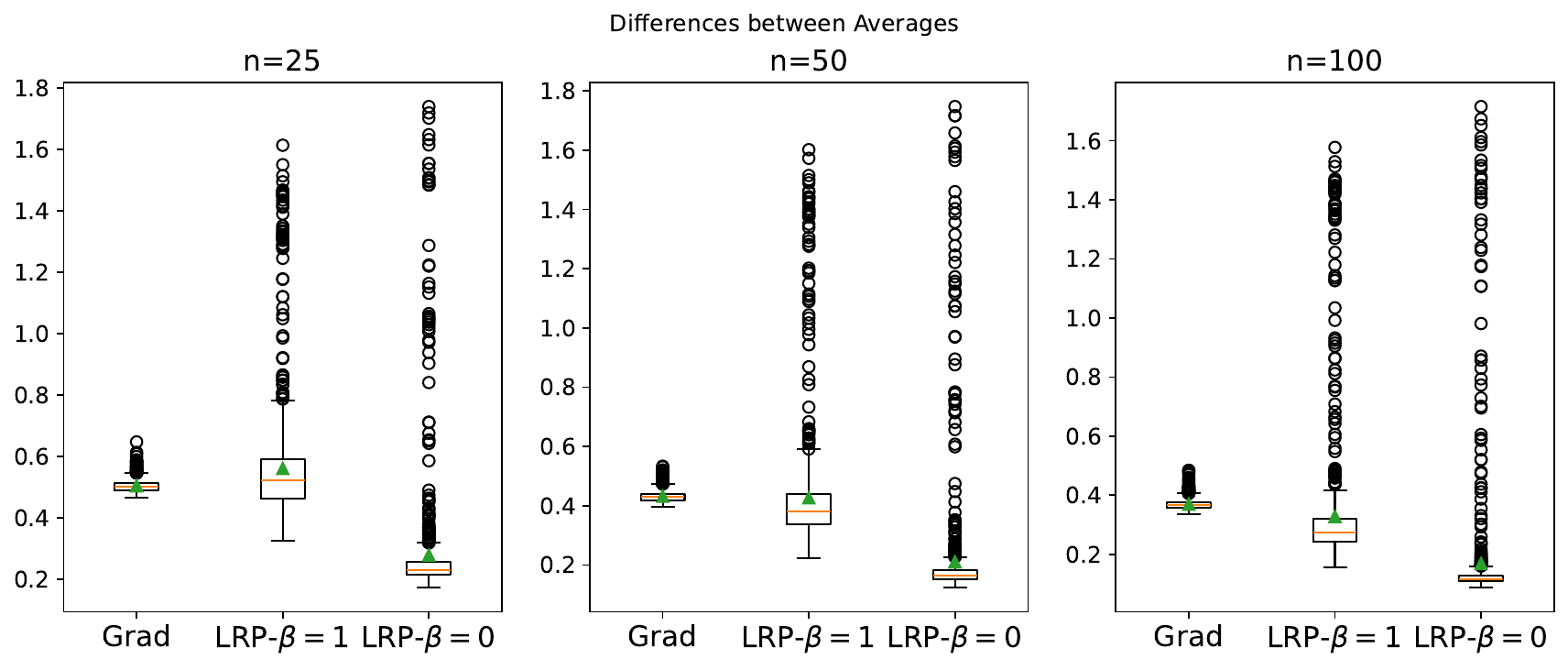}
    \caption{Convergence statistics for ResNet-50. Lower is better. First row: no normalization, photometric augmentation. Second row: no normalization, noise augmentation. third row: $\ell_2$-normalization, photometric augmentation. Fourth row: $\ell_2$-normalization, noise augmentation.}
    \label{fig:app:stats_grad_res}
\end{figure}

\section{Convergence Statistics for \texorpdfstring{LRP-$\beta$}{} and the gradient times input}\label{app:stats_gxi}
\begin{figure}[h!]
    \centering
    \includegraphics[width=0.72\linewidth]{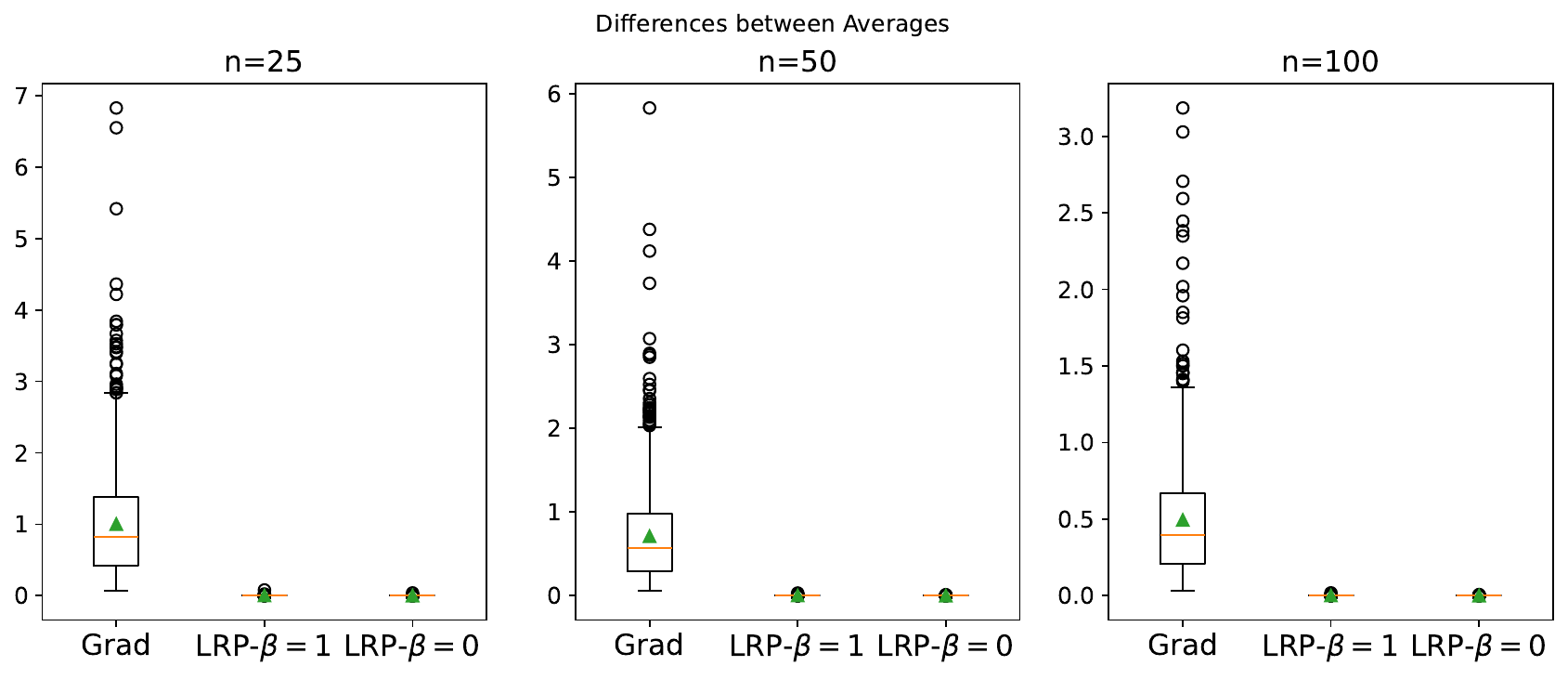}     \includegraphics[width=0.72\linewidth]{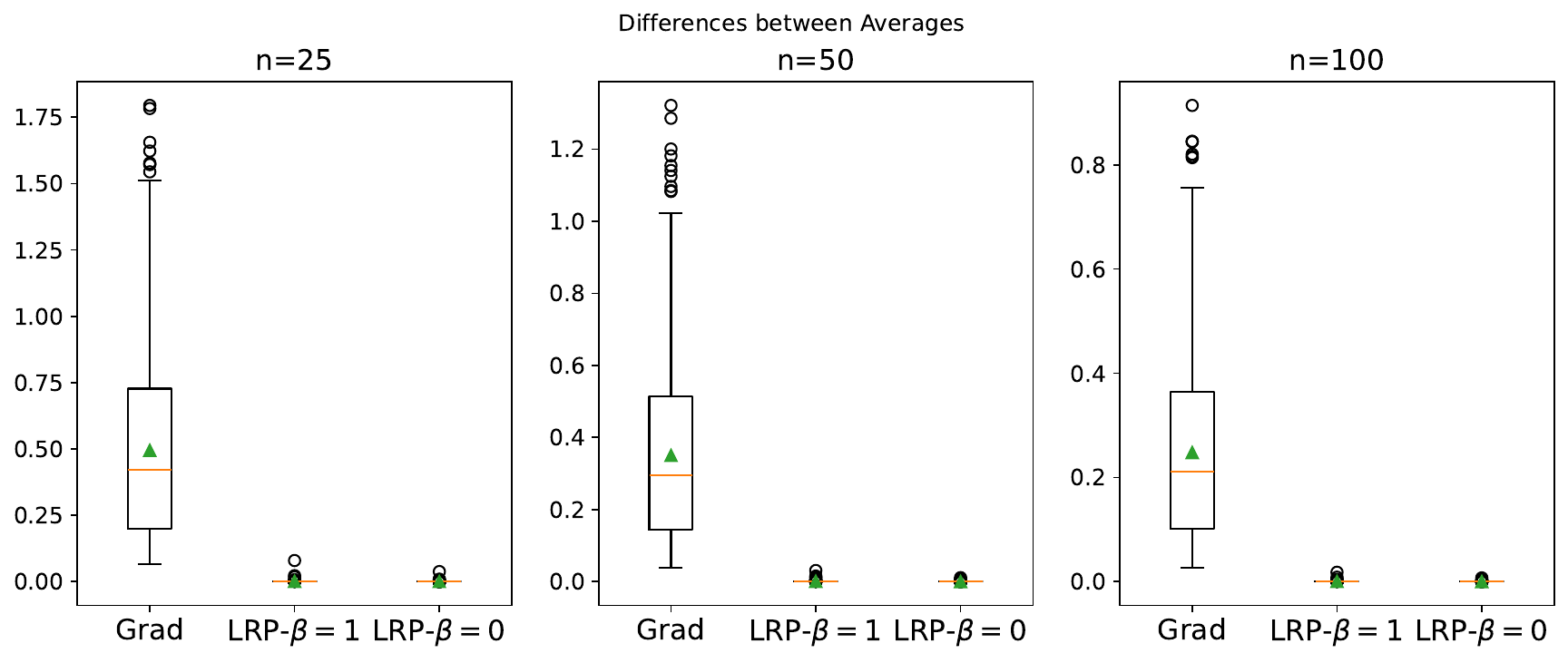}
    \includegraphics[width=0.72\linewidth]{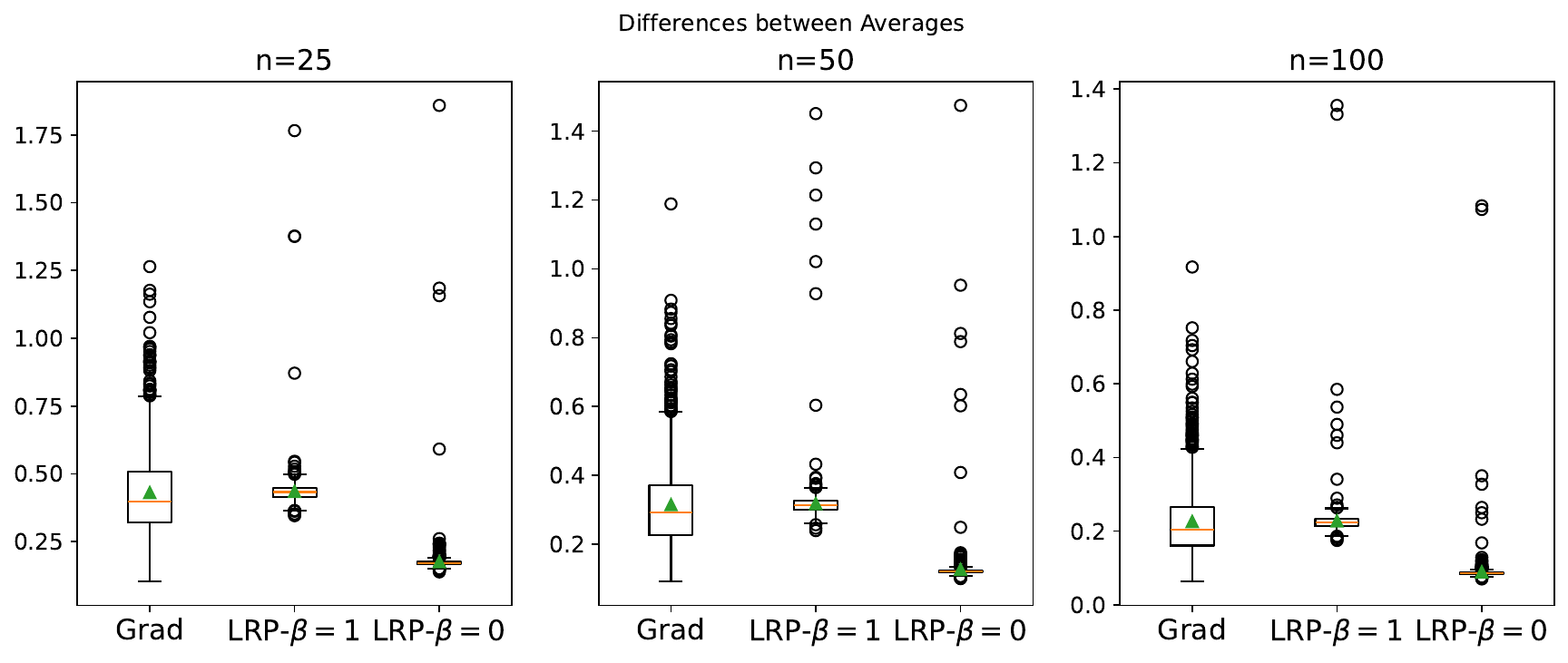}     \includegraphics[width=0.72\linewidth]{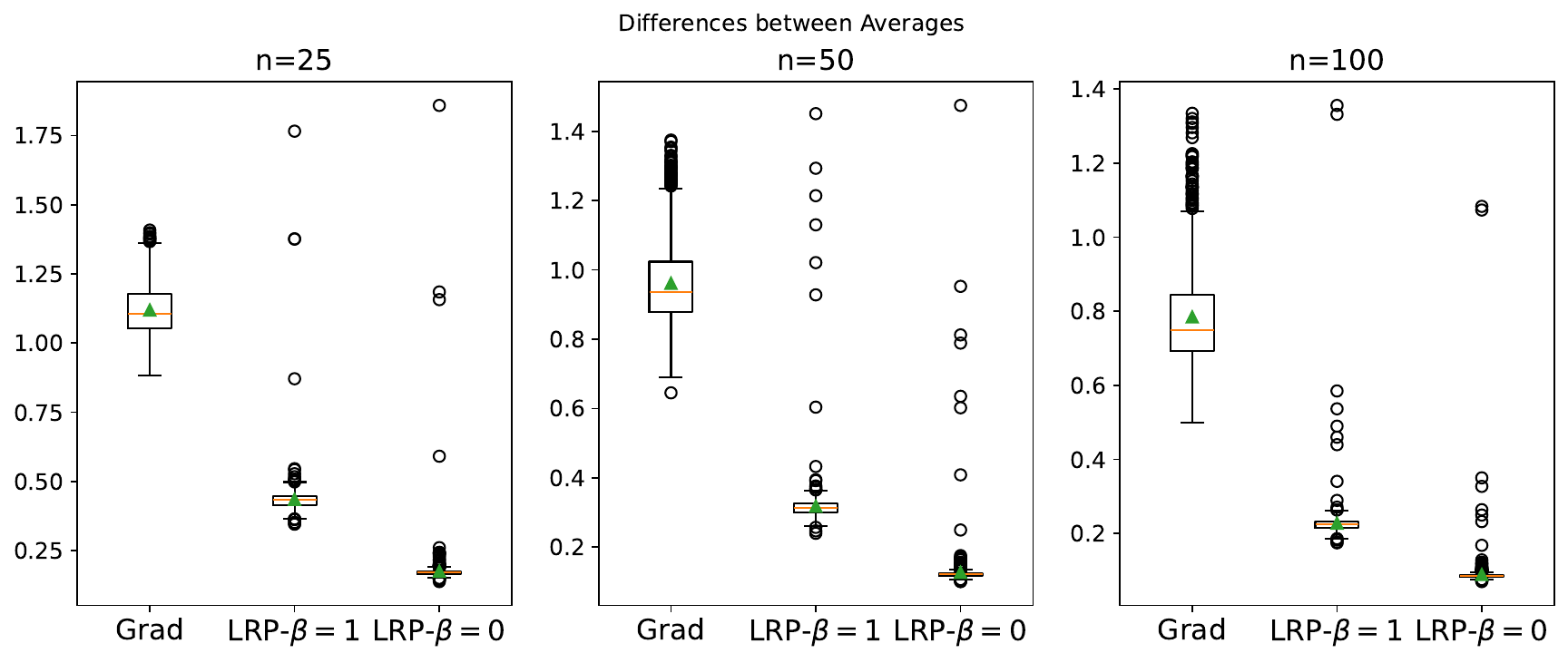}
    \caption{Convergence statistics for EfficientNet-V2-S. Lower is better. First row: no normalization, photometric augmentation. Second row: no normalization, noise augmentation. Third row: $\ell_2$-normalization, photometric augmentation. Fourth row: $\ell_2$-normalization, noise augmentation.}
    \label{fig:app:stats_gxi_eff}
\end{figure}

\begin{figure}[htbp]
    \centering
    \includegraphics[width=0.72\linewidth]{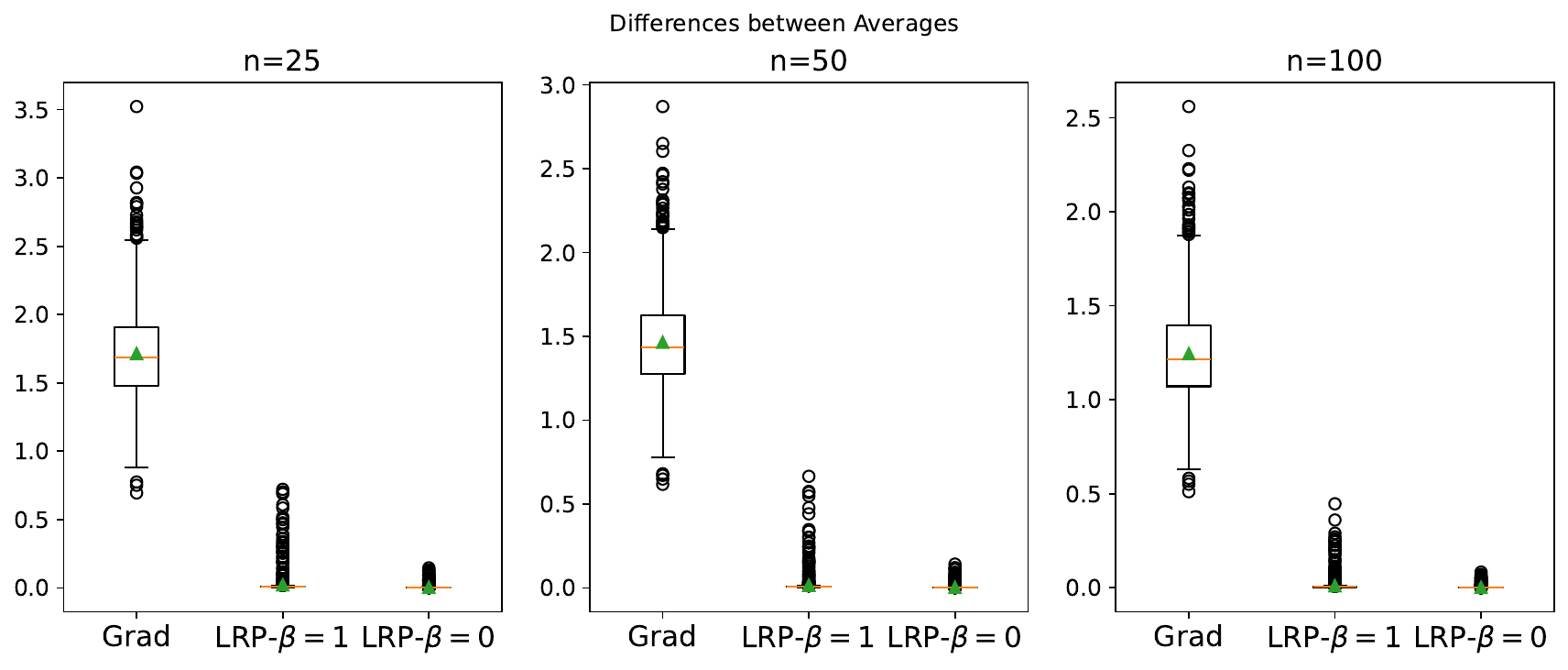}     \includegraphics[width=0.72\linewidth]{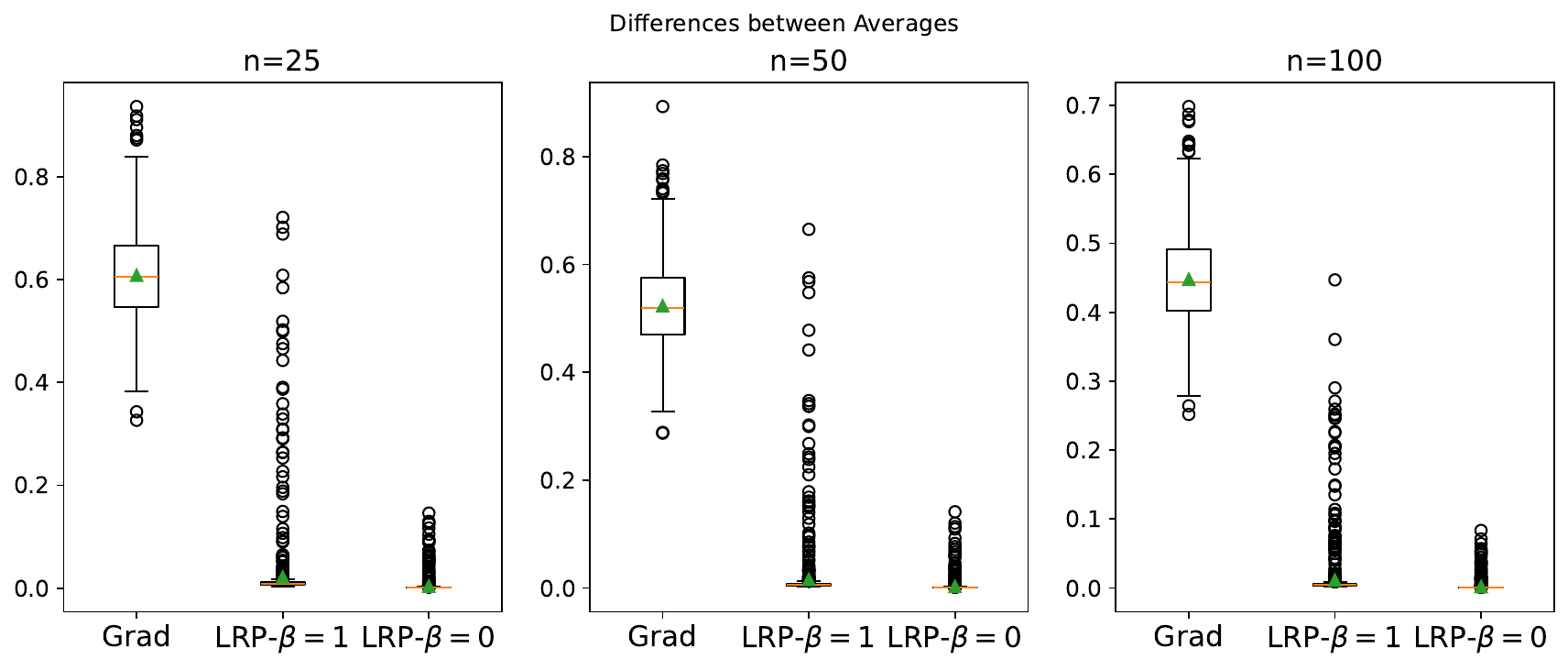}
    \includegraphics[width=0.72\linewidth]{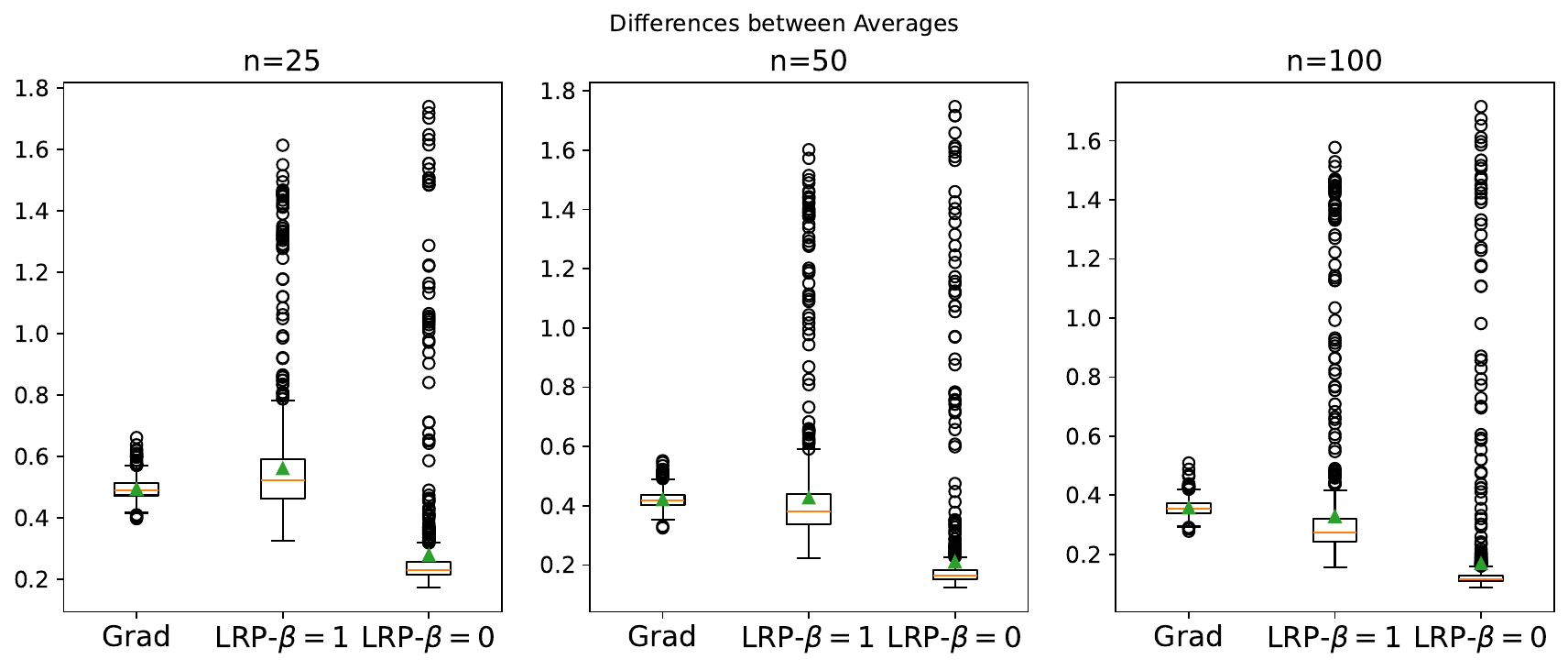}     \includegraphics[width=0.72\linewidth]{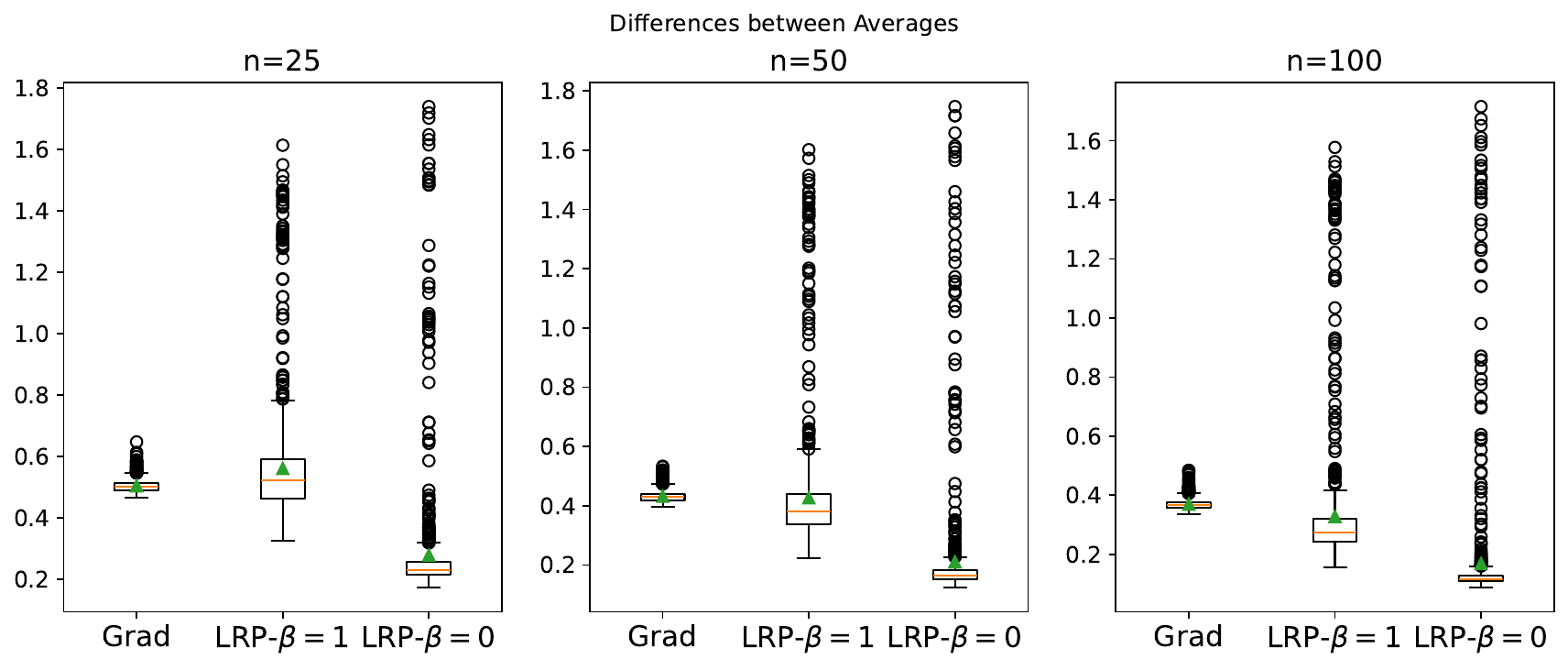}
    \caption{Convergence statistics for ResNet-50. Lower is better. First row: no normalization, photometric augmentation. Second row: no normalization, noise augmentation. third row: $\ell_2$-normalization, photometric augmentation. Fourth row: $\ell_2$-normalization, noise augmentation.}
    \label{fig:app:stats_gxi_res}
\end{figure}

\section{Convergence Statistics for for \texorpdfstring{LRP-$\gamma$}{} and the gradient}\label{app:stats_grad_gamma}

\begin{figure}[ht!]
    \centering
\includegraphics[width=0.72\linewidth]{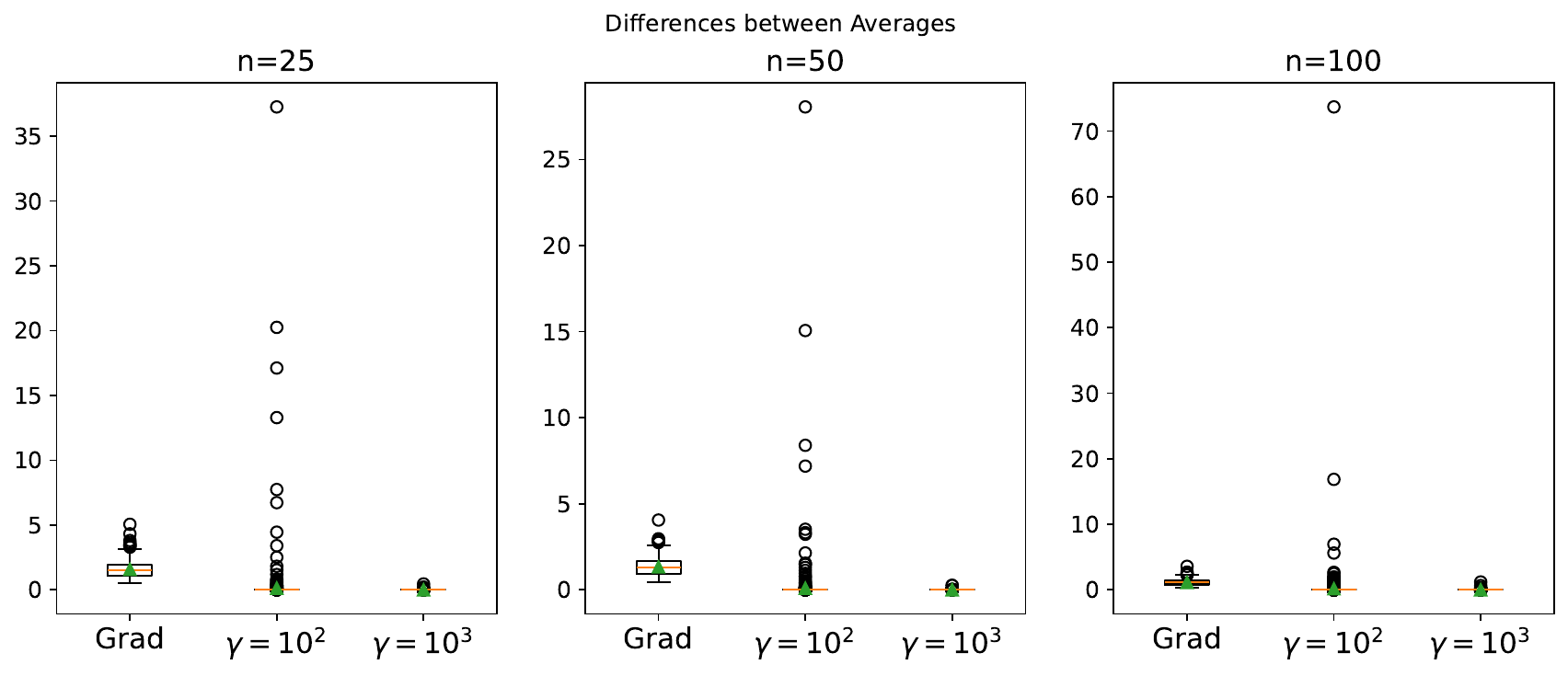}
\includegraphics[width=0.72\linewidth]{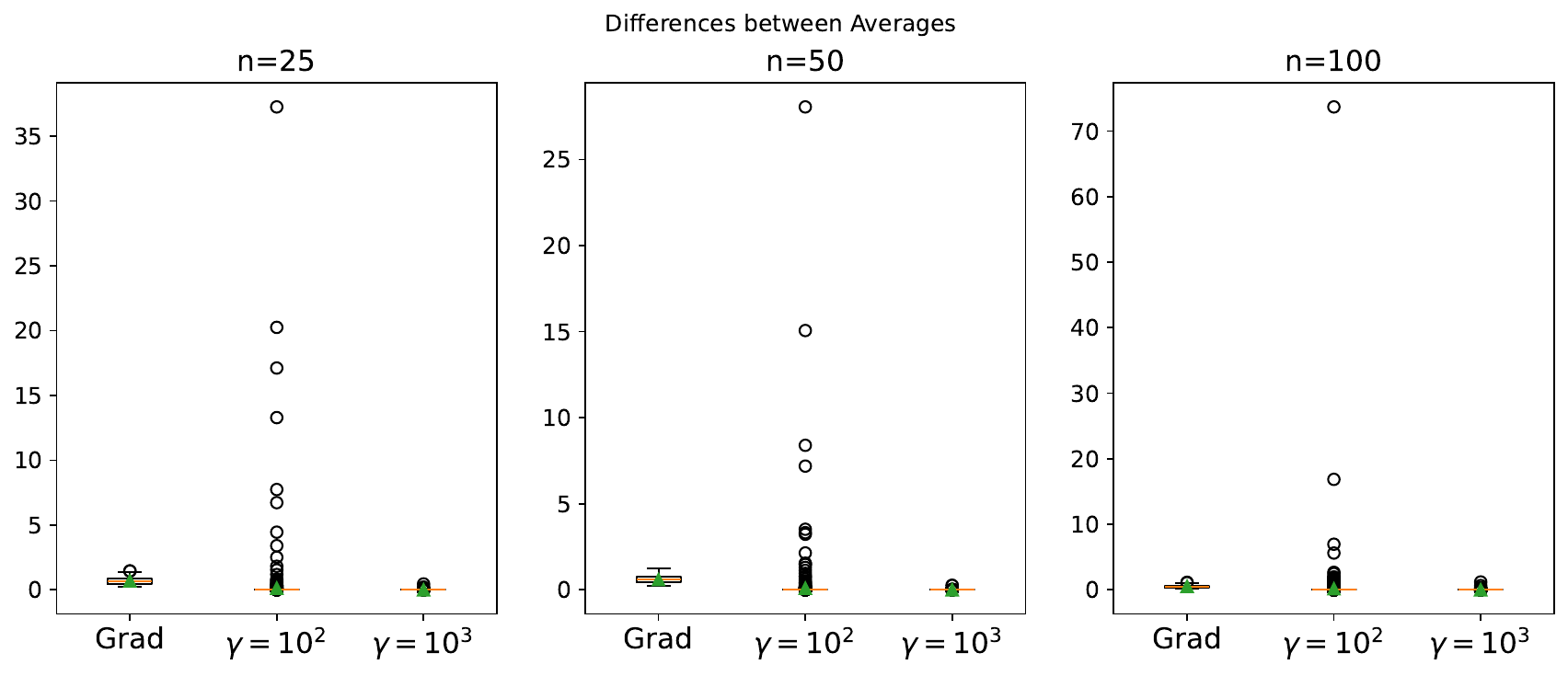}

\includegraphics[width=0.72\linewidth]{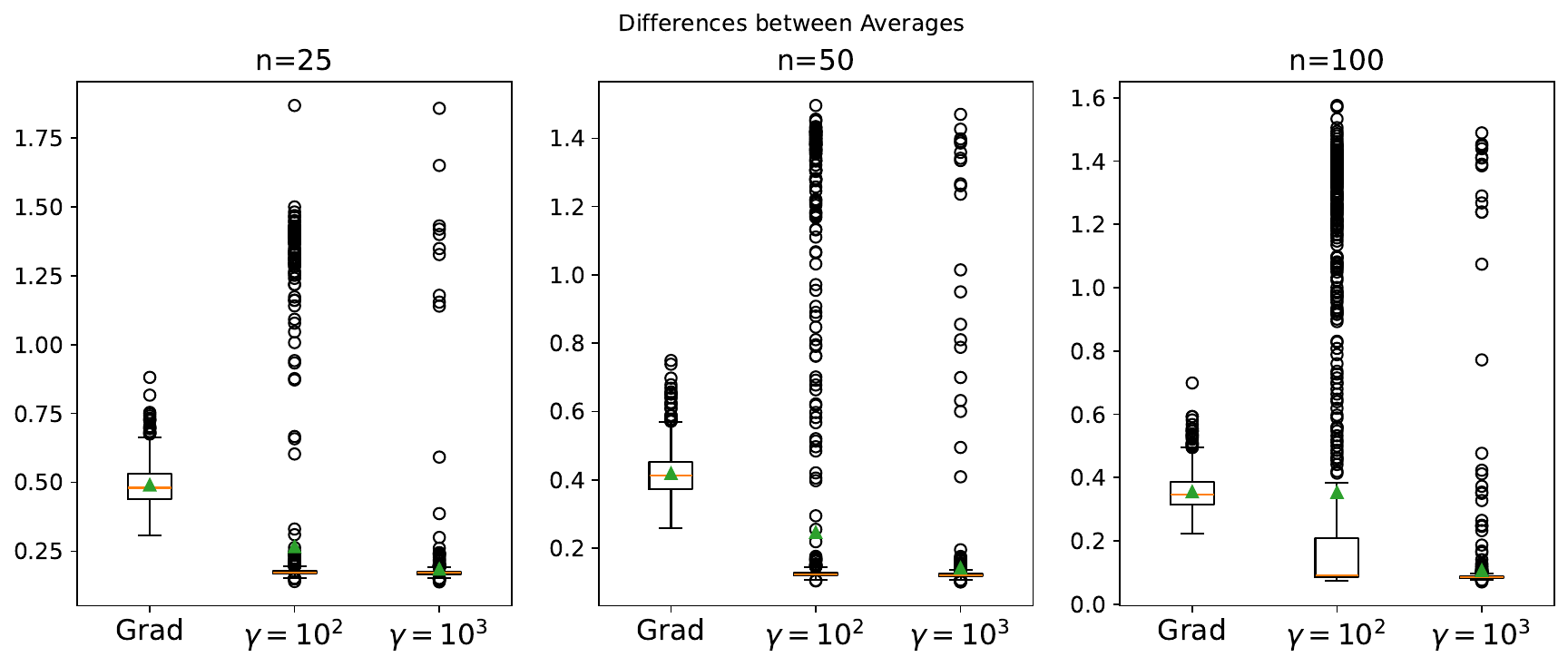}
\includegraphics[width=0.72\linewidth]{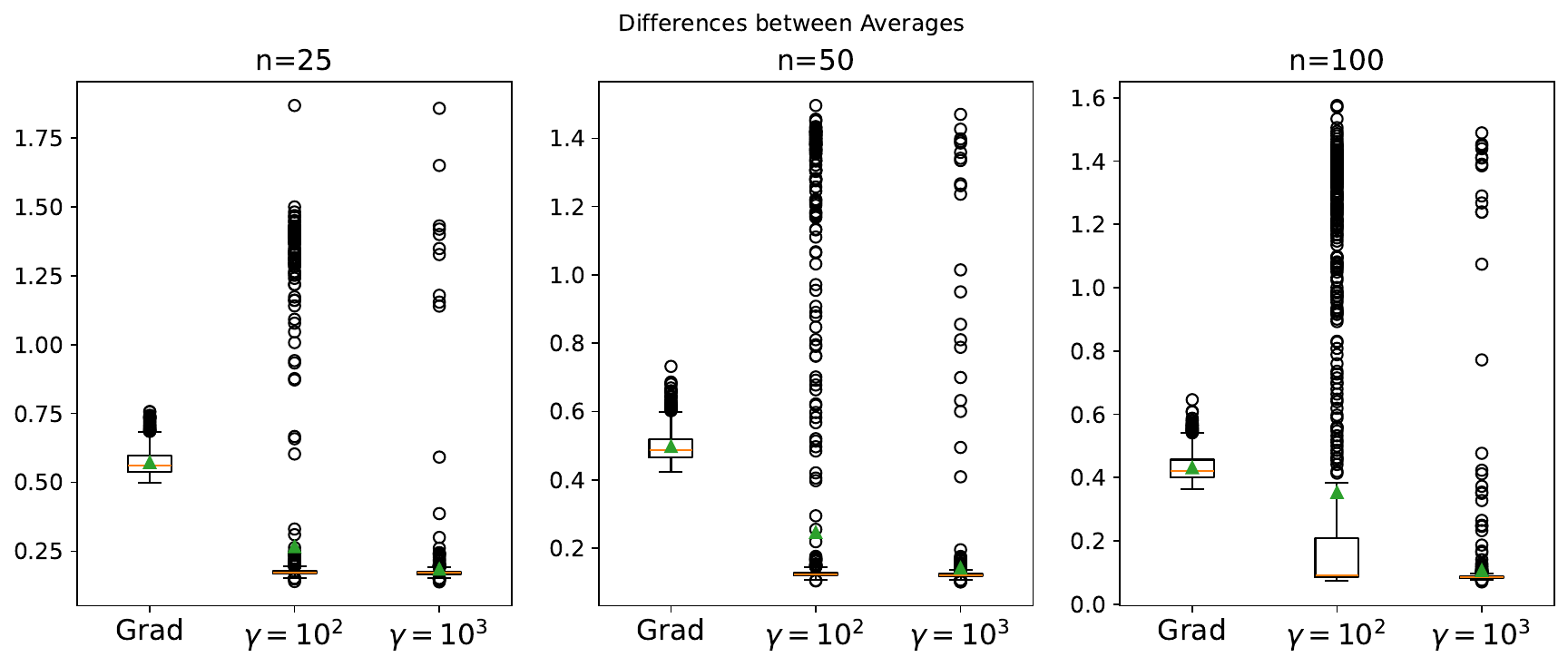}

    \caption{Convergence statistics for EfficientNet-V2-S. Lower is better. First row: no normalization, photometric augmentation. Second row: no normalization, noise augmentation. Third row: $\ell_2$-normalization, photometric augmentation. Fourth row: $\ell_2$-normalization, noise augmentation.}

\end{figure}

\begin{figure}[htbp]
    \centering
\includegraphics[width=0.72\linewidth]{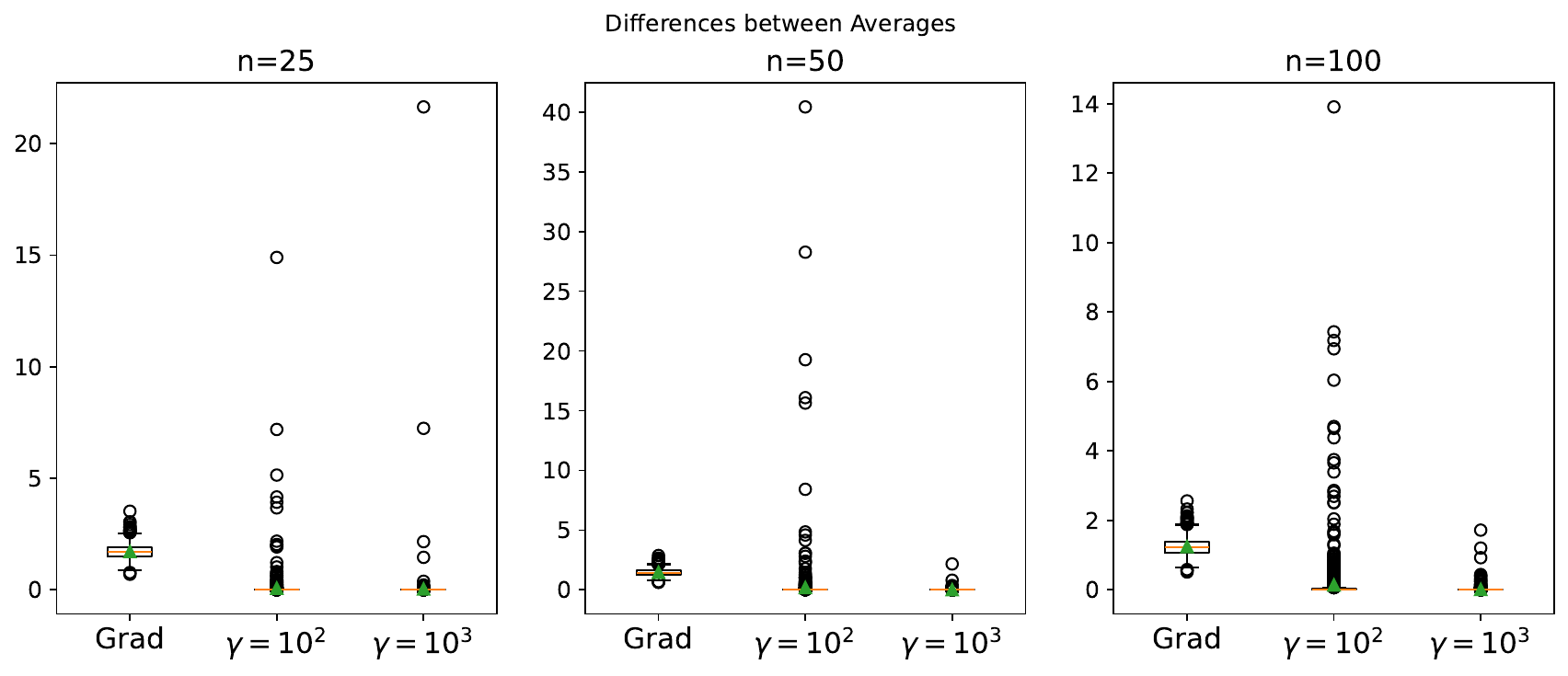}
\includegraphics[width=0.72\linewidth]{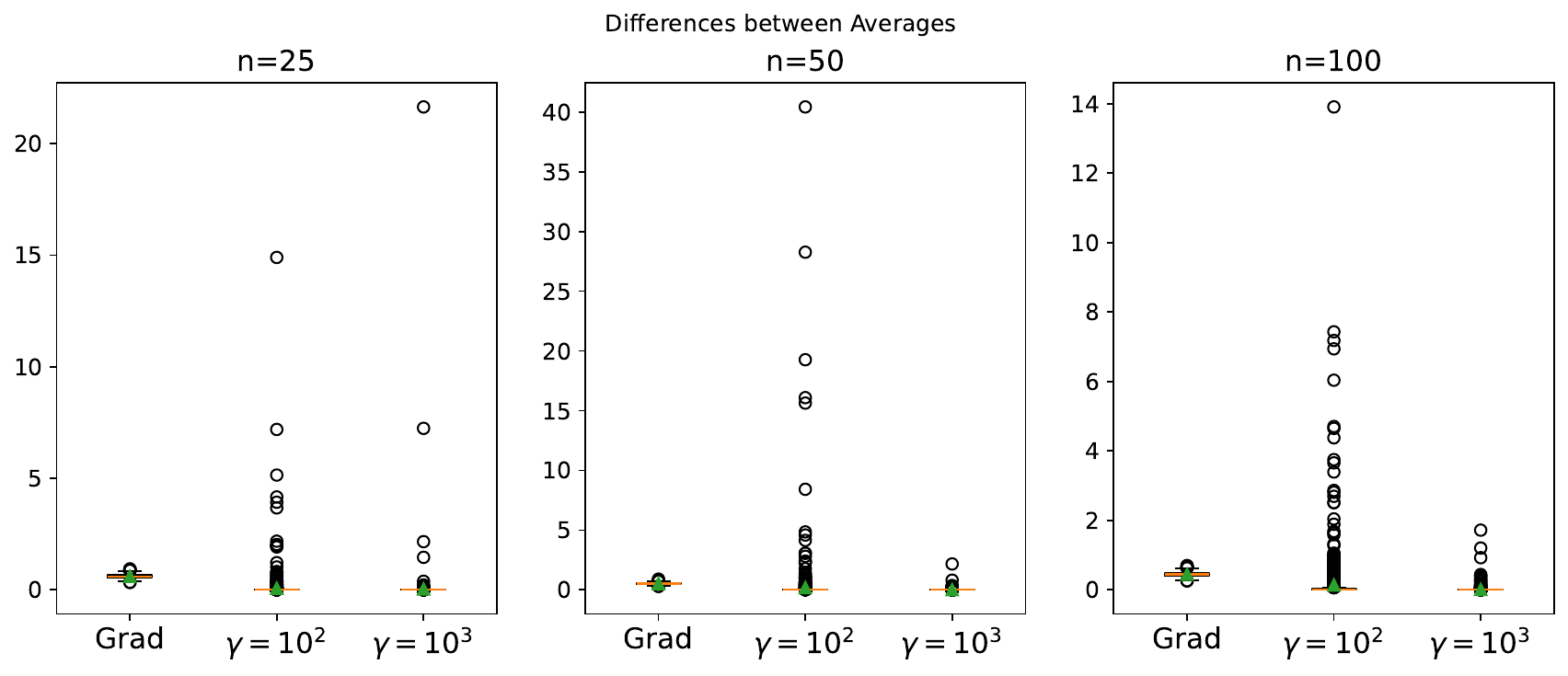}

\includegraphics[width=0.72\linewidth]{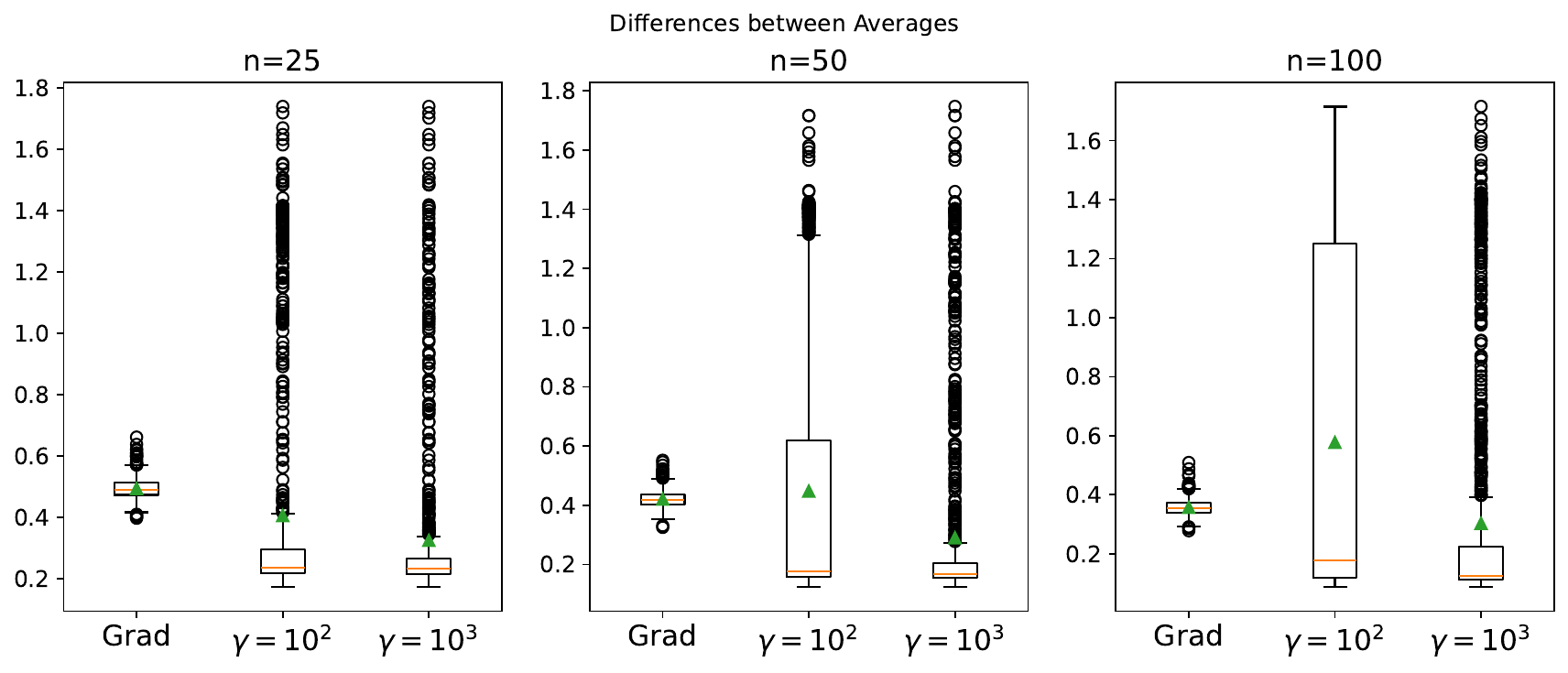}
\includegraphics[width=0.72\linewidth]{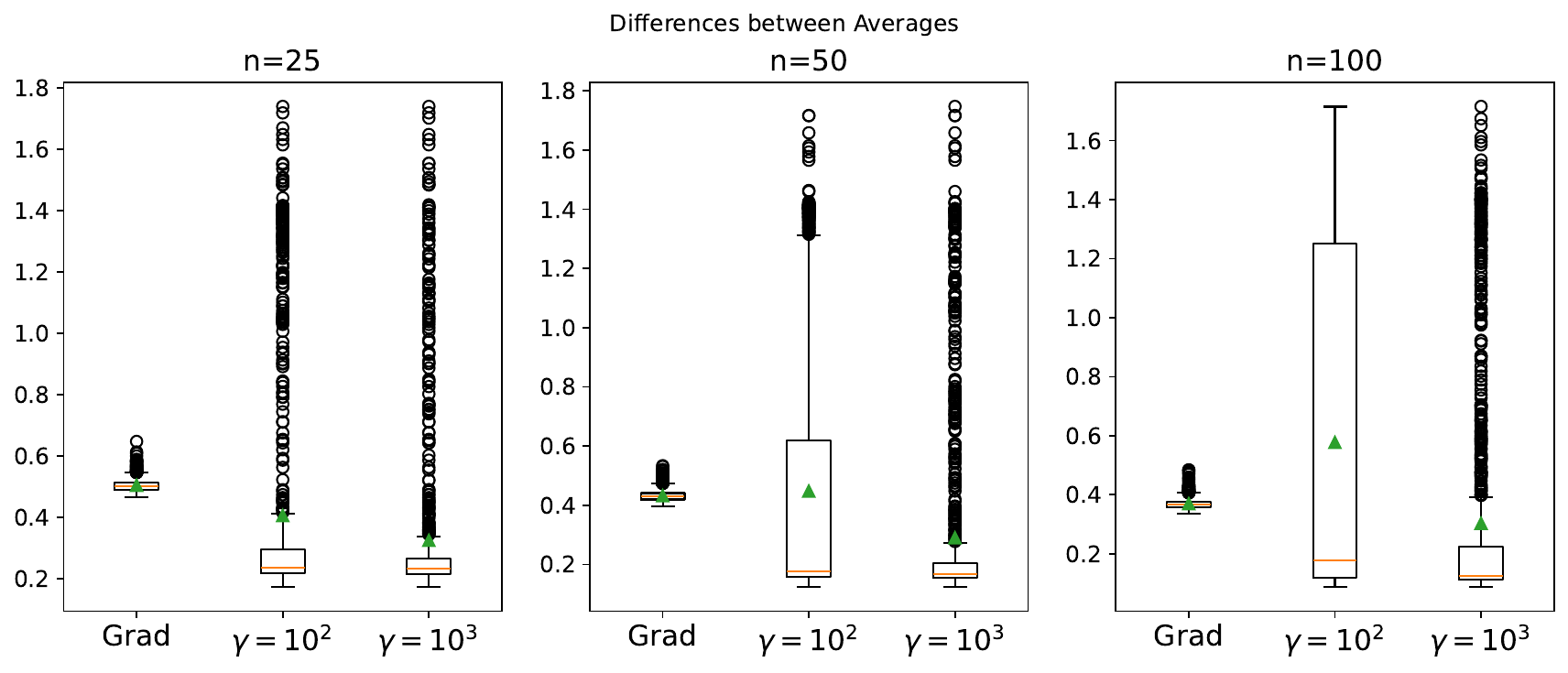}

    \caption{Convergence statistics for ResNet-50. Lower is better. First row: no normalization, photometric augmentation. Second row: no normalization, noise augmentation. Third row: $\ell_2$-normalization, photometric augmentation. Fourth row: $\ell_2$-normalization, noise augmentation.}

\end{figure}


\end{document}